\documentclass[10pt]{article}

\usepackage[accepted]{sty/tmlr}
\usepackage[pdfusetitle]{hyperref}
\usepackage{url}

\usepackage{amsmath, amsfonts, amssymb, amsthm, mathtools, empheq, dsfont}
\usepackage{multirow, booktabs}
\usepackage{siunitx}

\usepackage[ruled, lined, linesnumbered, noend]{algorithm2e}

\usepackage{float, wrapfig, subcaption}
\usepackage{tikz}

\usetikzlibrary{arrows, arrows.meta, patterns, automata, calc, positioning}
\allowdisplaybreaks
\newenvironment{algo}[1][0.8\textwidth]{%
    \begin{center}\begin{minipage}{#1}\begin{algorithm}[H]}{%
    \end{algorithm}\end{minipage}\end{center}}

\def\ie{i.e.\@}

\def\resp{resp.\@}

\def\epsilon{\varepsilon}

\def\A{\mathcal A}
\def\D{\mathcal D}

\def\S{\mathcal S}
\def\O{\mathcal O}

\def\Q{\mathcal Q}

\def\B{\mathcal B}

\def\C{\mathcal C}
\def\F{\mathcal F}
\def\U{\mathcal U}
\def\P{\mathcal P}
\def\X{\mathcal X}
\def\Y{\mathcal Y}
\def\L{\mathcal L}
\def\M{\mathcal M}
\def\H{\mathcal H}
\def\R{\mathbb R}

\def\N{\mathbb N}
\def\a{\mathbf a}

\def\c{\mathbf c}

\def\h{\mathbf h}
\def\i{\mathbf i}
\def\m{\mathbf m}
\def\o{\mathbf o}

\def\s{\mathbf s}

\def\x{\mathbf x}
\def\y{\mathbf y}

\def\d#1{\operatorname{d}\!{#1}}

\def\0{\mathbf 0}
\def\1{\mathds 1}

\def\->{\rightarrow}
\def\=>{\Rightarrow}
\def\<=>{\Leftrightarrow}

\def\E#1#2{\mathop{\underset{\substack{#1}}{\mathbb{E}} \left[ #2 \right]}}
\DeclareMathOperator*{\argmax}{arg\,max}

\def\texpdf#1#2{\texorpdfstring{#1}{#2}}
\def\cqf{\texpdf{$\mathcal{Q}$}{Q}-function}
\def\qf{\texpdf{$Q$}{Q}-function}

\title{%
    Recurrent networks, hidden states and beliefs in partially observable
    environments}

\author{%
    \name Gaspard Lambrechts \email gaspard.lambrechts@uliege.be \\
    \addr Montefiore Institute, University of Liège
    \AND
    \name Adrien Bolland \email adrien.bolland@uliege.be \\
    \addr Montefiore Institute, University of Liège
    \AND
    \name Damien Ernst \email dernst@uliege.be \\
    \addr Montefiore Institute, University of Liège \\
        LTCI, Telecom Paris, Institut Polytechnique de Paris}

\def\month{08}
\def\year{2022}

\def\openreview{\url{https://openreview.net/forum?id=dkHfV3wB2l}}

\def\peartm{\num{0.8233}}
\def\pearltm{\num{0.5347}}
\def\pearstm{\num{0.5460}}
\def\pearmh{\num{0.5948}}
\def\pearvmh{\num{0.5982}}

\def\speartm{\num{0.6419}}
\def\spearltm{\num{0.6666}}
\def\spearstm{\num{0.6403}}
\def\spearmh{\num{0.2965}}
\def\spearvmh{\num{0.6176}}

\begin{document}

\maketitle

\begin{abstract}

    Reinforcement learning aims to learn optimal policies from interaction with
    environments whose dynamics are unknown. Many methods rely on the
    approximation of a value function to derive near-optimal policies. In
    partially observable environments, these functions depend on the complete
    sequence of observations and past actions, called the history. In this
    work, we show empirically that recurrent neural networks trained to
    approximate such value functions internally filter the posterior
    probability distribution of the current state given the history, called the
    belief. More precisely, we show that, as a recurrent neural network learns
    the \cqf{}, its hidden states become more and more correlated with the
    beliefs of state variables that are relevant to optimal control. This
    correlation is measured through their mutual information. In addition, we
    show that the expected return of an agent increases with the ability of its
    recurrent architecture to reach a high mutual information between its
    hidden states and the beliefs. Finally, we show that the mutual information
    between the hidden states and the beliefs of variables that are irrelevant
    for optimal control decreases through the learning process. In summary,
    this work shows that in its hidden states, a recurrent neural network
    approximating the \cqf{} of a partially observable environment reproduces a
    sufficient statistic from the history that is correlated to the relevant
    part of the belief for taking optimal actions.

\end{abstract}

\section{Introduction} \label{sec:intro}

Latest advances in reinforcement learning (RL) rely heavily on the ability to
approximate a value function (\ie, state or state-action value function).
Modern RL algorithms have been shown to be able to produce approximations of
the value functions of Markov decision processes (MDPs) from which high-quality
policies can be derived, even in the case of continuous and high-dimensional
state and action spaces \citep{mnih2015human, lillicrap2015continuous,
mnih2016asynchronous, haarnoja2018soft, hessel2018rainbow}. The adaptation of
these techniques to partially observable MDPs (POMDPs) is not straightforward.
Indeed, in such environments, the agent only receives partial observations of
the underlying states of the environment. Unlike MDPs where the value functions
are written as functions of the current state, in POMDPs the value functions
are written as functions of the complete sequence of observations and past
actions, called the history. Moreover, the value functions of a history can
equivalently be written as functions of the posterior probability distribution
over the current state given this history \citep{bertsekas2012dynamic}. This
posterior probability distribution is called the belief and is said to be a
sufficient statistic from the history for the value functions of the POMDP.
However, the computation of the belief requires one to know the POMDP model and
is generally intractable with large or continuous state spaces. For these two
reasons, practical RL algorithms rely on the definition of the value functions
as functions of the complete history (\ie, history or history-action value
function), while the definition of the value functions as functions of the
belief (\ie, belief or belief-action value function) is more of theoretical
interest.

Approximating the value functions as functions of the histories requires one to
use function approximators that are able to process sequences of arbitrary
length. In practice, RNNs are good candidates for such approximators
\citep{bakker2001reinforcement, hausknecht2015deep, heess2015memory}. RNNs are
parametric approximators that process sequences, time step by time step,
exhibiting memory through a hidden state that is passed recurrently over time.
The RNN is thus tasked with outputting the value directly from the history. We
focus on the approximation of the history-action value function, or \cqf{}, in
POMDPs using a parametric recurrent Q-learning (PRQL) algorithm. More
precisely, RNNs are trained with the deep recurrent Q-network (DRQN) algorithm
\citep{hausknecht2015deep, zhu2017improving}.

Since we know that the belief is a sufficient statistic from the history for
the \cqf{} of this history \citep{bertsekas2012dynamic}, we investigate if
RNNs, once trained, reproduce the belief filter when processing a history. This
investigation is conducted in this work by studying the performance of the
different agents with regard to the mutual information (MI) between their
hidden states and the belief. We focus on POMDPs for which the models are
known. The benchmark problems chosen are the T-Maze environments
\citep{bakker2001reinforcement} and the Mountain Hike environments
\citep{igl2018deep}. The first ones present a discrete state space, allowing
one to compute the belief using Bayes' rule, and representing this distribution
over the states in a vector whose dimension is equal to the number of distinct
states. The second ones present a continuous state space, making the belief
update intractable. We thus rely on particle filtering in order to approximate
the belief by a set of states, called particles, distributed according to the
belief distribution. The MI between the hidden states and the beliefs is
periodically estimated during training, using the mutual information neural
estimator (MINE) algorithm \citep{belghazi2018mutual}. The MINE estimator is
extended with the Deep Set architecture \citep{zaheer2017deep} in order to
process sets of particles in the case of POMDPs with continuous state-spaces.
This methodology allows one to measure the ability and tendency of recurrent
architecture to reproduce the belief filter when trained to approximate the
\cqf{}.

In \citep{mikulik2020meta}, a similar study is performed in the meta-learning
setting. In this setting, an MDP is drawn from a distribution of MDPs at each
episode. This problem can be equivalently modeled as a particular subclass of
POMDP. The authors show empirically, among others, that the hidden state of an
RNN-based policy and the statistic of the optimal policy can be mapped one into
the other with a low dissimilarity measure. In contrast, we consider arbitrary
POMDPs and show empirically that information about the belief, a statistic
known to be sufficient for the optimal control, is encoded in the hidden
states.

In \autoref{sec:background}, we formalise the problem of optimal control in
POMDPs, we present the PRQL algorithms for deriving near-optimal policies and
we explain the MINE algorithm for estimating the MI. In \autoref{sec:method},
the beliefs and hidden states are defined as random variables whose MI is
measured. Afterwards, \autoref{sec:experiments} displays the main results
obtained for the previously mentioned POMDPs. Finally,
\autoref{sec:conclusions} concludes and proposes several future works and
algorithms motivated by our results.

\section{Background} \label{sec:background}

In \autoref{subsec:pomdp}, POMDPs are introduced, along with the belief,
policy, and \qf{}s associated with such decision processes. Afterwards, in
\autoref{subsec:prql}, we introduce the DRQN algorithm that is used in our
experiments. This algorithm is a particular instance of the PRQL class of
algorithms that allows to approximate the \cqf{} for deriving a near-optimal
policy in a POMDP. Finally, in \autoref{subsec:mine}, we present the MINE
algorithm that is used for estimating the MI between the hidden states and
beliefs in our experiments.

\subsection{Partially observable Markov decision processes}
\label{subsec:pomdp}

In this work, the environments are modelled as POMDPs. Formally, a POMDP $P$ is
an $8$-tuple $P = (\S, \A, \O, p_0, T, R, O, \gamma)$ where $\S$ is the state
space, $\A$ is the action space, and $\O$ is the observation space. The initial
state distribution $p_0$ gives the probability $p_0(\s_0)$ of $\s_0 \in \S$
being the initial state of the decision process. The dynamics are described by
the transition distribution $T$ that gives the probability $T(\s_{t+1} \mid
\s_t, \a_t)$ of $\s_{t+1} \in \S$ being the state resulting from action $\a_t
\in \A$ in state $\s_t \in \S$. The reward function $R$ gives the immediate
reward $r_t = R(\s_t, \a_t, \s_{t+1})$ obtained after each transition. The
observation distribution $O$ gives the probability $O(\o_t \mid \s_t)$ to get
observation $\o_t \in \O$ in state $\s_t \in \S$. Finally, the discount factor
$\gamma \in [0, 1[$ gives the relative importance of future rewards.

Taking a sequence of $t$ actions ($\a_{0:t-1}$) in the POMDP conditions its
execution and provides a sequence of $t+1$ observations ($\o_{0:t}$). Together,
they compose the history $\eta_{0:t} = (\o_{0:t}, \a_{0:t-1}) \in \H_{0:t}$
until time step $t$, where $\H_{0:t}$ is the set of such histories. Let $\eta
\in \H$ denote a history of arbitrary length sampled in the POMDP, and let $\H
= \bigcup_{t = 0}^\infty \H_{0:t}$ denote the set of histories of arbitrary
length.

A policy $\pi \in \Pi$ in a POMDP is a mapping from histories to actions, where
$\Pi = \H \-> \A$ is the set of such mappings. A policy $\pi^* \in \Pi$ is said
to be optimal when it maximises the expected discounted sum of future rewards
starting from any history $\eta_{0:t} \in \H_{0:t}$ at time $t \in \N_0$
\begin{equation}
    \pi^* \in \argmax_{\pi \in \Pi} \E{\pi, P}{
        \sum_{t' = t}^{\infty} \gamma^{t' - t} r_{t'}
        \;\middle\vert\; \eta_{0:t}
    }, \; \forall \eta_{0:t} \in \H_{0:t}, \; \forall t \in \N_0 .
    \label{eq:optimal_policy}
\end{equation}
The history-action value function, or \cqf{}, is defined as the maximal
expected discounted reward that can be gathered, starting from a history
$\eta_{0:t} \in \H_{0:t}$ at time $t \in \N_0$ and an action $\a_t \in \A$
\begin{equation}
    \Q(\eta_{0:t}, \a_t) = \max_{\pi \in \Pi} \E{\pi, P}{
        \sum_{t' = t}^{\infty} \gamma^{t' - t} r_{t'}
        \;\middle\vert\; \eta_{0:t}, \a_t
    }, \;
    \forall \eta_{0:t} \in \H_{0:t}, \;
    \forall \a_t \in \A, \;
    \forall t \in \N_0 .
    \label{eq:cqf}
\end{equation}
The \cqf{} is also the unique solution of the Bellman equation
\citep{smallwood1973optimal, kaelbling1998planning, porta2004value}
\begin{equation}
    \Q(\eta, \a) = \E{P}{
        r + \gamma \max_{\a' \in \A} \Q(\eta', \a')
        \;\middle\vert\; \eta, \a
    }, \;
    \forall \eta \in \H, \;
    \forall \a \in \A
    \label{eq:cqf_bellman}
\end{equation}
where $\eta' = \eta \cup (\a, \o')$ and $r$ is the immediate reward obtained
when taking action $\a$ in history $\eta$. From equation
\eqref{eq:optimal_policy} and equation \eqref{eq:cqf}, it can be observed that
any optimal policy satisfies
\begin{equation}
    \pi^*(\eta) \in \argmax_{\a \in \A} \Q(\eta, \a),
    \; \forall \eta \in \H .
    \label{eq:q_policy}
\end{equation}

Let $\P(\S)$ be the set of probability measures over the state space $\S$. The
belief $b \in \P(\S)$ of a history $\eta \in \H$ is defined as the posterior
probability distribution over the states given the history, such that $b(\s) =
p(\s \mid \eta), \; \forall \s \in \S$ \citep{thrun2002probabilistic}. The
belief filter $f^*$ is defined as the function that maps a history $\eta$ to
its corresponding belief $b$
\begin{equation}
    f^*(\eta) = b, \; \forall \eta \in \H. \label{eq:belief_filter}
\end{equation}
Formally, for an initial observation $\eta = (\o)$, the belief $b = f^*(\eta)$
is defined by
\begin{equation}
    b(\s)
    = \frac{
        p_0(\s) O(\o \mid \s)
    }{
        \int_{\S} p_0(\s') O(\o \mid \s') \d{\s'}
    }, \; \forall \s \in \S
    \label{eq:initial_belief}
\end{equation}
and for a history $\eta' = \eta \cup (\a, \o')$, the belief $b' = f^*(\eta')$
is recursively defined by
\begin{equation}
    b'(\s')
    = \frac{
        O(\o' \mid \s')
        \int_\S T(\s' \mid \s, \a) \; b(\s) \d{\s}
    }{
        \int_\S
        O(\o' \mid \s')
        \int_\S T(\s' \mid \s, \a) \; b(\s) \d{\s} \d{\s'}
    }, \; \forall \s' \in \S .
    \label{eq:belief_update_integrals}
\end{equation}
where $b = f^*(\eta)$. Equation \eqref{eq:belief_update_integrals} provides a
way to update the belief $b$ to $b'$ through a filter step $f$ once observing
new information $(\a, \o')$
\begin{equation}
    b' = f(b; \a, \o') . \label{eq:belief_update}
\end{equation}

A statistic from the history is defined as any function of the history. The
belief is known to be a sufficient statistic from the history in order to act
optimally \citep{bertsekas2012dynamic}. It means that the \cqf{} only depends
on the history through the belief computed from this same history. It implies
in particular that the \cqf{} takes the following form
\begin{equation}
    \Q(\eta, \a) = Q(f^*(\eta), \a),
    \; \forall \eta \in \H, \; \forall \a \in \A
    \label{eq:cqf_composition}
\end{equation}
where $Q: \P(\S) \times \A \rightarrow \R$ is called the belief-action value
function, or \qf{}. This function gives the maximal expected discounted reward
starting from a belief $b \in \P(\S)$ and an action $\a \in \A$, where the
belief $b = f^*(\eta)$ results from an arbitrary history $\eta \in \H$.
Although the exact belief filter is often unknown or intractable, this
factorisation of the \cqf{} still motivates the compression of the history in a
statistic related to the belief, when processing the history for predicting the
\cqf{}.

\subsection{Parametric recurrent Q-learning} \label{subsec:prql}

We call PRQL the family of algorithms that aim at learning an approximation of
the \cqf{} with a recurrent architecture $\Q_\theta$, where $\theta \in
\R^{d_\theta}$ is the parameter vector. These algorithms are motivated by
equation \eqref{eq:q_policy} that shows that an optimal policy can be derived
from the \cqf{}. The strategy consists of minimising, with respect to $\theta$,
for all $(\eta, \a)$, the distance between the estimation $\Q_\theta(\eta, \a)$
of the LHS of equation \eqref{eq:cqf_bellman}, and the estimation of the
expectation $\mathbb{E}_P[r + \gamma \max_{\a' \in \A} \Q_\theta(\eta', \a')]$
of the RHS of equation \eqref{eq:cqf_bellman}. This is done by using
transitions $(\eta, \a, r, \o', \eta')$ sampled in the POMDP, with $\eta' =
\eta \cup (\a, \o')$. In its simplest form, given such a transition, the PRQL
algorithm updates the parameters $\theta \in \R^{d_\theta}$ of the function
approximator according to
\begin{align}
    \theta \leftarrow \theta + \alpha \left( r + \gamma \max_{\a' \in \A}
    \left\{ \Q_\theta(\eta', \a') \right\} - \Q_\theta(\eta, \a)
    \right) \nabla_{\!\theta}{\Q_\theta}(\eta, \a) .
    \label{eq:prql}
\end{align}
This update corresponds to a gradient step in the direction that minimises,
with respect to $\theta$ the squared distance between $\Q_\theta(\eta, \a)$ and
the target $r + \gamma \max_{\a' \in \A} \left\{ \Q_\theta(\eta', \a')
\right\}$ considered independent of $\theta$. It can be noted that, in
practice, such algorithms introduce a truncation horizon $H$ such that the
histories generated in the POMDP have a maximum length of $H$. From the
approximation $\Q_\theta$, the policy $\pi_\theta$ is given by
$\pi_\theta(\eta) = \argmax_{\a \in \A} \Q_\theta(\eta, \a)$. Equation
\eqref{eq:q_policy} guarantees the optimality of this policy if $\Q_{\theta} =
\Q$. Even though it will alter the performance of the algorithm, any policy can
be used to sample the transitions $(\eta, \a, r, \o', \eta')$.

The function approximator $\Q_\theta$ of PRQL algorithms should be able to
process inputs $\eta \in \H$ of arbitrary length, making RNN approximators a
suitable choice. Indeed, RNNs process the inputs sequentially, exhibiting
memory through hidden states that are outputted after each time step, and
processed at the next time step along with the following input. More formally,
let $\x_{0:t} = [\x_0, \dots, \x_t]$ with $t \in \N_0$ be an input sequence. At
any step $k \in \{0, \dots, t\}$, RNNs maintain an internal memory state $\h_k$
through the update function \eqref{eq:rnn_update} and output a value $\y_k$
through the output function \eqref{eq:rnn_output}. The initial state $\h_{-1}$
is given by the initialization function \eqref{eq:rnn_initialization}.
\begin{align}
    \h_{k} &= u_\theta(\h_{k-1}, \x_k), \;
    \forall k \in \N_0, \label{eq:rnn_update} \\
    \y_k &= o_\theta(\h_k), \;
    \forall k \in \N_0, \label{eq:rnn_output} \\
    \h_{-1} &= i_\theta. \label{eq:rnn_initialization}
\end{align}
These networks are trained based on backpropagation through time where
gradients are computed in a backward pass through the complete sequence via the
hidden states \citep{werbos1990backpropagation}. The following recurrent
architectures are used in the experiments: the long short-term memory (LSTM) by
\citet{hochreiter1997long}, the gated recurrent unit (GRU) by
\citet{chung2014empirical}, the bistable recurrent cell (BRC) and recurrently
neuromodulated bistable recurrent cell (nBRC) by \citet{vecoven2021bio}, and
the minimal gated unit (MGU) by \citet{zhou2016minimal}.

In the experiments, we use the DRQN algorithm \citep{hausknecht2015deep,
zhu2017improving} to learn policies. This algorithm is a PRQL algorithm that
shows good convergence even for high-dimensional problems. The DRQN algorithm
is detailed in \autoref{algo:drqn} of \autoref{app:drqn}. In this algorithm,
for a given history $\eta_{0:t}$ of arbitrary length $t$, the inputs of the RNN
are $\x_k = (\a_{k-1}, \o_k), \; k = 1, \dots, t$ and $\x_0 = (\mathbf{0},
\o_0)$, and the output of the RNN at the last time step $\y_t = o_\theta(\h_t)
\in \R^{|\A|}$ gives $\y_t^{\a_t} = \Q_\theta(\eta_{0:t}, \a_t)$, for any $\a_t
\in \A$. We also define the composition $u^*_\theta: \H \rightarrow
\R^{d_\theta}$ of equation \eqref{eq:rnn_initialization} and equation
\eqref{eq:rnn_update} applied on the complete history, such that
\begin{align}
    \h_t
    = u^*_\theta(\eta_{0:t})
    = \begin{cases}
        u_\theta(u^*_\theta(\eta_{0:t-1}), \x_t), & t \geq 1 \\
        u_\theta(i_\theta, \x_t), & t = 0
    \end{cases}
    \label{eq:rnn_hidden_state}
\end{align}

\subsection{Mutual information neural estimator} \label{subsec:mine}

In this work, we are interested in establishing if a recurrent function
approximator reproduces the belief filter during PRQL. Formally, this is
performed by estimating the MI between the beliefs and the hidden states of the
RNN approximator $\Q_\theta$. In this subsection, we recall the concept of MI
and how it can be estimated in practice.

The MI is theoretically able to measure any kind of dependency between random
variables \citep{kraskov2004estimating}. The MI between two jointly continuous
random variables $X$ and $Y$ is defined as
\begin{equation}
    I(X; Y) = \int_{\X} \int_{\Y} p(x, y)
        \log \frac{p(x, y)}{p_X(x) \; p_Y(y)} \d{x} \d{y}
\end{equation}
where $\X$ and $\Y$ are the support of the random variables $X$ and $Y$
respectively, $p$ is the joint probability density function of $X$ and $Y$, and
$p_X$ and $p_Y$ are the marginal probability density functions of $X$ and $Y$,
respectively. It is worth noting that the MI can be defined in terms of the
Kulback-Leibler (KL) divergence between the joint $p$ and the product of the
marginals $q = p_X \otimes p_Y$, over the joint space $\mathcal{Z} = \X \times
\Y$
\begin{equation}
    I(X; Y) = D_\text{KL}(p \mid\mid q)
    = \int_{\mathcal{Z}} p(z) \log\left(\frac{p(z)}{q(z)}\right) \d{z}
\end{equation}

In order to estimate the MI between random variables $X$ and $Y$ from a dataset
$\left\{ (x_i, y_i) \right\}_{i = 1}^N$, we rely on the MINE algorithm
\citep{belghazi2018mutual}. This technique is a parametric approach where a
neural network outputs a lower bound on the MI, that is maximised by gradient
ascent. The lower bound is derived from the Donsker-Varhadan representation of
the KL-divergence \citep{donsker1975asymptotic}
\begin{equation}
    D_\text{KL}(p \mid\mid q) = \sup_{T: \mathcal{Z} \-> \mathbb{R}}
    \E{z \sim p}{T(z)} - \log\left(\E{z \sim q}{e^{T(z)}}\right)
\end{equation}
where the supremum is taken over all functions $T$ such that the two
expectations are finite. The lower bound $I_\Phi(X; Y)$ on the true MI $I(X;
Y)$ is obtained by replacing $T$ by a parameterised function $T_\phi:
\mathcal{Z} \-> \R$ with $\phi \in \Phi$, and taking the supremum over the
parameter space $\Phi$ of this function. If $\Phi$ corresponds to the parameter
space of a neural network, then this lower bound can be approached by gradient
ascent using empirical means as estimators of the expectations. The resulting
procedure for estimating the MI is given in \autoref{algo:mine} in
\autoref{app:mine}.

\section{Measuring the correlation between the hidden states and beliefs}
\label{sec:method}

In this work, we study if PRQL implicitly approximates the belief filter by
reaching a high MI between the RNN's hidden states and the beliefs, that are
both generated from random histories. In this section, we first explain the
intuition behind this hypothesis, then we define the joint probability
distribution over the hidden states and beliefs that defines the MI.

As explained in \autoref{sec:background}, the belief filter is generally
intractable. As a consequence, PRQL algorithms use approximators $\Q_\theta$
that directly take the histories as input. In the DRQN algorithm, these
histories are processed recurrently according to equation
\eqref{eq:rnn_update}, producing a new hidden state $\h_t$ after each input
$\x_t = (\a_{t-1}, \o_t)$
\begin{equation}
    \h_t = u_\theta(\h_{t-1}; (\a_{t-1}, \o_t)) .
    \label{eq:h_update}
\end{equation}
These hidden states should thus summarise all relevant information from past
inputs in order to predict the \cqf{} at all later time steps. The belief is
known to be a sufficient statistic from the history for these predictions
\eqref{eq:cqf_composition}. Moreover, the belief $b_t$ is also updated
recurrently, according to equation \eqref{eq:belief_update} after each
transition $(\a_{t-1}, \o_t)$
\begin{equation}
    b_t = f(b_{t-1}; \a_{t-1}, \o_t) .
    \label{eq:b_update}
\end{equation}
The parallel between equation \eqref{eq:h_update} and equation
\eqref{eq:b_update}, knowing the sufficiency of the belief
\eqref{eq:cqf_composition}, justifies the appropriateness of the belief filter
$f$ as the update function $u_\theta$ of the RNN approximator $\Q_\theta$. It
motivates the study of the reconstruction of the belief filter by the RNN.

In practice, this is done through the measurement of the MI between the hidden
state $\h_t$ and the belief $b_t$ at any time step $t \in \N_0$. Formally, for
a given history length $t \in \N_0$, the policy $\pi_{\theta}$ of the learning
algorithm, as defined in \autoref{subsec:prql}, induces a distribution
$p_{\pi_{\theta}}(\eta \mid t)$ over histories $\eta \in \H$. This conditional
probability distribution is zero for all history of length $t' \neq t$. Given a
distribution $p(t)$ over the length of trajectories, the joint distribution of
$\h$ and $b$ is given by
\begin{align}
    p(\h, b)
    &= \sum_{t = 0}^\infty \; p(t)
    \int_{\H}
        p(\h, b \mid \eta) \;
        p_{\pi_{\theta}}(\eta \mid t)
        \d{\eta}
    \label{eq:hb_distribution}
\end{align}
where $p(\h, b \mid \eta)$ is a Dirac distribution for $\h = u^*_\theta(\eta)$
and $b = f^*(\eta)$ given by equation \eqref{eq:rnn_hidden_state} and equation
\eqref{eq:belief_filter}, respectively. In the following, we estimate the MI
between $\h$ and $b$ under their joint distribution \eqref{eq:hb_distribution}.

\section{Experiments} \label{sec:experiments}

In this section, the experimental protocol and environments are described and
the results are given. More specifically, in \autoref{subsec:protocol}, we
describe the estimates that are reported in the figures. The results are
reported for four different POMDPs: the T-Maze and Stochastic T-Maze in
\autoref{subsec:tmaze}, and the Mountain Hike and Varying Mountain Hike in
\autoref{subsec:hike}. Afterwards, in \autoref{subsec:irrelevant}, irrelevant
state variables and observations are added to the decision processes, and the
MI is measured separately between the hidden states and the belief of the
relevant and irrelevant variables. Finally, in \autoref{subsec:discussion}, we
discuss the results obtained in this section, and propose an additional
protocol to study their generalisation.

\subsection{Experimental protocol} \label{subsec:protocol}

As explained in \autoref{subsec:prql}, the parameters $\theta$ of the
approximation $\Q_\theta$ are optimised with the DRQN algorithm. After $e$
episodes of interaction with the POMDP, the DRQN algorithm gives the policy
$\pi_{\theta_e}(\eta) = \argmax_{\a \in \A} \Q_{\theta_e}(\eta, \a)$. In the
experiments, the empirical cumulative reward $\hat{J}(\theta_e)$ of the policy
$\pi_{\theta_e}$ is reported, along with the estimated MI $\hat{I}(\theta_e)$
between the random variables $\h$ and $b$ under the distribution
\eqref{eq:hb_distribution} implied by $\pi_{\theta_e}$. Each estimate is
reported averaged over four training sessions. In addition, confidence
intervals show the minimum and maximum of these estimates.

The empirical return is defined as $\hat{J}(\theta_e) = \frac{1}{I}\sum_{i =
0}^{I-1} \sum_{t = 0}^{H-1} \gamma^t r_t^{i}$, where $I$ is the number of Monte
Carlo rollouts, $H$ the truncation horizon of the DRQN algorithm, and $r_t^i$
is the reward obtained at time step $t$ of Monte Carlo rollout $i$. As far as
the estimation of the MI is concerned, we sample time steps with equal
probability $p(t) = 1/H, \; t \in \{0, \dots, H - 1\}$, where $H$ is the
truncation horizon of the DRQN algorithm. The uniform distribution over time
steps and the current policy $\pi_{\theta_e}$ define the probability
distribution \eqref{eq:hb_distribution} over the hidden states and beliefs. The
MI is estimated from samples of this distribution using the MINE estimator
$\hat{I}(\theta_e)$ (see \autoref{subsec:mine_estimator} for details). The
hyperparameters of the DRQN and MINE algorithms are given in
\autoref{app:hyperparameters}.

For POMDPs with continuous state spaces, the computation of the belief $b$ is
intractable. However, a set of state particles $S$ that follows the belief
distribution $f^*(\eta)$ can be sampled, using particle filtering (see
\autoref{app:particle}). This set of particles could be used to construct an
approximation of the belief in order to estimate the MI. This density
estimation procedure is nonetheless unnecessary as the MINE network can
directly process the set of particles by producing a permutation-invariant
embedding of the belief using the Deep Set architecture \citep{zaheer2017deep},
see \autoref{subsec:deepsets} for details.

\subsection{Deterministic and Stochastic T-Mazes} \label{subsec:tmaze}

\begin{wrapfigure}{r}{0.35\textwidth}
    \centering
    \vspace{-1.5em}
    \resizebox{0.35\textwidth}{!}{\begin{tikzpicture}[scale=1.5]

    \pgfmathsetmacro{\L}{6}

    \node (mup) at (0.5, 5.5) {$\text{Up}$} ;
    \node (mup) at (0.5, 1.5) {$\text{Down}$} ;

    \foreach \s in {0, 4} {
        \draw[lightgray] (0, \s) grid (\L, \s + 1) ;
        \draw[lightgray] (\L, \s - 1) grid (\L + 1, \s + 2) ;

        \fill[blue, opacity=0.2] (0, \s) rectangle (1, \s + 1) ;
        \fill[pattern=north west lines, pattern color=blue!20]
            (\L, 6 * \s / 4 - 1) rectangle (\L + 1, 6 * \s / 4) ;
        \fill[gray, opacity=0.2] (\L, \s + 1) rectangle (\L + 1, \s + 2) ;
        \fill[gray, opacity=0.2] (\L, \s - 1) rectangle (\L + 1, \s + 0) ;

        \pgfmathsetmacro{\Lminustwo}{\L-2}
        \foreach \i in {0,...,\Lminustwo} {
            \node (c\i) at (\i + 0.5, \s + 0.5) {$(\i, 0)$} ;
        }

        \node (cdot) at (\L - 0.5, \s + 0.5) {$\dots$} ;
        \node (cL) at (\L + 0.5, \s + 0.5) {$(L, 0)$} ;
        \node (cL+1) at (\L + 0.5, \s + 1.5) {$(L, 1)$} ;
        \node (cL+2) at (\L + 0.5, \s - 0.5) {$(L, -1)$} ;
    }

\end{tikzpicture}}
    \vspace{-1em}
    \caption{T-Maze state space.}
    \vspace{-1em}
    \label{fig:tmaze_body}
\end{wrapfigure}
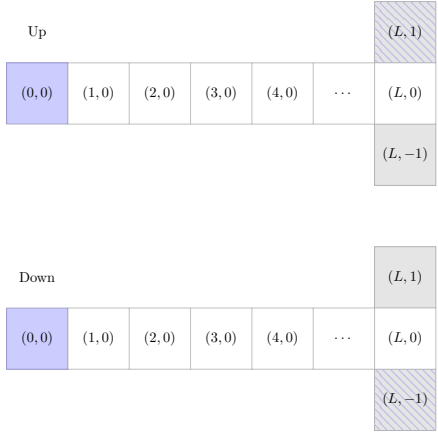

The T-Maze is a POMDP where the agent is tasked with finding the treasure in a
T-shaped maze (see \autoref{fig:tmaze_body}). The state is given by the
position of the agent in the maze and the maze layout that indicates whether
the treasure lies up or down after the crossroads. The initial state determines
the maze layout, and it never changes afterwards. The initial observation made
by the agent indicates the layout. Navigating in the maze provides zero reward,
except when bouncing onto a wall, in which case a reward of \num{-0.1} is
received. Finding the treasure provides a reward of \num{4}. Beyond the
crossroads, the states are always terminal. The optimal policy thus consists of
going through the maze, while remembering the initial observation in order to
take the correct direction at the crossroads. This POMDP is parameterised by
the corridor length $L \in \N$ and stochasticity rate $\lambda \in [0, 1]$ that
gives the probability of moving in a random direction at any time step. The
Deterministic T-Maze ($\lambda = 0$) was originally proposed in
\citep{bakker2001reinforcement}. The discount factor is $\gamma = 0.98$. This
POMDP is formally defined in \autoref{app:tmaze}.

As explained in \autoref{subsec:prql}, the histories can be sampled with an
arbitrary policy in PRQL algorithms. In practice, the DRQN algorithm uses an
$\varepsilon$-greedy stochastic policy that selects its action according to the
current policy with probability $1 - \varepsilon$, and according to the
exploration policy $\mathcal{E}(\A)$ with probability $\varepsilon$. Usually,
the exploration policy is chosen to be the uniform distribution $\U(\A)$ over
the action. However, for the T-Maze, the exploration policy $\mathcal{E}(\A)$
is tailored to this POMDP to alleviate the exploration problem, that is
independent of the study of this work. The exploration policy forces one to
walk through the right of the corridor with $\mathcal{E}(\text{Right}) = 1/2$
and $\mathcal{E}(\text{Other}) = 1/6$ where $\text{Other} \in \left\{
\text{Up}, \text{Left}, \text{Down} \right\}$.

\begin{figure}[!ht]
    \centering
    \begin{minipage}[t]{0.49\textwidth}
        \includegraphics[width=\textwidth]
            {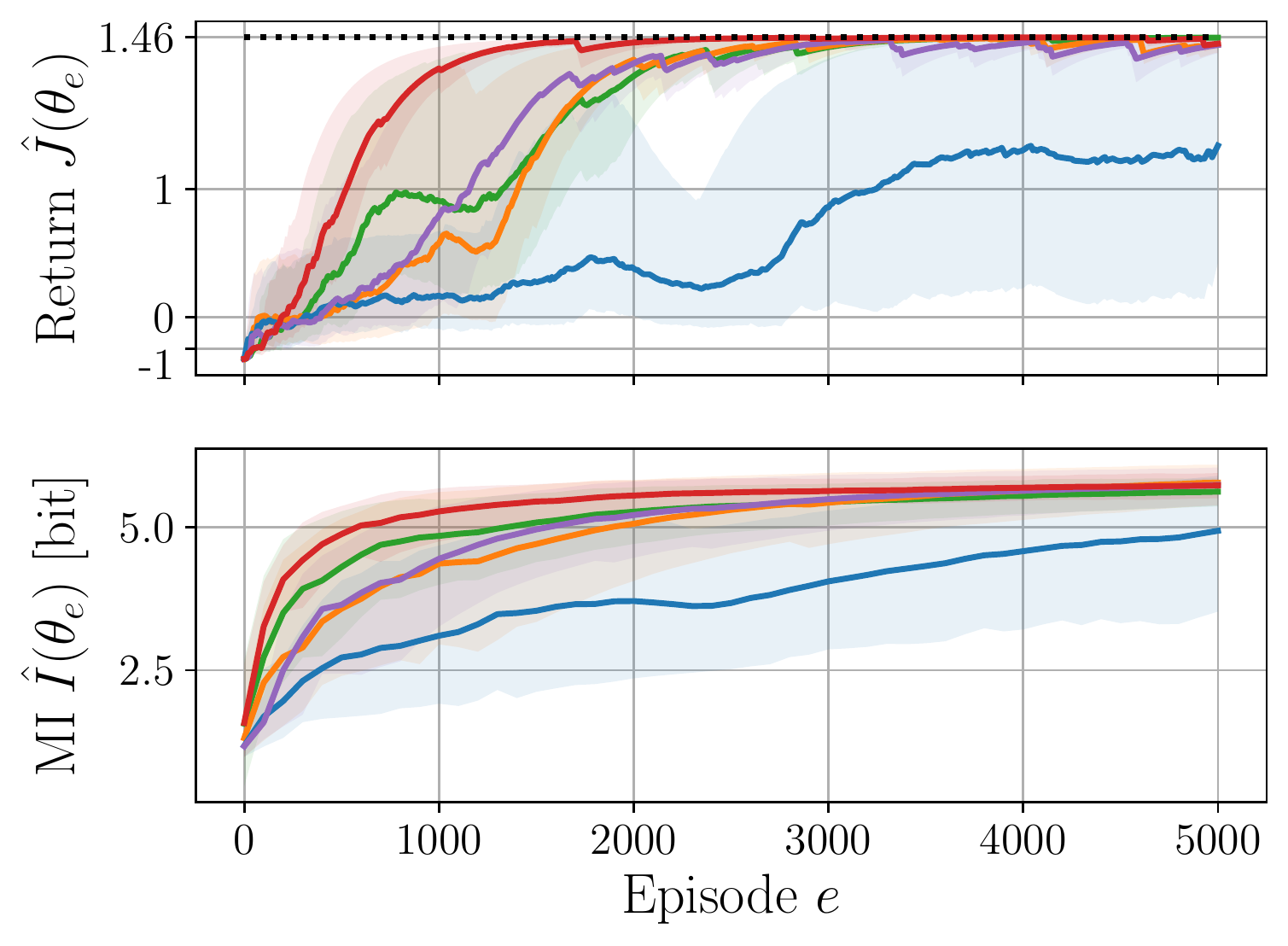}
        \label{fig:tmaze_50_learning}
    \end{minipage}
    \begin{minipage}[t]{0.49\textwidth}
        \includegraphics[width=\textwidth]
            {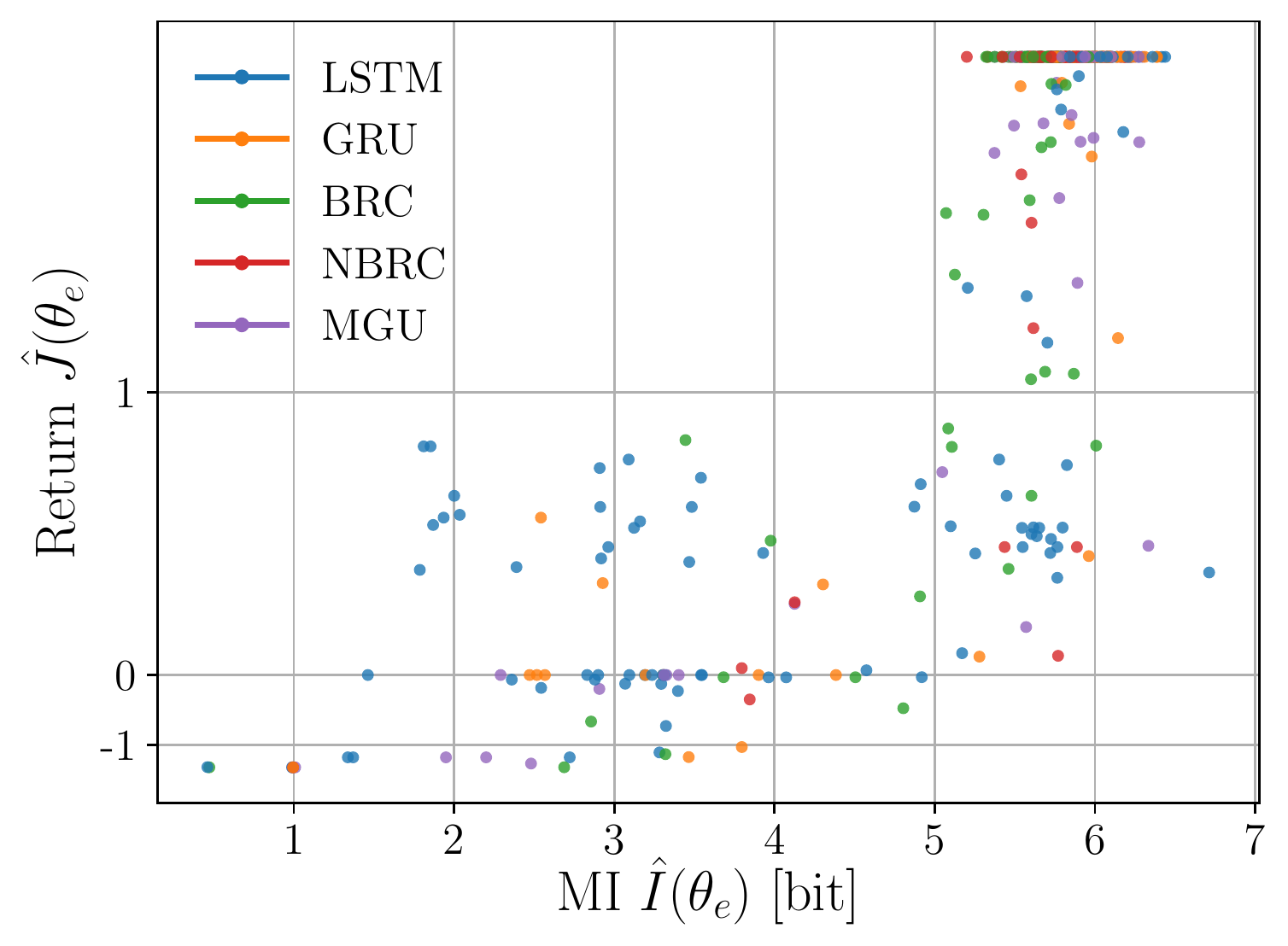}
        \label{fig:tmaze_50_corr}
    \end{minipage}
    \vspace{-1.5em}
    \caption{Deterministic T-Maze ($L = 50$).
        Evolution of the return $\hat{J}(\theta_e)$ and the MI
        $\hat{I}(\theta_e)$  after $e$ episodes (left), and the return
        $\hat{J}(\theta_e)$ with respect to the MI $\hat{I}(\theta_e)$
        (right). The maximal expected return is given by the dotted line.}
    \vspace{-0.5em}
    \label{fig:tmaze_50}
\end{figure}

On the left in \autoref{fig:tmaze_50}, the expected return is shown along with
the MI between the hidden states and the belief as a function of the number of
episodes, for a T-Maze of length $L = 50$. In order to better disambiguate
between high-quality policies, the empirical return is displayed with an
exponential scale in the following graphs. Both the performance of the policy
and the MI increase during training. We also observe that, at any given
episode, RNNs that have a higher return, such as the nBRC or the BRC,
correspond to cells that have a higher MI between their hidden states and the
belief. Furthermore, the LSTM that struggles to achieve a high return has a
significantly lower MI than the other cells. Finally, we can see that the
evolution of the MI and the return are correlated, which is highlighted on the
right in \autoref{fig:tmaze_50}. Indeed, the return increases with the MI, with
a linear correlation coefficient of \peartm{} and a rank correlation
coefficient of \speartm{}. These correlations coefficients are also detailed
for each cell separately in \autoref{app:correlations}. It can also be noted
that no RNN with less than \num{5} bits of MI reaches the maximal return.

\begin{figure}[!ht]
    \centering
    \begin{minipage}[t]{0.49\textwidth}
        \includegraphics[width=\textwidth]
            {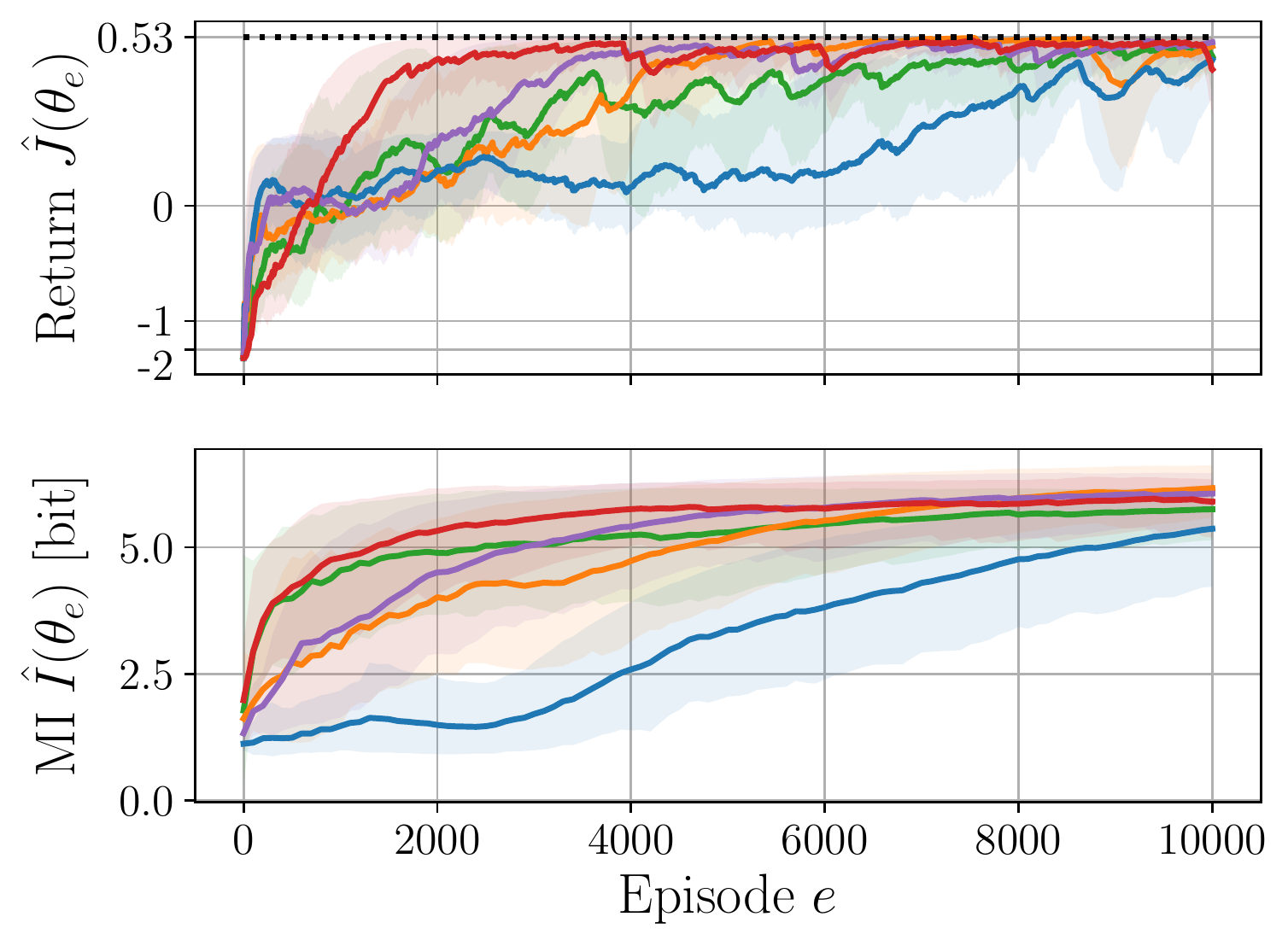}
        \label{fig:tmaze_100_learning}
    \end{minipage}
    \begin{minipage}[t]{0.49\textwidth}
        \includegraphics[width=\textwidth]
            {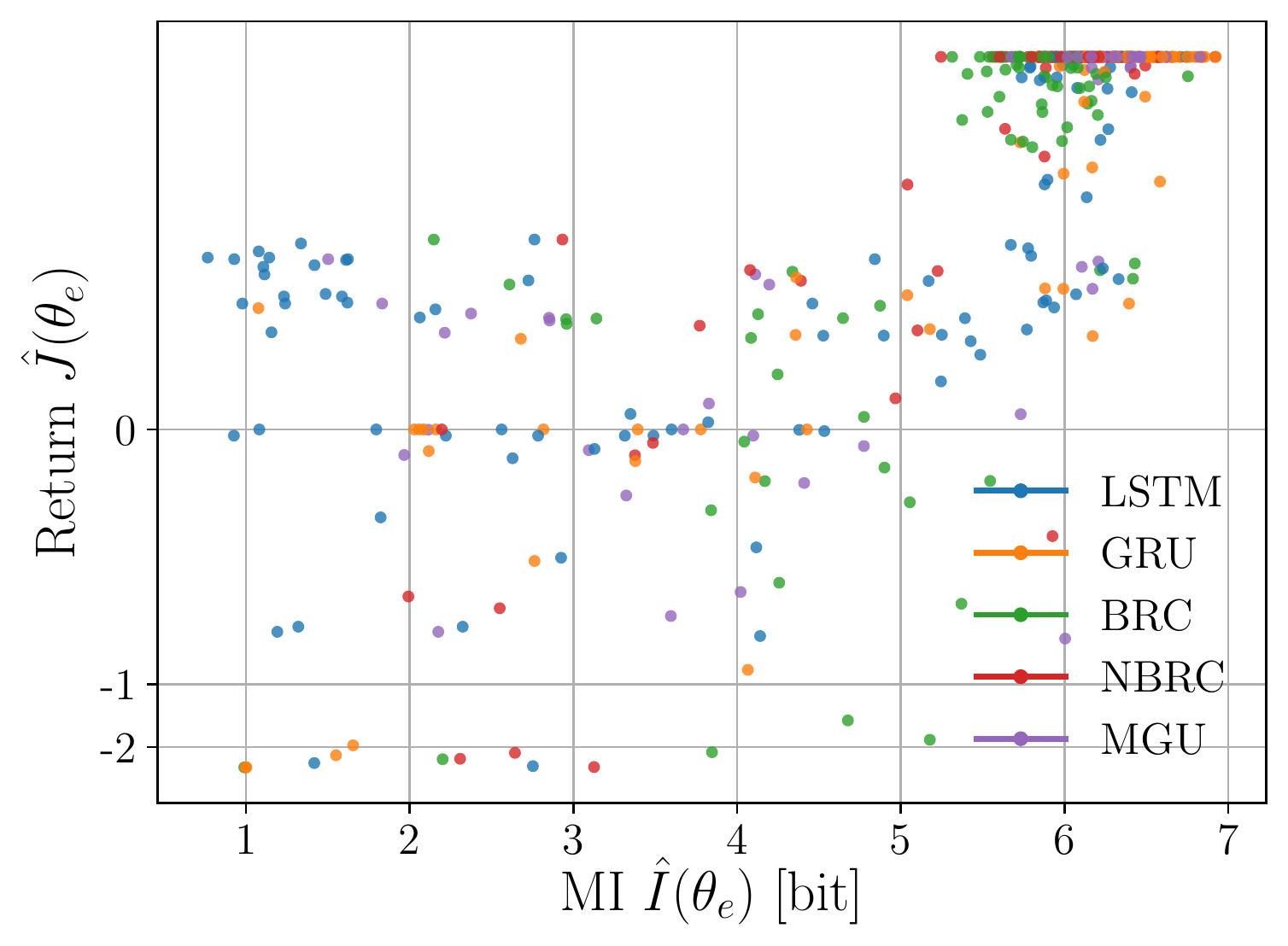}
        \label{fig:tmaze_100_corr}
    \end{minipage}
    \vspace{-1.5em}
    \caption{Deterministic T-Maze ($L = 100$).
        Evolution of the return $\hat{J}(\theta_e)$ and the MI
        $\hat{I}(\theta_e)$  after $e$ episodes (left), and the return
        $\hat{J}(\theta_e)$ with respect to the MI $\hat{I}(\theta_e)$
        (right). The maximal expected return is given by the dotted line.}
    \vspace{-0.5em}
    \label{fig:tmaze_100}
\end{figure}

In \autoref{fig:tmaze_100}, we can see that all previous observations also hold
for a T-Maze of length $L = 100$. On the left, we can see that the lower the
MI, the lower the return of the policy. For this length, in addition to the
LSTM, the GRU struggles to achieve the maximal return, which is reflected in
the evolution of its MI that increases more slowly than for the other RNNs. It
is also interesting to notice that, on average, the MGU overtake the BRC in
term of return after \num{2000} episodes, which is also the case for the MI.
Here, the linear correlation coefficient between the MI and the return is
\pearltm{} and the rank correlation coefficient is \spearltm{}. Once again, we
observe that a minimum amount of MI between the hidden states and the belief is
required for the policy to be optimal. Here, at least \num{5.0} bits of MI is
necessary.

\begin{figure}[!ht]
    \centering
    \begin{minipage}[t]{0.49\textwidth}
        \includegraphics[width=\textwidth]
            {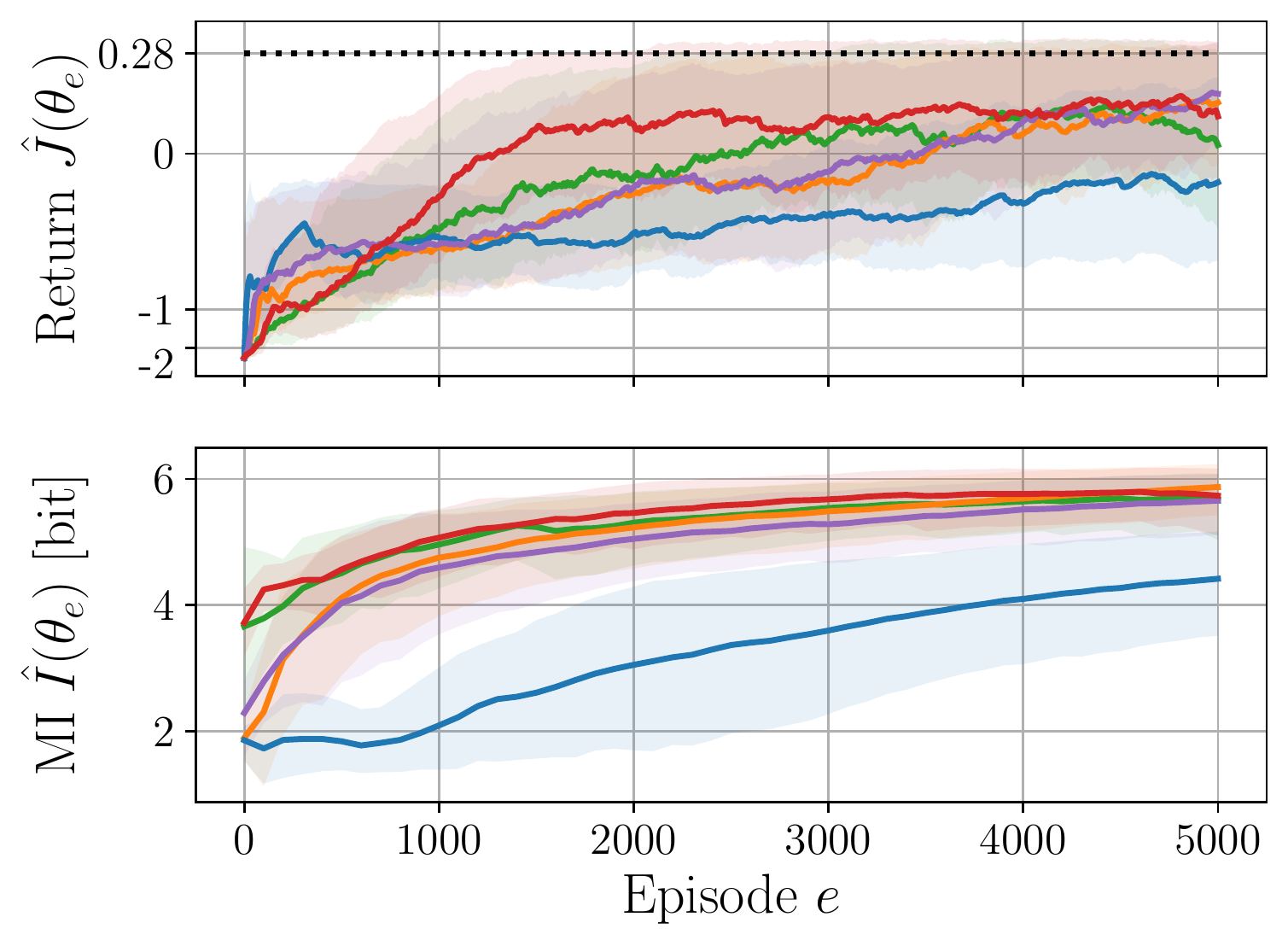}
        \label{fig:stmaze_50_learning}
    \end{minipage}
    \begin{minipage}[t]{0.49\textwidth}
        \includegraphics[width=\textwidth]
            {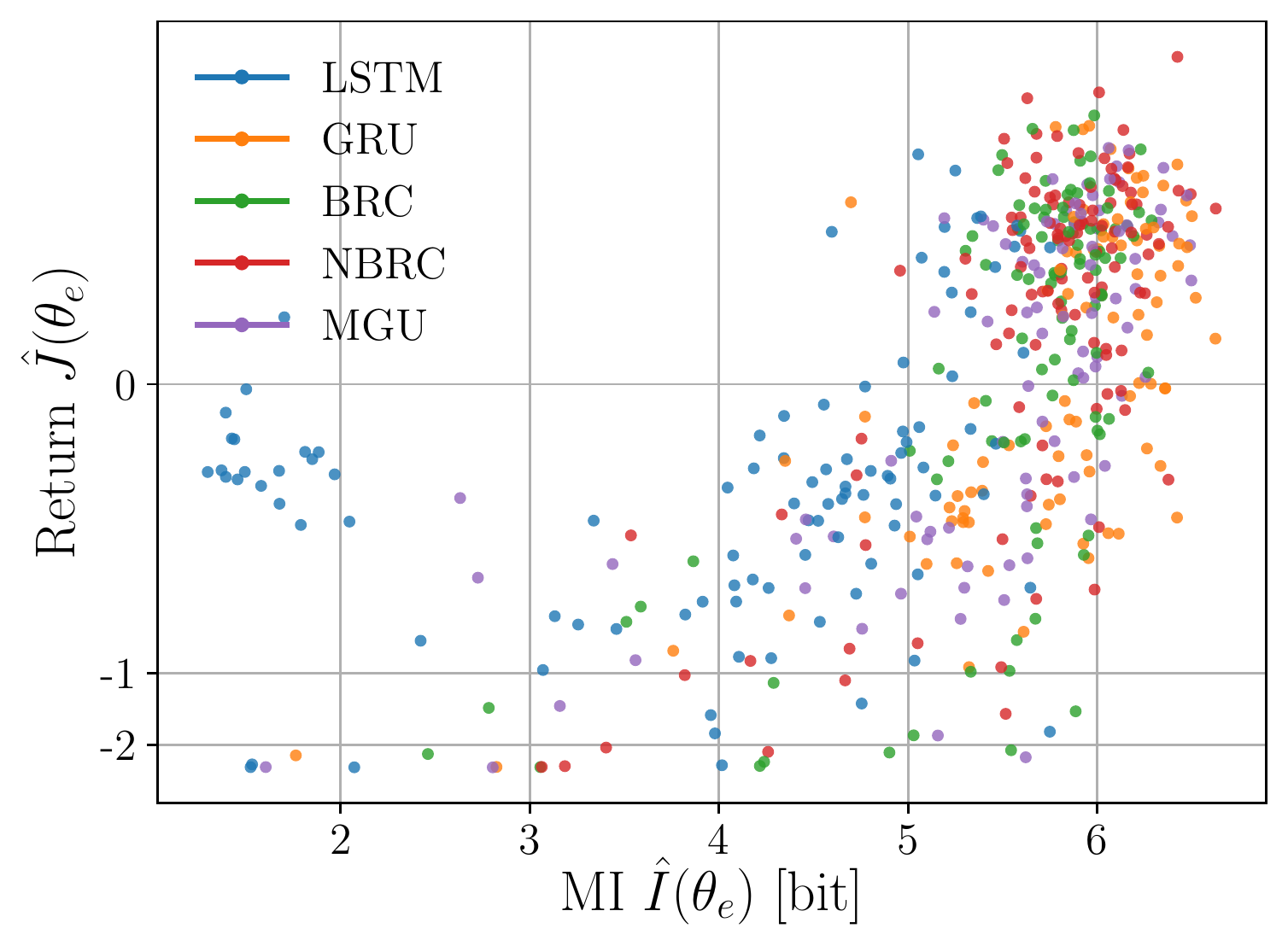}
        \label{fig:stmaze_50_corr}
    \end{minipage}
    \vspace{-1.5em}
    \caption{Stochastic T-Maze ($L = 50$, $\lambda = 0.3$).
        Evolution of the return $\hat{J}(\theta_e)$ and the MI
        $\hat{I}(\theta_e)$  after $e$ episodes (left), and the return
        $\hat{J}(\theta_e)$ with respect to the MI $\hat{I}(\theta_e)$
        (right). The maximal expected return is given by the dotted line.}
    \vspace{-0.5em}
    \label{fig:stmaze_50}
\end{figure}

In \autoref{fig:stmaze_50}, the results are shown for the Stochastic T-Maze
with $L = 50$ and $\lambda = 0.3$. On the contrary to the Deterministic T-Maze,
where the belief is a Dirac distribution over the states, there is uncertainty
on the true state in this environment. We can nevertheless observe that
previous observations hold for this environment too. The MI and the expected
return are indeed both increasing throughout the training process, and the best
performing RNNs, such as the BRC and nBRC, have a MI that increases faster and
stays higher, while the LSTM struggles to reach both a high return and a high
MI. Here, the linear correlation coefficient between the MI and the return is
\pearstm{} and the rank correlation coefficient is \spearstm{}. It can also be
noticed on the right that the best performing policies have a MI of at least
\num{4.5} bits in practice.

In the Deterministic T-Maze, it can be observed that the estimated lower bounds
$I_\phi(\h, b)$ on the MI that are obtained by the MINE estimator are tight.
Indeed, in this environment, the hidden state and belief are discrete random
variables and their mutual information is thus upper bounded by the entropy of
the belief. Moreover, the belief is a Dirac distribution that gives the actual
state with probability one. Under the optimal policy, each state is visited
with equal probability, such that the entropy of the belief is given by
$\log_2(102) = 6.6724$ for the Deterministic T-Maze of length $L = 50$, where
\num{102} is the number of non terminal states. As can be seen in
\autoref{fig:tmaze_50}, the optimal policies reach an estimated MI around $6.5$
at maximum, which nearly equals the upper bound. The same results is obtained
for the Deterministic T-Maze of length $L = 100$, where the entropy of the
belief is given by $\log_2(202) = 7.658$ and the optimal policies reach an
estimated MI around $7.0$ at maximum, as can be seen in
\autoref{fig:tmaze_100}. We expect this result to generalise to other
environments even if this would be difficult to verify in practice for random
variables with large or continuous spaces.

\subsection{Mountain Hike and Varying Mountain Hike} \label{subsec:hike}

\begin{wrapfigure}{r}{0.4\textwidth}
    \centering
    \vspace{-1.5em}
    \includegraphics[width=0.35\textwidth]{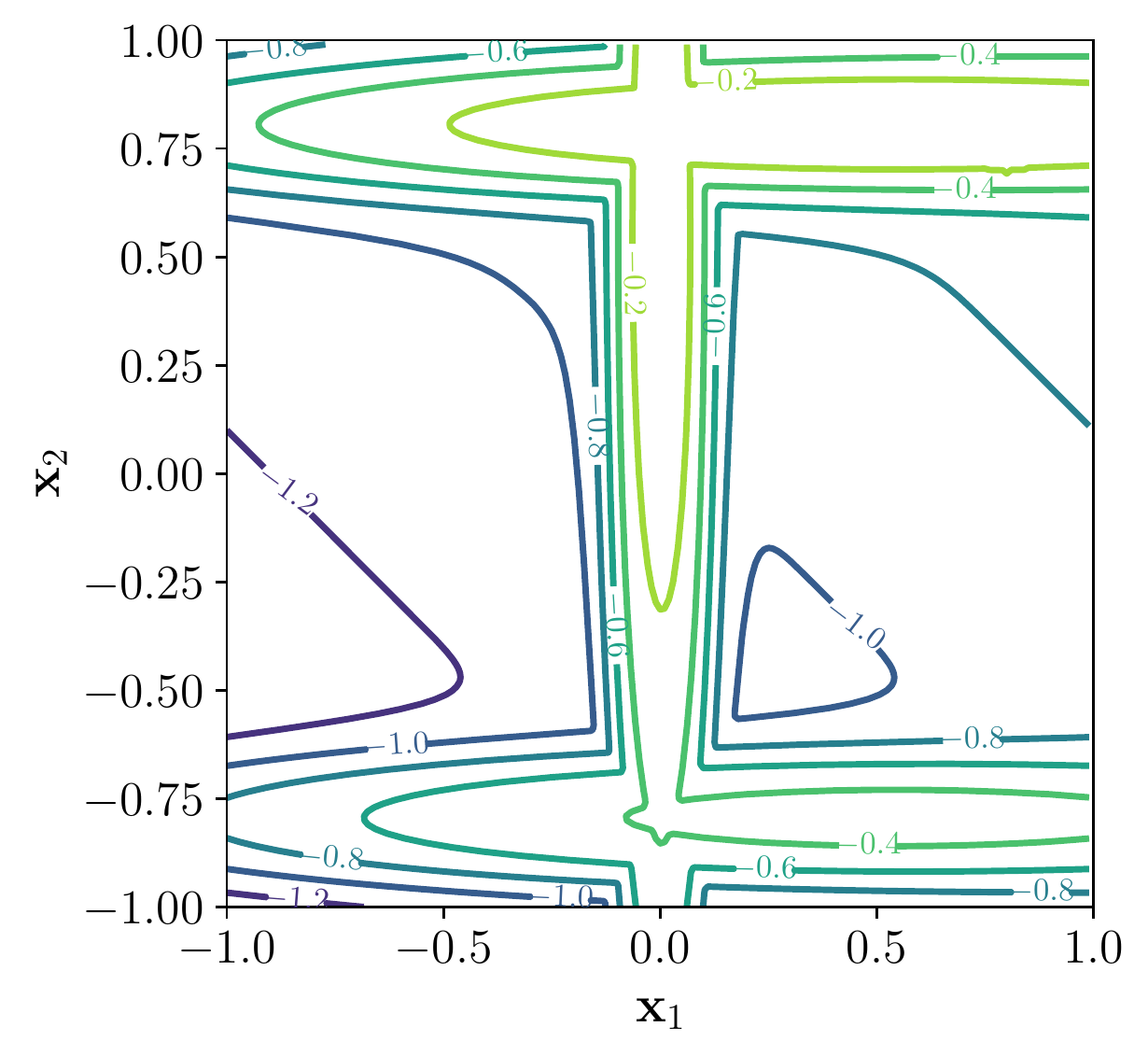}
    \vspace{-1em}
    \caption{Mountain hike altitude function.}
    \vspace{-1em}
    \label{fig:hike_reward_2d}
\end{wrapfigure}

The Mountain Hike environment is a POMDP modelling an agent walking through a
mountainous terrain. The agent has a position on a two-dimensional map and can
take actions to move in four directions relative to its initial orientation:
Forward, Backward, Right and Left. First, we consider that its initial
orientation is always North. Taking an action results in a noisy translation in
the corresponding direction. The translation noise is Gaussian with a standard
deviation of $\sigma_T = 0.05$. The only observation available is a noisy
measure of its relative altitude to the mountain top, that is always negative.
The observation noise is Gaussian with a standard deviation of $\sigma_O =
0.1$. The reward is also given by this relative altitude, such that the goal of
this POMDP is to to obtain the highest possible cumulative altitude. Around the
mountain top, the states are terminal. The optimal policy thus consists of
going as fast as possible towards those terminal states while staying on the
crests in order to get less negative rewards than in the valleys. This
environment is represented in \autoref{fig:hike_reward_2d}. This POMDP is
inspired by the Mountain Hike environment described in \citep{igl2018deep}. The
discount factor is $\gamma = 0.99$. We also consider the Varying Mountain Hike
in the experiments, a more difficult version of the Mountain Hike where the
agent randomly faces one of the four cardinal directions (\ie, North, West,
South, East) depending on the initial state. The agent does not observe its
orientation. As a consequence, the agent needs to maintain a belief about its
orientation given the observations in order to act optimally. This POMDP is
formally defined in \autoref{app:hike}.

\begin{figure}[!ht]
    \centering
    \begin{minipage}[t]{0.49\textwidth}
        \includegraphics[width=\textwidth]
            {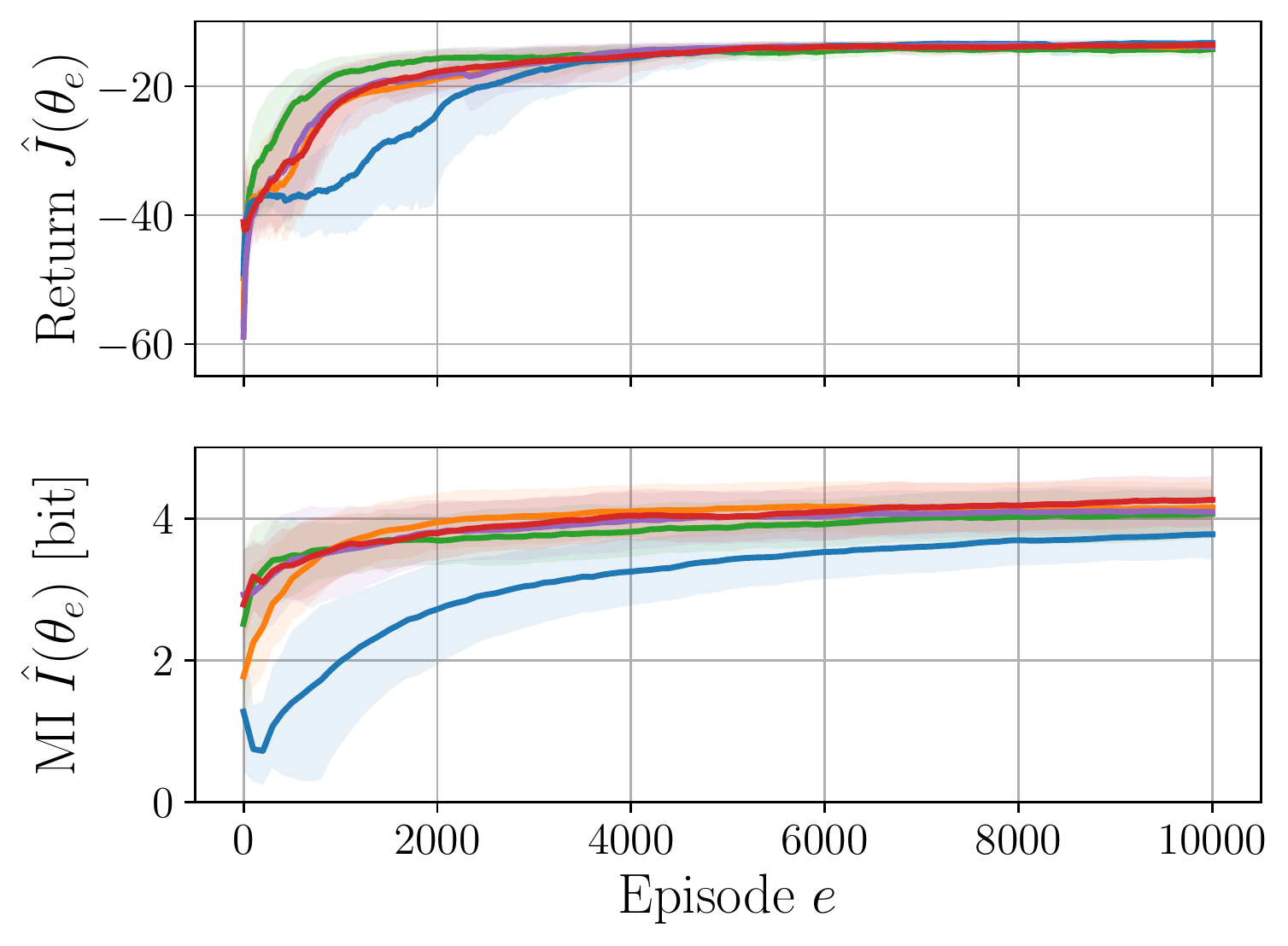}
        \label{fig:hike_learning}
    \end{minipage}
    \begin{minipage}[t]{0.49\textwidth}
        \includegraphics[width=\textwidth]
            {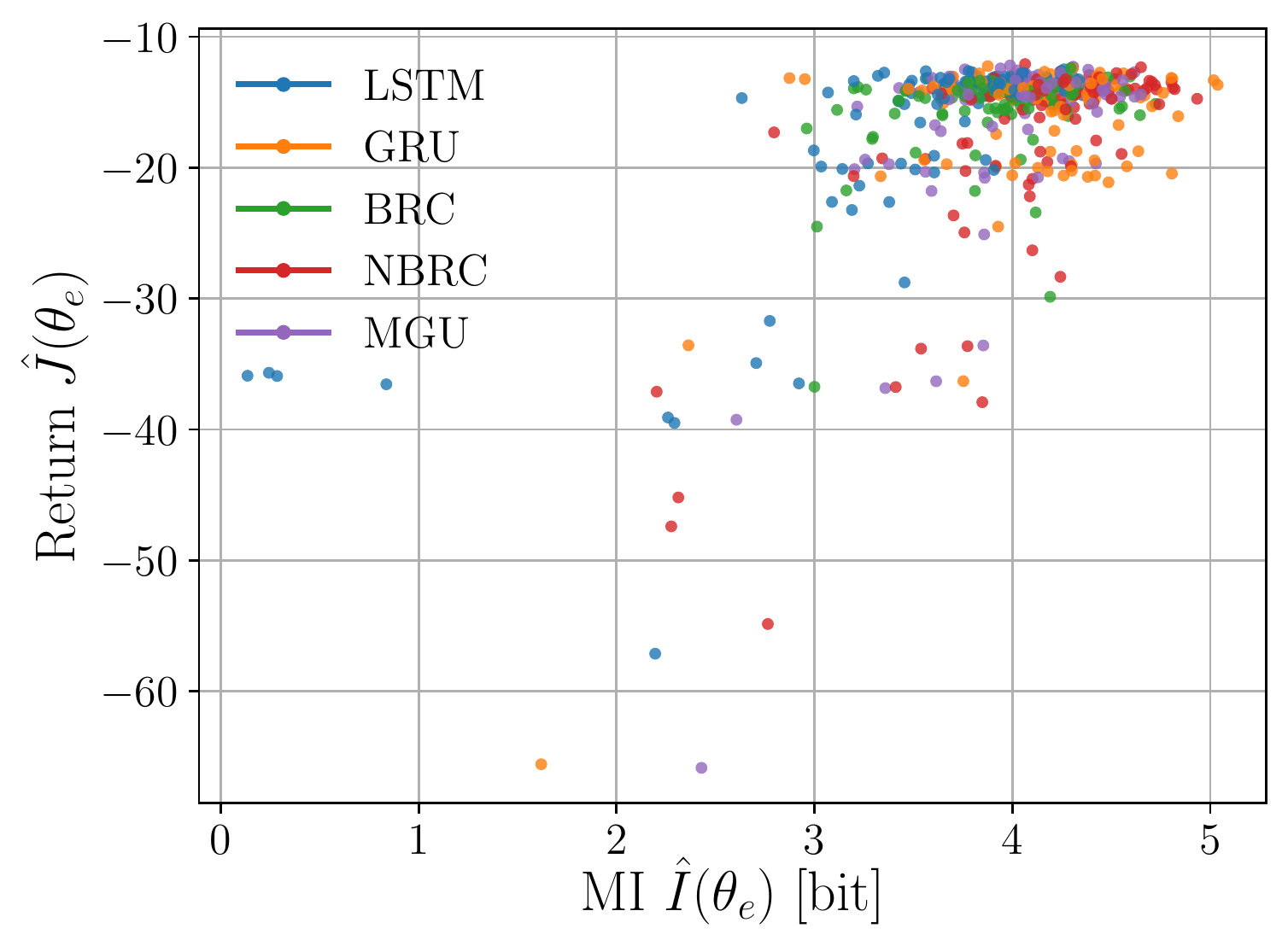}
        \label{fig:hike_corr}
    \end{minipage}
    \vspace{-1.5em}
    \caption{Mountain Hike.
        Evolution of the return $\hat{J}(\theta_e)$ and the MI
        $\hat{I}(\theta_e)$  after $e$ episodes (left), and the return
        $\hat{J}(\theta_e)$ with respect to the MI $\hat{I}(\theta_e)$
        (right).}
    \vspace{-0.5em}
    \label{fig:hike}
\end{figure}

\autoref{fig:hike} shows on the left the expected return and the MI during
training for the Mountain Hike environment. It is clear that the DRQN algorithm
promotes a high MI between the belief and the hidden states of the RNN, even in
continuous-state environments. It can also be seen that the evolution of the MI
and the evolution of the return are strongly linked throughout the training
process, for all RNNs. We can also see on the right in \autoref{fig:hike} that
the correlation between MI and performances appears clearly for each RNN. For
all RNNs, the linear correlation coefficient is \pearmh{} and the rank
correlation coefficient is \spearmh{}. In particular, we see that the best
policies, with a return around \num{-20}, are clearly separated from the others
and have a significantly higher MI on average.

\begin{figure}[!ht]
    \centering
    \begin{minipage}[t]{0.49\textwidth}
        \includegraphics[width=\textwidth]
            {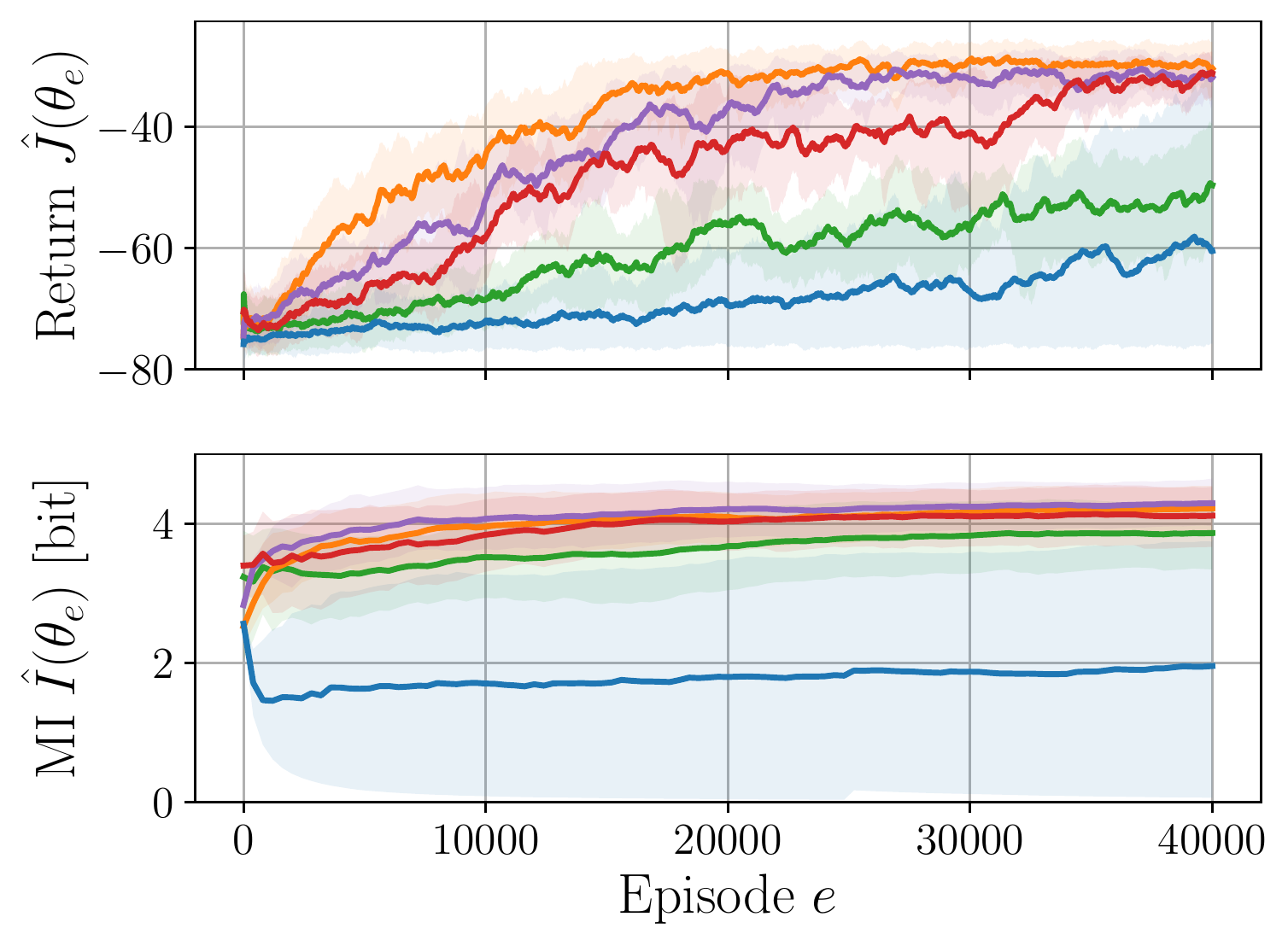}
        \label{fig:vhike_learning}
    \end{minipage}
    \begin{minipage}[t]{0.49\textwidth}
        \includegraphics[width=\textwidth]
            {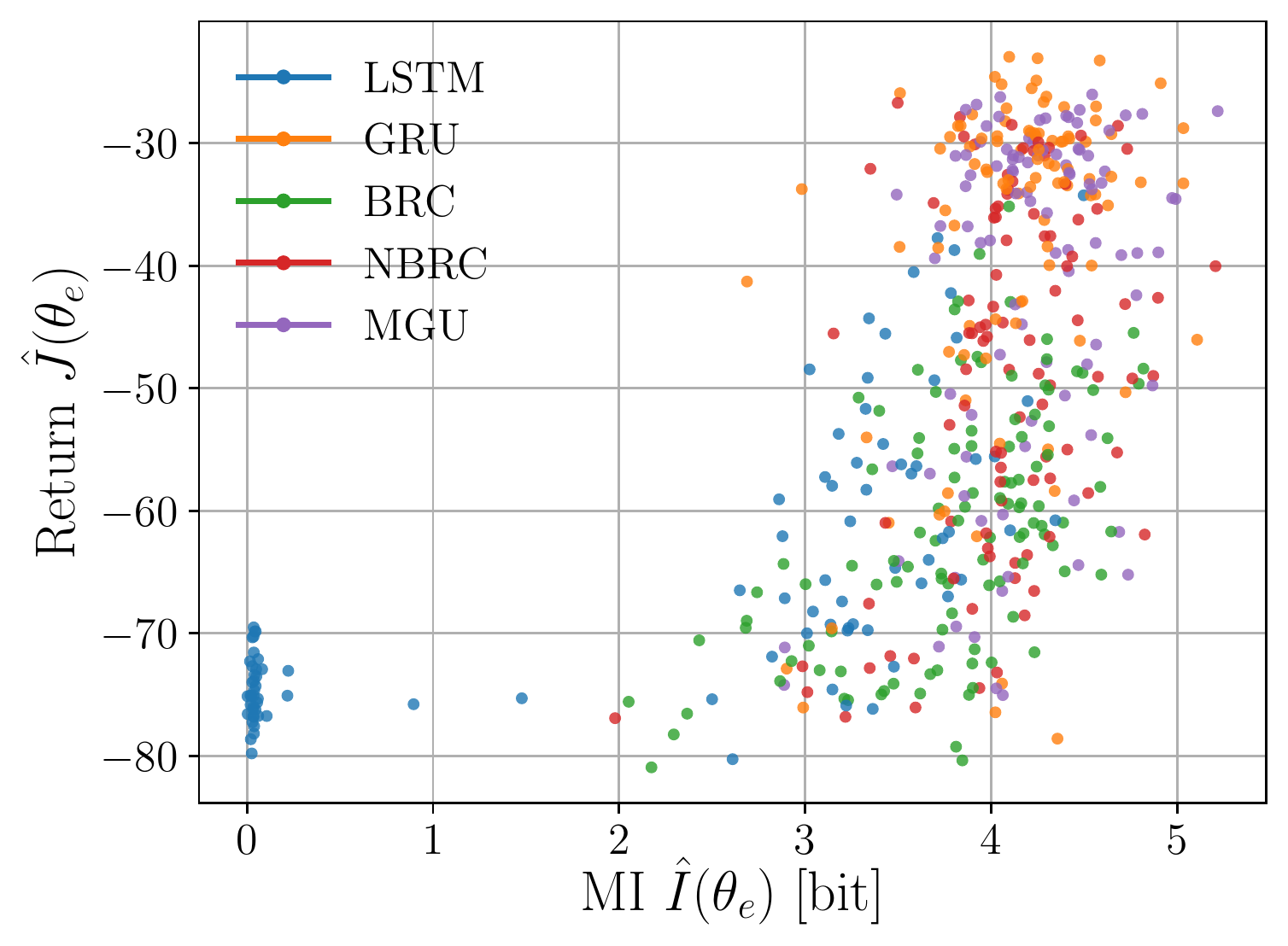}
        \label{fig:vhike_corr}
    \end{minipage}
    \vspace{-1.5em}
    \caption{Varying Mountain Hike.
        Evolution of the return $\hat{J}(\theta_e)$ and the MI
        $\hat{I}(\theta_e)$  after $e$ episodes (left), and the return
        $\hat{J}(\theta_e)$ with respect to the MI $\hat{I}(\theta_e)$
        (right).}
    \vspace{-0.5em}
    \label{fig:vhike}
\end{figure}

In \autoref{fig:vhike}, we can see the evolution and the correlation between
the return and the MI for the Varying Mountain Hike environment. The
correlation is even clearer than for the other environments. This may be due to
the fact that differences in term of performances are more pronounced than for
the other experiments. Again, the worse RNNs such as the LSTM and the BRC have
a significantly lower MI compared to the other cells. In addition, the
performances of any RNN is strongly correlated to their ability to reproduce
the belief filter, as can be seen on the right, with a sharp increase in
empirical return as the MI increases from \num{2.5} to \num{4.5} bits. More
precisely, the linear correlation coefficient between the MI and the return is
\pearvmh{} and the rank correlation coefficient is \spearvmh{}. This increase
occurs throughout the training process, as can be seen on the left.

\subsection{Belief of variables irrelevant for the optimal control}
\label{subsec:irrelevant}

Despite the belief being a sufficient statistic from the history in order to
act optimally, it may be that only the belief of some state variables is
necessary for optimal control. In this subsection, we show that approximating
the \cqf{} with an RNN will only tend to reconstruct the necessary part,
naturally filtering away the belief of irrelevant state variables.

In order to study this phenomenon, we construct a new POMDP $P'$ from a POMDP
$P$ by adding new state variables, independent of the original ones, and
irrelevant for optimal control. More precisely, we add $d$ irrelevant state
variables $\s_t^I$ that follows a Gaussian random walk. In addition, the agent
acting in the POMDP $P'$ obtains partial observations $\o_t^I$ of the new state
variables through an unbiased Gaussian observation model. Formally, the new
states and observations are distributed according to
\begin{align}
    p(\s_0^I) &=
        \phi(\s_0^I ; \mathbf{0}, \1) \\
    p(\s_{t+1}^I \mid \s_t^I) &=
        \phi(\s_{t+1}^I ; \s_t^I, \1), \;
        \forall t \in \N_0, \\
    p(\o_t^I \mid \s_t^I) &=
        \phi(\o_t^I ; s_t^I, \1), \;
        \forall t \in \N_0,
\end{align}
where $\phi(\x; \mu, \Sigma)$ is the probability density function of a
multivariate random variable of mean $\mu \in \R^d$ and covariance matrix
$\Sigma \in \R^{d \times d}$, evaluated at $\x \in \R^d$, and $\1$ is the
identity matrix.

\begin{figure}[!ht]
    \centering
    \begin{subfigure}[t]{0.49\textwidth}
        \includegraphics[width=\textwidth]
            {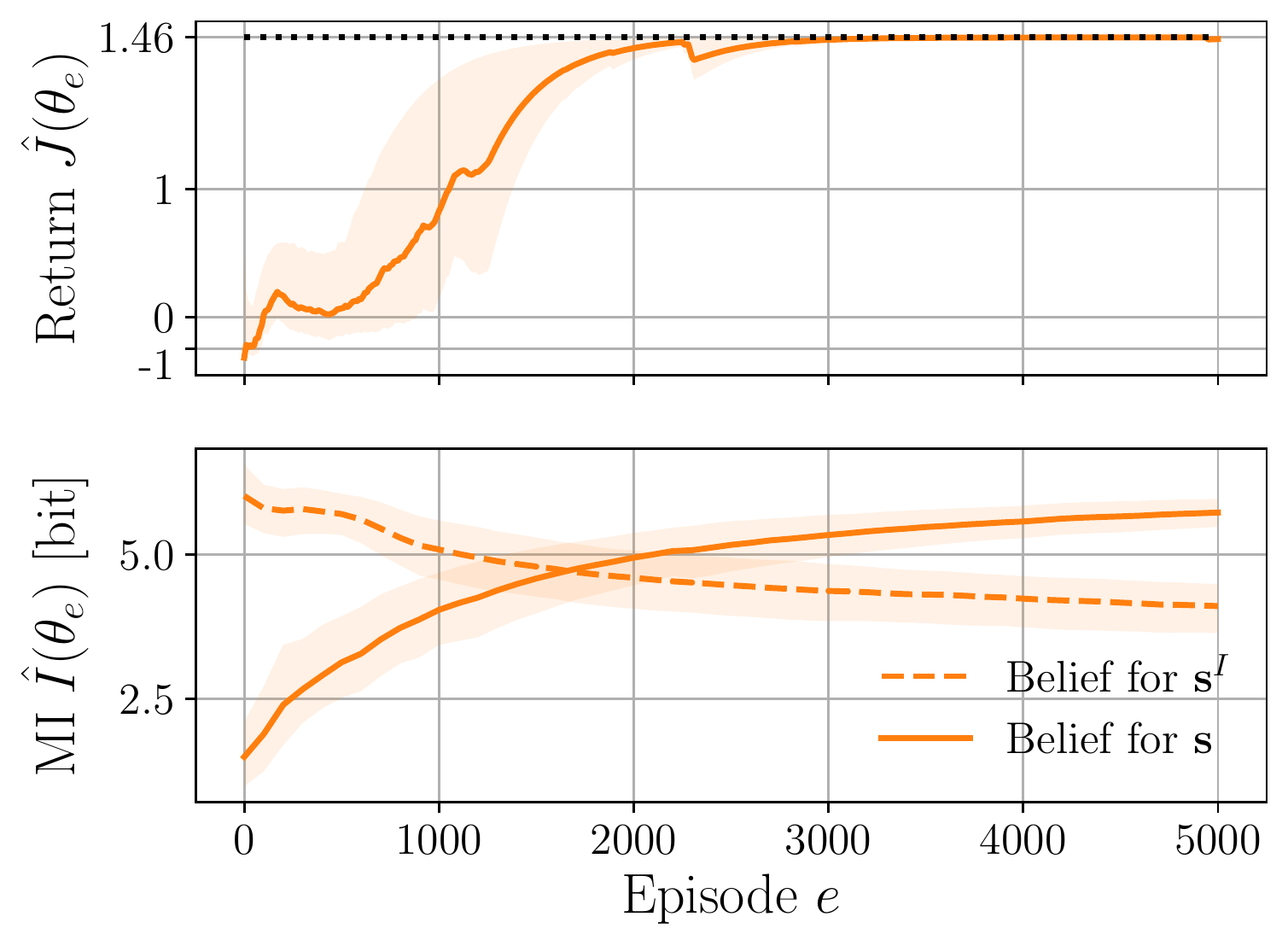}
        \vspace{-1.7em}
        \caption{$d = 1$}
        \vspace{1em}
        \label{fig:tmaze_irr_1_gru}
    \end{subfigure}
    \begin{subfigure}[t]{0.49\textwidth}
        \includegraphics[width=\textwidth]
            {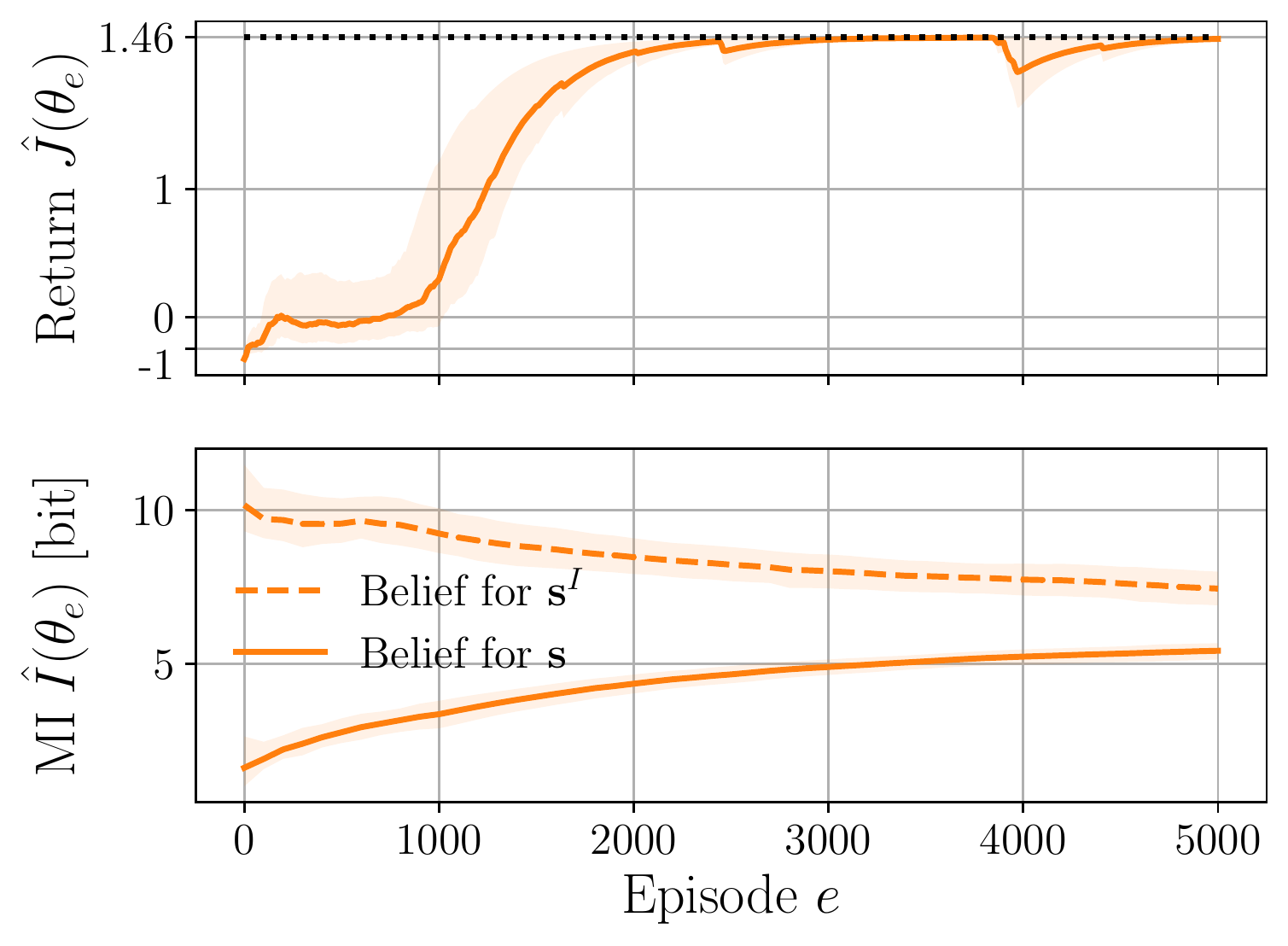}
        \vspace{-1.7em}
        \caption{$d = 4$}
        \vspace{1em}
        \label{fig:tmaze_irr_4_gru}
    \end{subfigure}
    \vspace{-1em}
    \caption{Deterministic T-Maze ($L = 50$) with $d$ irrelevant state
        variables.
        Evolution of the return $\hat{J}(\theta_e)$ and the MI
        $\hat{I}(\theta_e)$ for the belief of the irrelevant and relevant state
        variables after $e$ episodes, for the GRU cell. The maximal
        expected return is given by the dotted line.} \vspace{-0.5em}
    \label{fig:tmaze_irr_gru}
\end{figure}

\autoref{fig:tmaze_irr_gru} shows the return and the MI measured for the GRU on
the T-Maze environment with $L = 50$. It can be observed, as for the classic
T-Maze environment, that the MI between the hidden states and the belief of
state variables that are relevant to optimal control increases with the return.
In addition, the MI with the belief of irrelevant variables decreases during
training. It can also be seen that, for $d = 4$, the MI with the belief of
irrelevant variables remains higher than the MI with the belief of relevant
variables, due to the high entropy of this irrelevant process. Finally, it is
interesting to note that the MI continues to increase (\resp{} decrease) with
the belief of relevant (\resp{} irrelevant) variables long after the optimal
policy is reached, suggesting that the hidden states of the RNN still change
substantially. Similar results are obtained for the other cells (see
\autoref{subsec:irrelevant_additional}).

\begin{figure}[!ht]
    \centering
    \begin{subfigure}[t]{0.49\textwidth}
        \includegraphics[width=\textwidth]
            {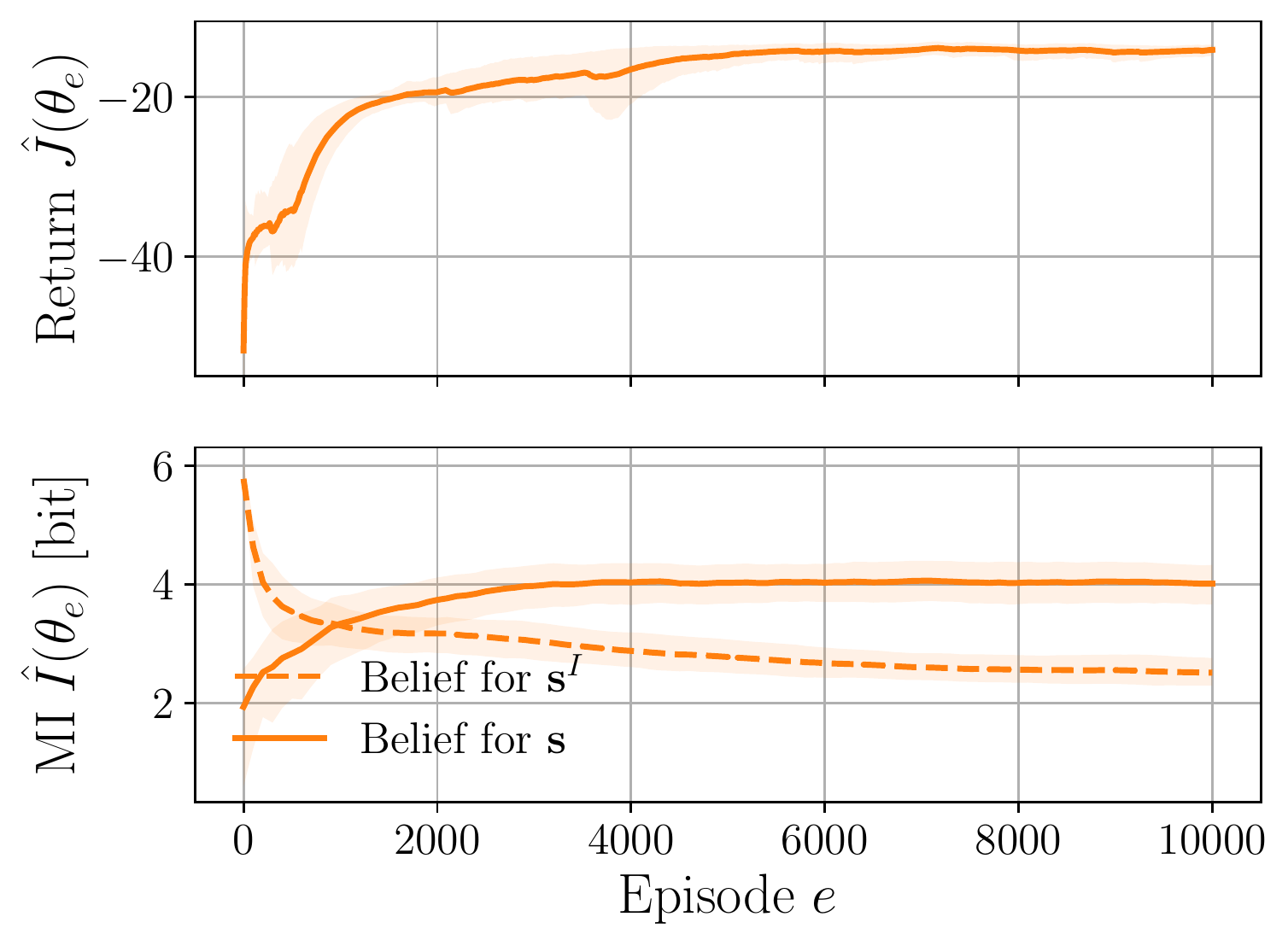}
        \vspace{-1.7em}
        \caption{$d = 1$}
        \vspace{1em}
        \label{fig:hike_irr_1_gru}
    \end{subfigure}
    \begin{subfigure}[t]{0.49\textwidth}
        \includegraphics[width=\textwidth]
            {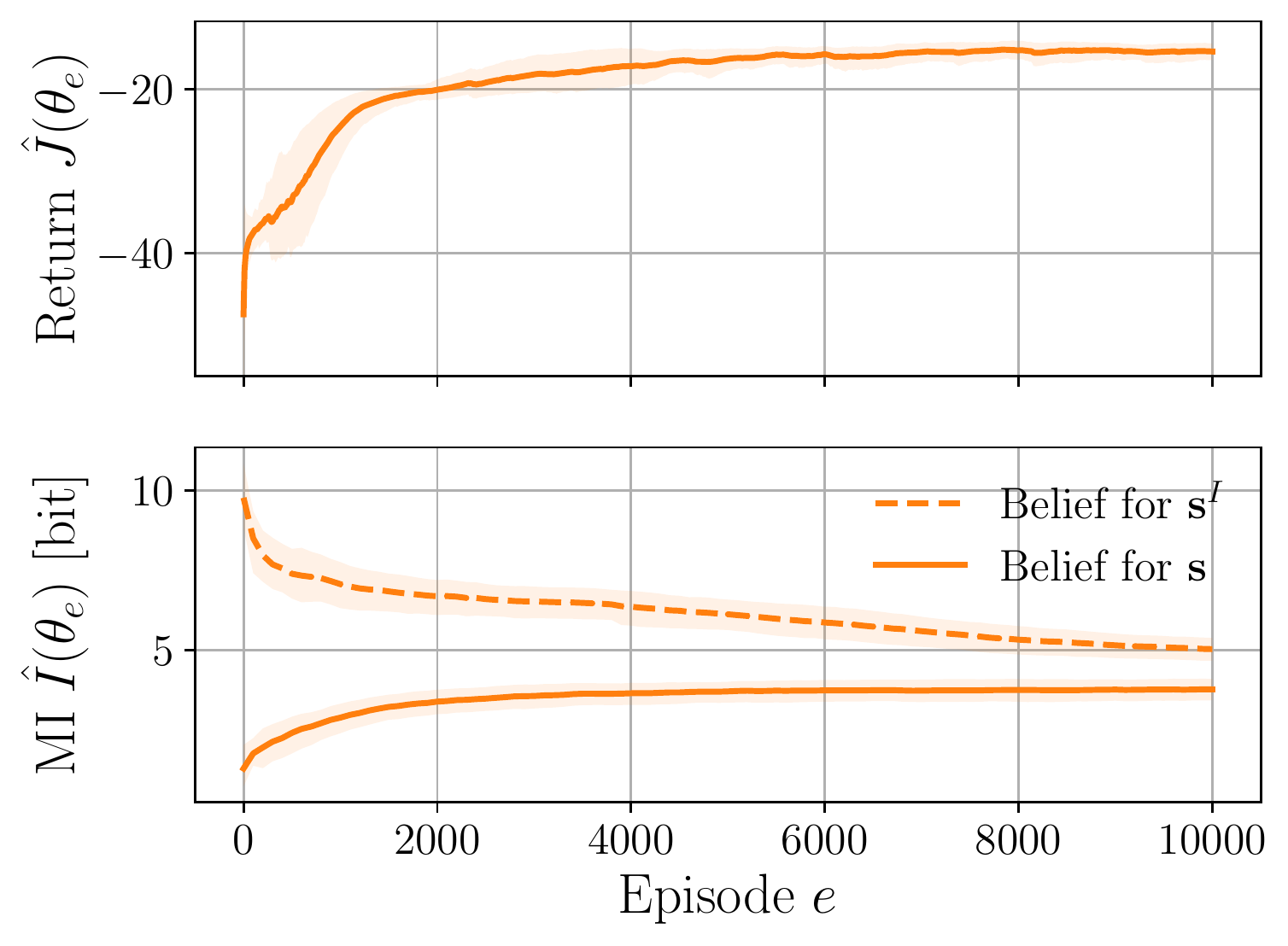}
        \vspace{-1.7em}
        \caption{$d = 4$}
        \vspace{1em}
        \label{fig:hike_irr_4_gru}
    \end{subfigure}
    \vspace{-1em}
    \caption{Mountain Hike with with $d$ irrelevant state variables.
        Evolution of the return $\hat{J}(\theta_e)$ and the MI
        $\hat{I}(\theta_e)$ for the belief of the irrelevant and relevant state
        variables after $e$ episodes, for the GRU cell.}
    \vspace{-0.5em}
    \label{fig:hike_irr_gru}
\end{figure}

\autoref{fig:hike_irr_gru} shows the return and the MI measured for the GRU on
the Mountain Hike environment. The same conclusions as for the T-Maze can be
drawn, with a clear increase of the MI for the relevant variables throughout
the training process, and a clear decrease of the MI for the irrelevant
variables. In addition, it can be seen that the optimal policy is reached later
when there are more irrelevant variables. It is also clear that adding more
irrelevant variables increases the entropy of the irrelevant process, which
leads to a higher MI between the hidden states and the irrelevant state
variables. Similar results are obtained for the other cells (see
\autoref{subsec:irrelevant_additional}).

\subsection{Discussion} \label{subsec:discussion}

As shown in the experiments, under the distribution induced by a recurrent
policy trained using recurrent Q-learning, its hidden state provide a high
amount of information about the belief of relevant state variables, at any time
step. The hidden state of the RNN is thus a statistic from the history that
encodes information about the belief. In addition, at any time step, the
network performs an update of this statistic, based on the actions and
observations that are observed. The RNN thus implements a filter that provides
a statistic encoding the belief.

However, it was only shown that the RNN produces such a statistic under the
distribution of histories induced by the learned policy. For the sake of
robustness of the policy to perturbations of histories, we might want this
statistic to also provide information about the belief under other distribution
of histories. In \autoref{app:generalisation}, we propose an experimental
protocol to study the generalisation of the learned statistics. The results
show that the MI between the hidden states and the beliefs also increases
throughout the training process, under distributions induced by various
$\varepsilon$-greedy policies, even the fully random policy. We impute those
results to the following reasons. First, the DRQN algorithm approximates the
\cqf{}, which generally requires a richer statistic from the history than the
optimal policy. Second, the DRQN algorithm makes use of exploration, which
allows the RNN to learn from histories that are diverse. However, we still
observe that the higher the noise, the lower the MI. From these results, we
conclude that the statistic that is learned by the network generalises
reasonably well to other distributions of histories.

\section{Conclusions} \label{sec:conclusions}

In this work, we have shown empirically for several POMDPs that RNNs
approximating the Q-function with a recurrent Q-learning algorithm
\citep{hausknecht2015deep, zhu2017improving} produces a statistic in their
hidden states that provide a high amount of information about the belief of
state variables that are relevant for optimal control. More precisely, we have
shown that the MI between the hidden states of the RNN and the belief of states
variables that are relevant for optimal control was increasing throughout the
training process. In addition, we have shown that the ability of a recurrent
architecture to reproduce, through a high MI, the belief filter conditions the
performance of its policy. Finally, we showed that the MI between the hidden
states and the beliefs of state variables that are irrelevant for optimal
control decreases through the training process, suggesting that RNNs only focus
on the relevant part of the belief.

This work also opens up several paths for future work. First, this work
suggests that enforcing a high MI between the hidden states and the beliefs
leads to an increase in the performances of the algorithm and in the return of
the resulting policy. While other works have focused on an explicit
representation of the belief in the hidden states \citep{karkus2017qmdp,
igl2018deep}, which required to design specific recurrent architectures, we
propose to implicitly embed the belief in the hidden state of any recurrent
architecture by maximising their MI. When the belief or state particles are
available, this can be done by adding an auxiliary loss such that the RNN also
maximises the MI. In practice, this can be implemented by backpropagating the
MINE loss beyond the MINE architecture through the unrolled RNN architecture,
such that the hidden states are optimized to get a higher MI with the beliefs.

Moreover, this work could be extended to algorithms that approximate other
functions of the histories than the \cqf{}. Notably, this study could be
extended to the hidden states of a recurrent policy learned by policy-gradient
algorithms or to the hidden states of the actor and the critic in actor-critic
methods. We may nevertheless expect to find similar results since the value
function of a policy tends towards the optimal value function when the policy
tends towards the optimal policy.

\section*{Acknowledgments}

Gaspard Lambrechts gratefully acknowledges the financial support of the
\emph{Wallonia-Brussels Federation} for his FRIA grant and the financial
support of the \emph{Walloon Region} for Grant No.\@ 2010235 – ARIAC by DW4AI.
Adrien Bolland gratefully acknowledges the financial support of the
\emph{Wallonia-Brussels Federation} for his FNRS grant. Computational resources
have been provided by the \emph{Consortium des Équipements de Calcul Intensif}
(CÉCI), funded by the \emph{Fonds de la Recherche Scientifique de Belgique}
(F.R.S.-FNRS) under Grant No. 2502011 and by the Walloon Region.

\bibliography{bib/bibliography}

\begin{thebibliography}{27}
\providecommand{\natexlab}[1]{#1}
\providecommand{\url}[1]{\texttt{#1}}
\expandafter\ifx\csname urlstyle\endcsname\relax
  \providecommand{\doi}[1]{doi: #1}\else
  \providecommand{\doi}{doi: \begingroup \urlstyle{rm}\Url}\fi

\bibitem[Bakker(2001)]{bakker2001reinforcement}
Bram Bakker.
\newblock Reinforcement learning with long short-term memory.
\newblock \emph{Advances in neural information processing systems}, 14, 2001.

\bibitem[Belghazi et~al.(2018)Belghazi, Baratin, Rajeshwar, Ozair, Bengio,
  Courville, and Hjelm]{belghazi2018mutual}
Mohamed~Ishmael Belghazi, Aristide Baratin, Sai Rajeshwar, Sherjil Ozair,
  Yoshua Bengio, Aaron Courville, and Devon Hjelm.
\newblock Mutual information neural estimation.
\newblock In \emph{International conference on machine learning}, pp.\
  531--540, 2018.

\bibitem[Bertsekas(2012)]{bertsekas2012dynamic}
Dimitri Bertsekas.
\newblock \emph{Dynamic programming and optimal control: Volume I}, volume~1.
\newblock Athena scientific, 2012.

\bibitem[Chung et~al.(2014)Chung, Gulcehre, Cho, and
  Bengio]{chung2014empirical}
Junyoung Chung, Caglar Gulcehre, KyungHyun Cho, and Yoshua Bengio.
\newblock Empirical evaluation of gated recurrent neural networks on sequence
  modeling.
\newblock \emph{arXiv preprint arXiv:1412.3555}, 2014.

\bibitem[Donsker \& Varadhan(1975)Donsker and Varadhan]{donsker1975asymptotic}
Monroe~D Donsker and SR~Srinivasa Varadhan.
\newblock Asymptotic evaluation of certain {M}arkov process expectations for
  large time, {I}.
\newblock \emph{Communications on Pure and Applied Mathematics}, 28\penalty0
  (1):\penalty0 1--47, 1975.

\bibitem[Haarnoja et~al.(2018)Haarnoja, Zhou, Abbeel, and
  Levine]{haarnoja2018soft}
Tuomas Haarnoja, Aurick Zhou, Pieter Abbeel, and Sergey Levine.
\newblock Soft actor-critic: Off-policy maximum entropy deep reinforcement
  learning with a stochastic actor.
\newblock In \emph{International conference on machine learning}, pp.\
  1861--1870, 2018.

\bibitem[Hausknecht \& Stone(2015)Hausknecht and Stone]{hausknecht2015deep}
Matthew Hausknecht and Peter Stone.
\newblock Deep recurrent {Q}-learning for partially observable {MDP}s.
\newblock In \emph{Association for the advancement of artificial intelligence
  fall symposium series}, 2015.

\bibitem[Heess et~al.(2015)Heess, Hunt, Lillicrap, and Silver]{heess2015memory}
Nicolas Heess, Jonathan~J Hunt, Timothy~P Lillicrap, and David Silver.
\newblock Memory-based control with recurrent neural networks.
\newblock \emph{arXiv preprint arXiv:1512.04455}, 2015.

\bibitem[Hessel et~al.(2018)Hessel, Modayil, Van~Hasselt, Schaul, Ostrovski,
  Dabney, Horgan, Piot, Azar, and Silver]{hessel2018rainbow}
Matteo Hessel, Joseph Modayil, Hado Van~Hasselt, Tom Schaul, Georg Ostrovski,
  Will Dabney, Dan Horgan, Bilal Piot, Mohammad Azar, and David Silver.
\newblock Rainbow: Combining improvements in deep reinforcement learning.
\newblock In \emph{Thirty-second association for the advancement of artificial
  intelligence conference on artificial intelligence}, 2018.

\bibitem[Hochreiter \& Schmidhuber(1997)Hochreiter and
  Schmidhuber]{hochreiter1997long}
Sepp Hochreiter and J{\"u}rgen Schmidhuber.
\newblock Long short-term memory.
\newblock \emph{Neural computation}, 9\penalty0 (8):\penalty0 1735--1780, 1997.

\bibitem[Igl et~al.(2018)Igl, Zintgraf, Le, Wood, and Whiteson]{igl2018deep}
Maximilian Igl, Luisa Zintgraf, Tuan~Anh Le, Frank Wood, and Shimon Whiteson.
\newblock Deep variational reinforcement learning for {POMDP}s.
\newblock In \emph{International Conference on Machine Learning}, pp.\
  2117--2126, 2018.

\bibitem[Kaelbling et~al.(1998)Kaelbling, Littman, and
  Cassandra]{kaelbling1998planning}
Leslie~Pack Kaelbling, Michael~L Littman, and Anthony~R Cassandra.
\newblock Planning and acting in partially observable stochastic domains.
\newblock \emph{Artificial intelligence}, 101\penalty0 (1-2):\penalty0 99--134,
  1998.

\bibitem[Karkus et~al.(2017)Karkus, Hsu, and Lee]{karkus2017qmdp}
Peter Karkus, David Hsu, and Wee~Sun Lee.
\newblock {QMDP}-net: Deep learning for planning under partial observability.
\newblock \emph{Advances in neural information processing systems}, 30, 2017.

\bibitem[Kingma \& Ba(2014)Kingma and Ba]{kingma2014adam}
Diederik~P Kingma and Jimmy Ba.
\newblock Adam: A method for stochastic optimization.
\newblock \emph{arXiv preprint arXiv:1412.6980}, 2014.

\bibitem[Kraskov et~al.(2004)Kraskov, St{\"o}gbauer, and
  Grassberger]{kraskov2004estimating}
Alexander Kraskov, Harald St{\"o}gbauer, and Peter Grassberger.
\newblock Estimating mutual information.
\newblock \emph{Physical review E}, 69\penalty0 (6):\penalty0 066138, 2004.

\bibitem[Lillicrap et~al.(2015)Lillicrap, Hunt, Pritzel, Heess, Erez, Tassa,
  Silver, and Wierstra]{lillicrap2015continuous}
Timothy~P Lillicrap, Jonathan~J Hunt, Alexander Pritzel, Nicolas Heess, Tom
  Erez, Yuval Tassa, David Silver, and Daan Wierstra.
\newblock Continuous control with deep reinforcement learning.
\newblock \emph{arXiv preprint arXiv:1509.02971}, 2015.

\bibitem[Mikulik et~al.(2020)Mikulik, Del{\'e}tang, McGrath, Genewein, Martic,
  Legg, and Ortega]{mikulik2020meta}
Vladimir Mikulik, Gr{\'e}goire Del{\'e}tang, Tom McGrath, Tim Genewein, Miljan
  Martic, Shane Legg, and Pedro Ortega.
\newblock Meta-trained agents implement {B}ayes-optimal agents.
\newblock \emph{Advances in neural information processing systems},
  33:\penalty0 18691--18703, 2020.

\bibitem[Mnih et~al.(2015)Mnih, Kavukcuoglu, Silver, Rusu, Veness, Bellemare,
  Graves, Riedmiller, Fidjeland, Ostrovski, et~al.]{mnih2015human}
Volodymyr Mnih, Koray Kavukcuoglu, David Silver, Andrei~A Rusu, Joel Veness,
  Marc~G Bellemare, Alex Graves, Martin Riedmiller, Andreas~K Fidjeland, Georg
  Ostrovski, et~al.
\newblock Human-level control through deep reinforcement learning.
\newblock \emph{Nature}, 518\penalty0 (7540):\penalty0 529--533, 2015.

\bibitem[Mnih et~al.(2016)Mnih, Badia, Mirza, Graves, Lillicrap, Harley,
  Silver, and Kavukcuoglu]{mnih2016asynchronous}
Volodymyr Mnih, Adria~Puigdomenech Badia, Mehdi Mirza, Alex Graves, Timothy
  Lillicrap, Tim Harley, David Silver, and Koray Kavukcuoglu.
\newblock Asynchronous methods for deep reinforcement learning.
\newblock In \emph{International conference on machine learning}, pp.\
  1928--1937, 2016.

\bibitem[Porta et~al.(2004)Porta, Spaan, and Vlassis]{porta2004value}
Josep~M. Porta, Matthijs T.~J. Spaan, and Nikos Vlassis.
\newblock Value iteration for continuous-state {POMDP}s.
\newblock \emph{Technical Report IAS-UVA-04-04}, December 2004.

\bibitem[Smallwood \& Sondik(1973)Smallwood and Sondik]{smallwood1973optimal}
Richard~D Smallwood and Edward~J Sondik.
\newblock The optimal control of partially observable markov processes over a
  finite horizon.
\newblock \emph{Operations research}, 21\penalty0 (5):\penalty0 1071--1088,
  1973.

\bibitem[Thrun(2002)]{thrun2002probabilistic}
Sebastian Thrun.
\newblock Probabilistic robotics.
\newblock \emph{Communications of the ACM}, 45\penalty0 (3):\penalty0 52--57,
  2002.

\bibitem[Vecoven et~al.(2021)Vecoven, Ernst, and Drion]{vecoven2021bio}
Nicolas Vecoven, Damien Ernst, and Guillaume Drion.
\newblock A bio-inspired bistable recurrent cell allows for long-lasting
  memory.
\newblock \emph{Plos one}, 16\penalty0 (6):\penalty0 e0252676, 2021.

\bibitem[Werbos(1990)]{werbos1990backpropagation}
Paul~J Werbos.
\newblock Backpropagation through time: what it does and how to do it.
\newblock \emph{Proceedings of the institute of electrical and electronics
  engineers}, 78\penalty0 (10):\penalty0 1550--1560, 1990.

\bibitem[Zaheer et~al.(2017)Zaheer, Kottur, Ravanbakhsh, Poczos, Salakhutdinov,
  and Smola]{zaheer2017deep}
Manzil Zaheer, Satwik Kottur, Siamak Ravanbakhsh, Barnabas Poczos, Russ~R
  Salakhutdinov, and Alexander~J Smola.
\newblock Deep sets.
\newblock \emph{Advances in neural information processing systems}, 30, 2017.

\bibitem[Zhou et~al.(2016)Zhou, Wu, Zhang, and Zhou]{zhou2016minimal}
Guo-Bing Zhou, Jianxin Wu, Chen-Lin Zhang, and Zhi-Hua Zhou.
\newblock Minimal gated unit for recurrent neural networks.
\newblock \emph{International Journal of Automation and Computing}, 13\penalty0
  (3):\penalty0 226--234, 2016.

\bibitem[Zhu et~al.(2017)Zhu, Li, Poupart, and Miao]{zhu2017improving}
Pengfei Zhu, Xin Li, Pascal Poupart, and Guanghui Miao.
\newblock On improving deep reinforcement learning for {POMDP}s.
\newblock \emph{arXiv preprint arXiv:1704.07978}, 2017.

\end{thebibliography}
\bibliographystyle{bst/tmlr}

\appendix

\newpage

\section{Environments} \label{app:environments}

In this section, the class of environments that are considered in this work are
introduced. Then, the environments are formally defined.

\subsection{Class of environments} \label{app:environments_class}

In the experiments, the class of POMDPs that are considered is restricted to
those where we can observe from $\o_t$ if a state $\s_t$ is terminal. A state
$\s \in \S$ is said to be terminal if, and only if
\begin{empheq}[left = \empheqlbrace]{align}
    T(\s' \mid \s, \a) &= \delta_{\s}(\s'),
    \; \forall \s' \in \S, \forall \a \in \A \\
    R(\s, \a, \s) &= 0, \; \forall \a \in \A
\end{empheq}
where $\delta_\s$ denotes the Dirac distribution centred in $\s \in \S$. As can
be noted, the expected cumulative reward of any policy when starting in a
terminal state is zero. As a consequence, the \cqf{} of a history for which we
observe a terminal state is also zero for any initial action. The PRQL
algorithm thus only has to learn the \cqf{} of histories that have not yet
reached a terminal state. It implies that the histories that are generated in
the POMDP can be interrupted as soon as a terminal state is observed.

\subsection{T-Maze environments} \label{app:tmaze}

The T-Maze environment is a POMDP $(\S, \A, \O, p_0, T, R, O, \gamma)$
parameterised by the maze length $L \in \N$ and the stochasticity rate $\lambda
\in [0, 1]$. The formal definition of this environment is given below.

\begin{figure}[!ht]
    \centering
    \resizebox{0.6\textwidth}{!}{\begin{tikzpicture}[scale=2]

    \pgfmathsetmacro{\L}{8}

    \node (mup) at (0.5, 5.5) {$\mathbf{m} = \text{Up}$} ;
    \node (mup) at (0.5, 1.5) {$\mathbf{m} = \text{Down}$} ;

    \foreach \s in {0, 4} {
        \draw[lightgray] (0, \s) grid (\L, \s + 1) ;
        \draw[lightgray] (\L, \s - 1) grid (\L + 1, \s + 2) ;

        \fill[blue, opacity=0.2] (0, \s) rectangle (1, \s + 1) ;
        \fill[pattern=north west lines, pattern color=blue!20]
            (\L, 6 * \s / 4 - 1) rectangle (\L + 1, 6 * \s / 4) ;
        \fill[gray, opacity=0.2] (\L, \s + 1) rectangle (\L + 1, \s + 2) ;
        \fill[gray, opacity=0.2] (\L, \s - 1) rectangle (\L + 1, \s + 0) ;

        \pgfmathsetmacro{\Lminustwo}{\L-2}
        \foreach \i in {0,...,\Lminustwo} {
            \node (c\i) at (\i + 0.5, \s + 0.5) {$\mathbf{c} = (\i, 0)$} ;
        }

        \node (cdot) at (\L - 0.5, \s + 0.5) {$\dots$} ;
        \node (cL) at (\L + 0.5, \s + 0.5) {$\mathbf{c} = (L, 0)$} ;
        \node (cL+1) at (\L + 0.5, \s + 1.5) {$\mathbf{c} = (L, 1)$} ;
        \node (cL+2) at (\L + 0.5, \s - 0.5) {$\mathbf{c} = (L, -1)$} ;
    }

\end{tikzpicture}}
    \caption{T-Maze state space. Initial states in blue, terminal states in
        grey, and treasure states hatched.}
    \label{fig:tmaze}
\end{figure}
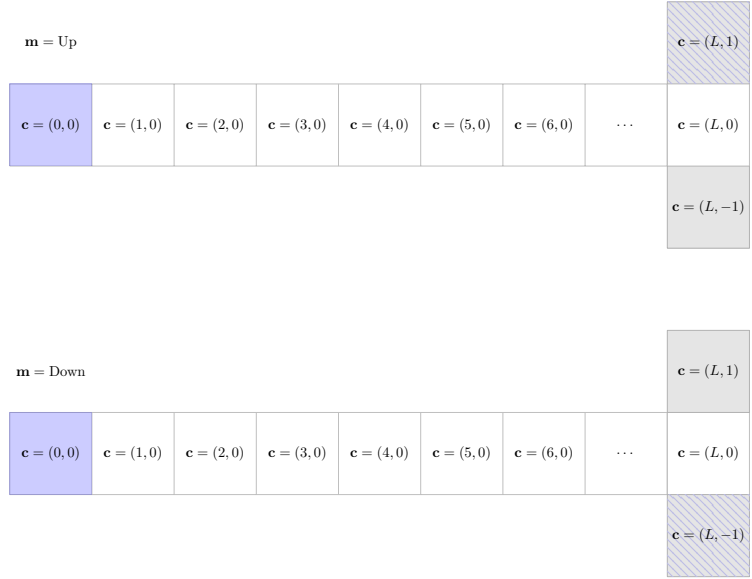

\paragraph{State space.}

The discrete state space $\S$ is composed of the set of positions $\C$ for the
agent in each of the two maze layouts $\M$. The maze layout determines the
position of the treasure. Formally, we have
\begin{empheq}[left = \empheqlbrace]{align}
    \S &= \M \times \C \\
    \M &= \{\text{Up}, \text{Down}\} \\
    \C &= \left\{ (0 ,  0), \dots, (L ,  0) \right\} \cup \{(L, 1), (L, -1)\}
\end{empheq}

A state $\s_t \in \S$ is thus defined by $\s_t = (\m_t, \c_t)$ with $\m_t \in
\M$ and $\c_t \in \C$. Let us also define $\F = \left\{ \s_t = (\m_t, \c_t) \in
\S \mid \c_t \in \{(L, 1), (L, -1)\} \right\}$ the set of terminal states, four
in number.

\paragraph{Action space.}

The discrete action space $\A$ is composed of the four possible moves that the
agent can take
\begin{equation}
    \A = \left\{
        (1,  0), (0,  1), (-1,  0), (0, -1)
    \right\}
\end{equation}
that correspond to Right, Up, Left and Down, respectively.

\paragraph{Observation space.}

The discrete observation space $\O$ is composed of the four partial
observations of the state that the agent can perceive
\begin{equation}
    \O = \left\{
        \text{Up}, \text{Down}, \text{Corridor}, \text{Junction}
    \right\} .
\end{equation}

\paragraph{Initial state distribution.}

The two possible initial states are $\s_0^\text{Up} = (\text{Up}, (0, 0))$ and
$\s_0^\text{Down} = (\text{Down}, (0 ,  0))$, depending on the maze in which
the agent lies. The initial state distribution $p_0: \S \-> [0, 1]$ is thus
given by
\begin{equation}
    p_0(\s_0) =
    \begin{cases}
        0.5 & \text{ if } \s_0 = \s_0^\text{Up} \\
        0.5 & \text{ if } \s_0 = \s_0^\text{Down} \\
        0 & \text{ otherwise }
    \end{cases}
\end{equation}

\paragraph{Transition distribution.}

The transition distribution function $T\colon \S \times \A \times \S \-> [0,
1]$ is given by
\begin{equation}
    T(\s_{t+1} \mid \s_t, \a_t) =
    \begin{cases}
        \delta_{\s_t}(\s_{t+1}) & \text{ if } \s_t \in \F \\
        (1 - \lambda) \delta_{f(\s_t, \a_t)}(\s_{t+1}) + \frac{\lambda}{4}
        \left(
            \sum_{\a \in \A}
            \delta_{f(\s_t, \a)}(\s_{t+1})
        \right)
            & \text{ otherwise }
    \end{cases}
\end{equation}
where $\s_t \in \S, \a_t \in \A$ and $\s_{t+1} \in \S$, and $f$ is given by
\begin{equation}
    f(\s_t, \a_t) =
    \begin{cases}
        \s_{t+1} = (\m_t, \c_t + \a_t)
            & \text{ if } \s_t \not \in \F, \c_t + \a_t \in \C \\
        \s_{t+1} = (\m_t, \c_t)
            & \text{ otherwise }
    \end{cases}
\end{equation}
where $\s_t = (\m_t, \c_t) \in \S$ and $\a_t \in \A$.

\paragraph{Reward function.}

The reward function $R: \S \times \A \times \S \-> \R$ is given by
\begin{equation}
    R(\s_t, \a_t, \s_{t+1}) =
    \begin{cases}
        0 & \text{ if } \s_t \in \F \\
        0 & \text{ if } \s_t \not \in \F, \s_{t+1} \not \in \F,
            \s_t \neq \s_{t+1} \\
        - 0.1 & \text { if } \s_t \not \in \F, \s_{t+1} \not \in \F,
            \s_t = \s_{t+1} \\
        4 & \text{ if } \s_t \not \in \F, \s_{t+1} \in \F, \c_{t+1} =
            \begin{cases}
                (L, 1) &\text{ if } \m_{t+1} = \text{Up} \\
                (L, -1) &\text{ if } \m_{t+1} = \text{Down}
            \end{cases} \\
        -0.1 & \text{ if } \s_t \not \in \F, \s_{t+1} \in \F, \c_{t+1}
            = \begin{cases}
                (L, -1) &\text{ if } \m_{t+1} = \text{Up} \\
                (L, +1) &\text{ if } \m_{t+1} = \text{Down}
            \end{cases}
    \end{cases}
\end{equation}
where $\s_t = (\m_t, \c_t) \in \S, \a_t \in \A$ and $\s_{t+1} = (\m_{t+1},
\c_{t+1}) \in \S$.

\paragraph{Observation distribution.}

In the T-Maze, the observations are deterministic. The observation distribution
$O\colon \S \times \O \-> [0, 1]$ is given by
\begin{equation}
    O(\o_t \mid \s_t) =
    \begin{cases}
        1 & \text{ if } \o_t = \text{Up}, \c_t = (0, 0), \m_t = \text{Up} \\
        1 & \text{ if } \o_t = \text{Down}, \c_t = (0, 0),
            \m_t = \text{Down} \\
        1 & \text{ if } \o_t = \text{Corridor}, \c_t \in \left\{
            (1 ,  0), \dots, (L-1 ,  0) \right\} \\
        1 & \text{ if } \o_t = \text{Junction}, \c_t \in \left\{
            (L ,  0), (L, 1), (L, -1) \right\} \\
        0 & \text{ otherwise }
    \end{cases}
\end{equation}
where $\s_t = (\m_t, \c_t) \in \S$ and $\o_t \in \O$.

\paragraph{Exploration policy.}

The exploration policy $\mathcal{E}: \A \-> [0, 1]$ is a stochastic policy that
is given by $\mathcal{E}(\text{Right}) = 1/2$ and $\mathcal{E}(\text{Other}) =
1/6$ where $\text{Other} \in \left\{ \text{Up}, \text{Left}, \text{Down}
\right\}$. It enforces the exploration of the right hand side of the maze
layouts. This exploration policy, tailored to the T-Maze environment, allows
one to speed up the training procedure, without interfering with the study of
this work.

\paragraph{Truncation horizon.}

The truncation horizon $H$ of the DRQN algorithm is chosen such that the
expected displacement of an agent moving according to the exploration policy in
a T-Maze with an infinite corridor on both sides is greater than $L$. Let $r =
\mathcal{E}(\text{Right})$ and $l = \mathcal{E}(\text{Left})$. In this infinite
T-Maze, the probability of increasing its position is $p = (1 - \lambda) r +
\lambda \frac{1}{4}$ and the probability of decreasing its position is $q = (1
- \lambda) l + \lambda \frac{1}{4}$. As a consequence, starting at \num{0}, the
expected displacement after one time step is $\bar{x}_1 = (1 - \lambda) (r -
l)$. By independence, $\bar{x}_H = H \bar{x}_1$ such that, for $\bar{x}_H \geq
L$, the time horizon is given by
\begin{equation}
    H = \left\lceil \frac{L}{(1 - \lambda)(r - l)} \right\rceil .
\end{equation}

\subsection{Mountain Hike environments} \label{app:hike}

The Varying Mountain Hike environment is a POMDP $(\S, \A, \O, p_0, T, R, O,
\gamma)$ parameterised by the sensor variance $\sigma_O \in \R$ and the
transition variance $\sigma_T \in \R$. The formal definition of this
environment is given below.

\begin{figure}[!ht]
    \centering
    \includegraphics[width=0.45\textwidth]{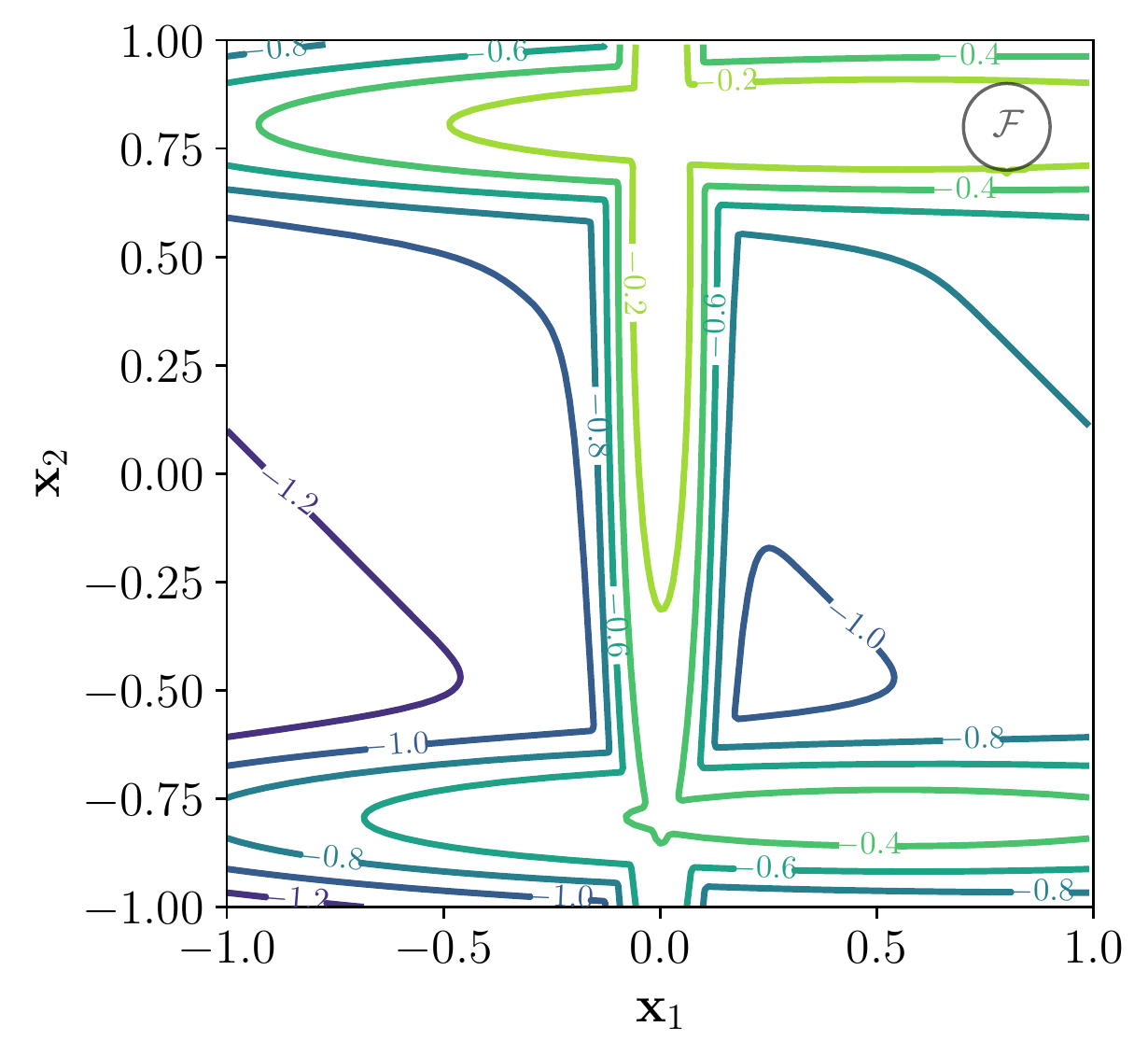}
    \includegraphics[width=0.49\textwidth]{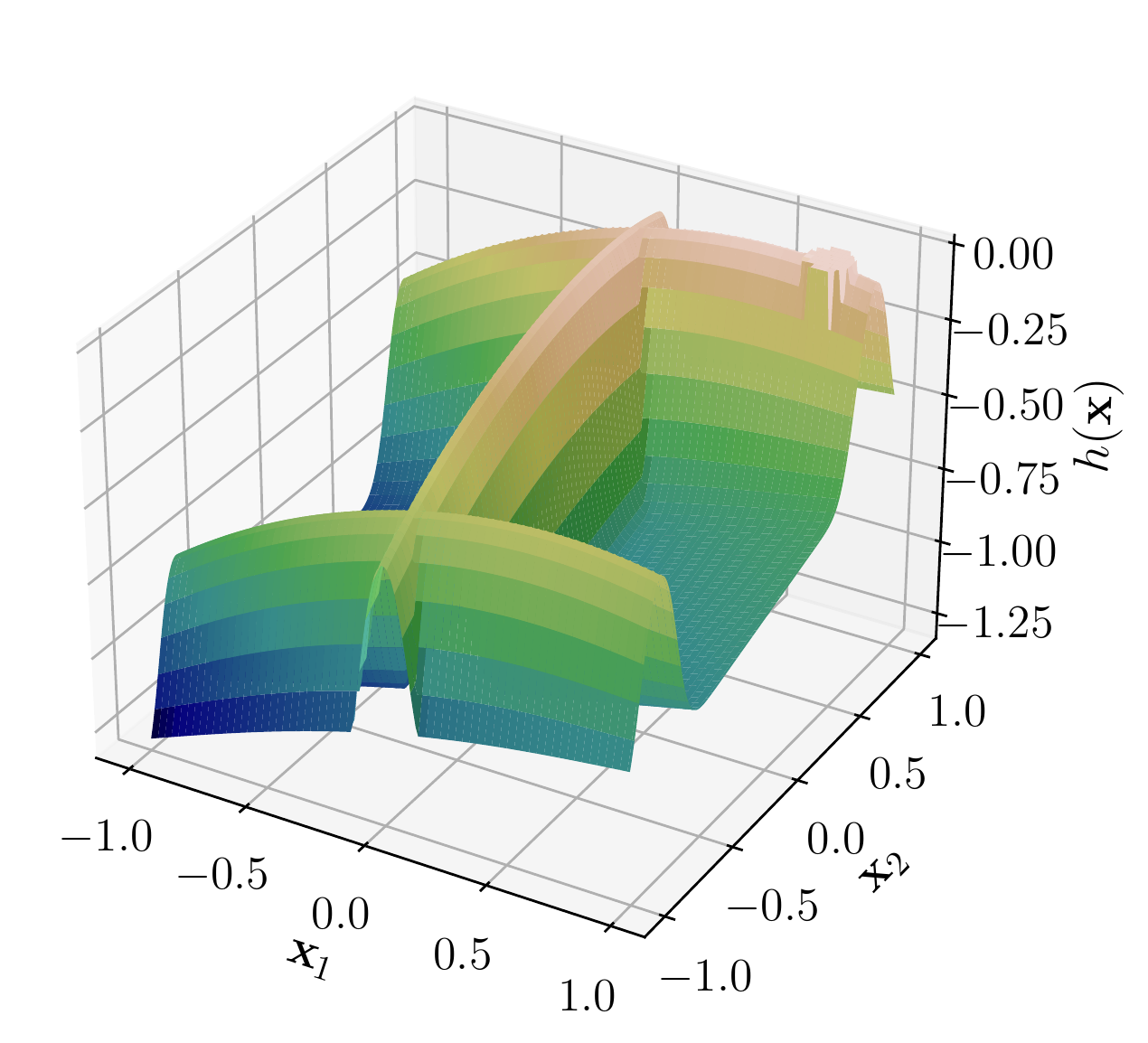}
    \caption{Mountain hike altitude function $h$ in $\X$.}
    \label{fig:hike_reward}
\end{figure}

\paragraph{State space.}

The state space $\S$ is the set of positions $\X$ and orientations $\C$ that
the agent can take. Formally, we have
\begin{empheq}[left = \empheqlbrace]{align}
    \S &= \X \times \C \\
    \X &= [-1, 1]^2 \\
    \C &= \{0^\circ, 90^\circ, 180^\circ, 270^\circ\}
\end{empheq}
The orientation $\c = 0^\circ, 90^\circ, 180^\circ$ and $270^\circ$ corresponds
to facing East, North, West and South, respectively. The set of terminal states
is $\F = \left\{\s = (\x, \c) \in \S \mid \lVert \x - (0.8, 0.8) \lVert <
0.1\right\}$.

\paragraph{Action space.}

The discrete action space $\A$ is composed of the four possible directions in
which the agent can move
\begin{equation}
    \A = \left\{ (0, 0.1), (-0.1, 0), (0, -0.1), (0.1, 0) \right\}
\end{equation}
that correspond to Forward, Left, Backward and Right, respectively.

\paragraph{Observation space.}

The continuous observation space is $\O = \mathbb{R}$.

\paragraph{Initial state distribution.}

The initial position is always is always $\x = (-0.8, -0.8)$ and the initial
orientation is sampled uniformly in $\C$, such that the initial state
distribution $p_0: \S \-> [0, 1]$ is given by
\begin{equation}
    p_0(\s_0) = \sum_{\c \in \C} \frac{1}{|\C|}
        \delta_{((-0.8, -0.8), \c)}(\s_0)
\end{equation}

\paragraph{Transition distribution.}

The transition distribution $T\colon \S \times \A \times \S \-> [0, 1]$ is
given by the conditional probability distribution of the random variable
$(\s_{t+1} \mid \s_t, \a_t)$ that is defined as
\begin{equation}
    \s_{t+1} = \begin{cases}
        \s_t & \text{ if } \s_t \in \F \\
        \operatorname{clamp}_\S \left(
            \s_t + R(\c) \, \a_t + \mathcal{N}(0, \sigma_T)
        \right) & \text{ otherwise }
    \end{cases}
\end{equation}
where $\operatorname{clamp}_\S(\s)$ is the function that maps $\s$ to the point
in $\S$ that minimizes its distance with $\s$, and
\begin{equation}
    R(\c) = \begin{pmatrix} \cos\c & -\sin\c \\ \sin\c & \cos\c \end{pmatrix}
\end{equation}
is the two-dimensional rotation matrix for an angle $\c$.

\paragraph{Reward function.}

The reward function $R\colon \S \times \A \times \S \-> \R$ is given by
\begin{equation}
    R(\s_t, \a_t, \s_{t+1}) =
    \begin{cases}
        0 & \text{ if } \s_t \in \F \\
        h(\s_{t+1}) & \text{ otherwise }
    \end{cases}
\end{equation}
where $\s_t \in \S, \a_t \in \A, \s_{t+1} \in \S$, and $h: \S \rightarrow \R^-$
is the function that gives the relative altitude to the mountain top in any
state. Note that the altitude is independent of the agent orientation.

\paragraph{Observation distribution.}

The observation distribution $O\colon \S \times \O \-> [0, 1]$ is given by
\begin{equation}
    O(\o_t \mid \s_t) = \phi(\o_t ; h(\s_t), \sigma_O^2)
\end{equation}
where $\s_t \in \S$ and $\o_t \in \O$, and where $\phi(\cdot ; \mu, \sigma^2)$
denotes the probability density function of a univariate Gaussian random
variable with mean $\mu$ and standard deviation $\sigma$.

\paragraph{Mountain Hike.}

The Mountain Hike environment is a POMDP $(\S, \A, \O, p_0, T, R, O, \gamma)$,
parameterised by the sensor variance $\sigma_O \in \R$ and the transition
variance $\sigma_T \in \R$. The formal definition of this environment is
identical to that of the Varying Mountain Hike, except that the initial
orientation of the agent is always North, which makes it an easier problem. The
initial state distribution is thus given by
\begin{equation}
    p_0(\s_0) = \delta_{((-0.8, -0.8), 90^\circ)}(\s_0) .
\end{equation}

\paragraph{Exploration policy.}

The uniform distribution $\U(\A)$ over the action space $\A$ is chosen as the
exploration policy $\mathcal{E}(\A)$.

\paragraph{Truncation horizon.}

The truncation horizon of the DRQN algorithm is chosen equal to $H = 80$ for
the Mountain Hike environment and $H = 160$ for the Varying Mountain Hike
environment.

\section{Deep recurrent Q-network} \label{app:drqn}

The DRQN algorithm is an instance of the PRQL algorithm that introduces several
improvements over vanilla PRQL. First, it is adapted to the online setting by
interleaving the generation of episodes and the update of the estimation
$\Q_\theta$. In addition, in the DRQN algorithm, the episodes are generated
with the $\varepsilon$-greedy policy $\sigma^\varepsilon_{\theta}: \H \->
\P(\A)$, derived from the current estimation $\Q_\theta$. This stochastic
policy selects actions according to $\argmax_{\a \in \A} \Q_\theta(\cdot, \a)$
with probability $1 - \varepsilon$, and according to an exploration policy
$\mathcal{E}(\A) \in \P(\A)$ with probability $\varepsilon$. In addition, a
replay buffer of histories is used and the gradient is evaluated on a batch of
histories sampled from this buffer. Furthermore, the parameters $\theta$ are
updated with the Adam algorithm \citep{kingma2014adam}. Finally, the target
$r_t + \gamma \max_{\a \in \A} \Q_{\theta'}(\eta_{0:t+1}, \a)$ is computed
using a past version $\Q_{\theta'}$ of the estimation $\Q_\theta$  with
parameters $\theta'$ that are updated to $\theta$ less frequently, which eases
the convergence towards the target, and ultimately towards the \cqf{}. The DRQN
training procedure is detailed in \autoref{algo:drqn}.

\begin{algo}
    \DontPrintSemicolon \footnotesize
    \caption{DRQN - \cqf{} approximation}
    \label{algo:drqn}

    \SetKwInOut{Parameters}{Parameters}
    \SetKwInOut{Inputs}{Inputs}

    \Parameters{%
        $N \in \N$ the buffer capacity. \\
        $C \in \N$ the target update period (in episodes). \\
        $E \in \N$ the number of episodes. \\
        $H \in \N$ the truncation horizon. \\
        $I \in \N$ the number of gradient steps after each episode. \\
        $\varepsilon \in \R$ the exploration rate. \\
        $\alpha \in \R$ the learning rate. \\
        $B \in \N$ the batch size.}

    \Inputs{%
        $(\S, \A, \O, T, R, O, p_0, \gamma)$ a POMDP. \\
        $\mathcal{E}(\A) \in \P(\A)$ the exploration policy.}

    Initialise empty replay buffer $\B$ \;
    Initialise parameters $\theta$ randomly \;

    \For{$e = 0, \dots, E - 1$}{

        \If{$e \bmod C = 0$}{%
            Update target network with $\theta' \leftarrow \theta$
        }

        \tcp{Generate new episode, store history and rewards}

        Draw an initial state $\s_0$ according to $p_0$ and observe $\o_0$ \;

        Let $\eta_{0:0} = (\o_0)$  \;

        \For{$t = 0, \dots, H - 1$}{
            Select $\a_t \sim \mathcal{E}(\A)$ with
            probability $\varepsilon$, otherwise select
            $\a_t = \argmax_{\a \in \A} \left\{
                \Q_\theta(\eta_{0:t}, \a )
            \right\}$ \;

            Take action $\a_t$ and observe $r_t$ and $\o_{t+1}$ \;
            Let $\eta_{0:t+1} = (\o_0, \a_0, \o_1, \dots, \o_{t+1})$ \;

            \lIf{$|\B| < N$}{
                add $(\eta_{0:t}, \a_t, r_t, \o_{t+1}, \eta_{0:t+1})$ in replay
                buffer $\B$
            }
            \lElse{%
                replace oldest transition in replay buffer $\B$ by
                $(\eta_{0:t}, \a_t, r_t, \o_{t+1}, \eta_{0:t+1})$
            }

            \If{$\o_{t+1}$ is terminal}{%
                \textbf{break}
            }
        }

        \tcp{Optimise recurrent Q-network}

        \For{$i = 0, \dots, I - 1$}{

            Sample $B$ transitions $(\eta_{0:t}^b, \a_t^b, r_t^b,
            \o_{t + 1}^b, \eta_{0:t+1}^b)$ uniformly from the
            replay buffer $\B$ \;

            Compute targets $y^b = \begin{cases}
                r_t^b + \gamma \max_{\a \in \A}
                    \left\{
                        \Q_{\theta'} (\eta_{0:t+1}^b, \a)
                    \right\}
                    & \text{if } \o_{t + 1}^b \text{ is not terminal} \\
                r_t^b & \text{otherwise}
            \end{cases}$ \;

            Compute loss $L = \sum_{b = 0}^{B - 1} \left( y^b -
            \Q_\theta(\eta_{0:t}^b, \a_t^b) \right)^2$ \;

            Compute direction $g$ using Adam optimiser, perform gradient step
            $\theta \leftarrow \theta + \alpha g$ \;
        }
    }
\end{algo}

\section{Particle filtering} \label{app:particle}

As explained in \autoref{sec:background}, the belief filter becomes intractable
for certain POMDPs. In particular, POMDPs with continuous state space require
one to perform an integration over the state space. Furthermore, in these
environments, the belief should be represented by a function over a continuous
domain instead of a finite-dimensional vector. Such arbitrary beliefs cannot be
represented in a digital computer.

To overcome these two difficulties, the particle filtering algorithm proposes
to represent an approximation of the belief by a finite set of samples that
follows the belief distribution. In other words, we represent $b_t \in \P(\S)$
by the set of $M$ samples
\begin{equation}
    S_t = \left \{ \s_t^{m} \right \}_{m = 0}^{M-1}
    \label{eq:sample_represented_belief}
\end{equation}
where $\s_t^m \in \S, \, m = 0, \dots, {M-1}$ being independent realisations of
the distribution $b_t$.

Particle filtering is a procedure that allows one to sample a set of states
$S_t$ that follow the belief distribution $b_t$. The set is thus updated each
time that a new action $\a_{t-1}$ is taken and a new observation $\o_t$ is
observed. Although this procedure does not require to evaluate expression
\eqref{eq:belief_update}, it is necessary to be able to sample from the initial
state distribution $p_0$ and from the transition distribution $T$, and to be
able to evaluate the observation distribution $O$. This process, illustrated in
\autoref{algo:particle}, guarantees that the successive sets $S_0, \dots, S_H$
have (weighted) samples following the probability distribution $b_0, \dots,
b_H$ defined by equation \eqref{eq:belief_update}.

\begin{algo}
    \DontPrintSemicolon \footnotesize
    \caption{Particle filtering}
    \label{algo:particle}

    \SetKwInOut{Parameters}{Parameters}
    \SetKwInOut{Inputs}{Inputs}

    \Parameters{%
        $M \in \N$ the number of particles}

    \Inputs{%
        $(\S, \A, \O, T, R, O, p_0, \gamma)$ a POMDP. \\
        $H \in \N$ the number of transitions \\
        $\eta_{0:H} = (\o_0, \a_0, \dots, \o_{H-1}, \a_{H-1}, \o_H) \in
            \H_{0:H}$ a history}

    \tcp{Generate weighted samples following the initial belief $b_0$}

    Sample $\s_0^0, \dots, \s_0^{M - 1} \sim p_0$ \;

    $\eta \leftarrow 0$ \;

    \For{$m = 0, \dots, M - 1$}{
        $w_0^m \leftarrow O(\o_0 \mid \s_0^m)$ \;

        $\eta \leftarrow \eta + w_0^m$ \;
    }

    \For{$m = 0, \dots, M - 1$}{
        $w_0^m \leftarrow w_0^m / \eta$ \;
    }

    $S_0 = \left\{ (\s_0^m, w_0^m) \right\}_{m = 0}^{M - 1}$ \;

    \tcp{Generate successive weighted samples following the beliefs
        $b_1, \dots, b_H$}

    \For{$t = 1, \dots, H$}{
        $\eta \leftarrow 0$ \;

        \For{$m = 0, \dots, {M - 1}$}{
            Sample $l \in \{0, \dots, M - 1\}$ according to $p(l) = w_{t-1}^l$
            \;

            Sample $\s_t^m \sim T(\cdot \mid \s_{t-1}^l, \a_{t-1})$ \;

            $w_t^m \leftarrow O(\o_t \mid \s_t^m)$ \;

            $\eta \leftarrow \eta + w_t^m$ \;
        }

        \For{$m = 0, \dots, {M - 1}$}{
            $w_t^m \leftarrow w_t^m / \eta$ \;
        }

        $S_t = \left\{ (\s_t^m, w_t^m) \right\}_{m = 0}^{M - 1}$ \;
    }
\end{algo}

\autoref{algo:particle} starts from $N$ samples from the initial distribution
$p_0$. These samples are initially weighted by their likelihood $O(\o_0 \mid
\s_0^n)$. Then, we have three steps that are repeated at each time step. First,
the samples are resampled according to their weights. Then, given the action,
the samples are updated by sampling from $T(\cdot \mid \s_t^n, \a_t)$. Finally,
these new samples are weighted by their likelihood $O(\o_{t+1} \mid
\s_{t+1}^n)$ given the new observation $\o_{t+1}$, as for the initial samples.
As stated above, this method ensures that the (weighted) samples follow the
distribution of the successive beliefs.

\section{Mutual information neural estimator} \label{app:mine}

In \autoref{subsec:mine_estimator}, the MI estimator that is used in the
experiments is formally defined, and the algorithm that is used to derive this
estimator is detailed. In \autoref{subsec:deepsets}, we formalise the extension
of the MINE algorithm with the Deep Set architecture.

\subsection{Estimator} \label{subsec:mine_estimator}

As explained in \autoref{subsec:mine}, the ideal MI neural estimator, for a
parameter space $\Phi$, is given by
\begin{align}
    I_\Phi(X; Y) &= \sup_{\phi \in \Phi} i_\phi(X; Y) \\
    i_\phi(X; Y) &= \E{z \sim p}{T_\phi(z)} -
    \log\left(\E{z \sim q}{e^{T_\phi(z)}}\right)
    \label{eq:mine}
\end{align}
However, both the estimation of the expectations and the computation of the
supremum are intractable. In practice, the expectations are thus estimated with
the empirical means over the set of samples $\left\{ (\x^n, \y^n) \right\}_{n =
0}^{N - 1}$ drawn from the joint distribution $p$ and the set of samples
$\left\{ (\x^n, \tilde{\y}^n) \right\}_{n = 0}^{N - 1}$ obtained by permuting
the samples from $Y$, such that the pairs follow the product of marginal
distributions $q = p_X \otimes p_Y$. In order to estimate the supremum over the
parameter space $\Phi$, the MINE algorithm proposes to maximise $i_\phi(X; Y)$
by stochastic gradient ascent over batches from the two sets of samples, as
detailed in \autoref{algo:mine}. The final parameters $\phi^*$ obtained by this
maximisation procedure define the estimator
\begin{equation}
    \hat{I} = \frac{1}{N} \sum_{n = 0}^{N-1}
    {T_{\phi^*}(\x^n, \y^n)} - \log\left(
        \frac{1}{N} \sum_{n = 0}^{N-1}
        {e^{T_{\phi^*}(\x^n, \tilde{\y}^n)}}
    \right)
    \label{eq:mine_estimate}
\end{equation}
that is used in the experiments. This algorithm was initially proposed in
\citep{belghazi2018mutual}.

\begin{algo}
    \DontPrintSemicolon \footnotesize
    \caption{MINE - lower bound optimization}
    \label{algo:mine}

    \SetKwInOut{Parameters}{Parameters}
    \SetKwInOut{Inputs}{Inputs}

    \Parameters{%
        $E \in \N$ the number of episodes. \\
        $B \in \N$ the batch size. \\
        $\alpha \in \R$ the learning rate.}

    \Inputs{%
        $N \in \N$ the number of samples. \\
        $\D = \left\{ (\x^n, \y^n) \right\}_{n = 0}^{N - 1}$ the set of samples
            from the joint distribution.}

    Initialise parameters $\phi$ randomly. \;

    \For{$e = 0, \dots, E - 1$}{

        Let $p$ a random permutation of $\{0, \dots, N - 1\}$. \;
        Let $\tilde{p}_1$ a random permutation of $\{0, \dots, N - 1\}$. \;
        Let $\tilde{p}_2$ a random permutation of $\{0, \dots, N - 1\}$. \;

        \While{$i = 0, \dots, \left\lfloor \frac{N}{B} \right\rfloor$}{

            Let $S \leftarrow \left\{
                (\x^{p(k)}, \y^{p(k)})
            \right\}_{k = iB}^{(i+1)B - 1}$
            a batch of samples from the joint distribution. \;

            Let $\tilde{S} \leftarrow \left\{
                (\x^{\tilde{p}_1(k)}, \y^{\tilde{p}_2(k)})
            \right\}_{k = iB}^{(i+1)B - 1} $
            a batch of samples from the product of marginal distributions \;

            Evaluate the lower bound
            $$
                L(\phi) \leftarrow \frac{1}{B} \sum_{(\x, \y) \in S}
                    T_\phi(\x, \y) - \log \left(
                    \frac{1}{B} \sum_{(\tilde{x}, \tilde{y}) \in \tilde{S}}
                    e^{T_\phi(\tilde{\x}, \tilde{\y})}
                \right)
            $$ \;

            Evaluate bias corrected gradients
            $G(\phi) \leftarrow \tilde{\nabla}_{\!\phi} L(\phi)$ \;

            Update network parameters with $\phi \leftarrow \phi + \alpha
            G(\phi)$ \;
        }
    }
\end{algo}

\subsection{Deep sets} \label{subsec:deepsets}

As explained in \autoref{subsec:protocol}, the belief computation is
intractable for environments with continuous state spaces. In the experiments,
the belief of such environments is approximated by a set of particles $S =
\left\{ \s^m \right\}_{m = 1}^M$ that are guaranteed to follow the belief
distribution, such that $\s^m \sim b, \; \forall \s^m \in S$ (see
\autoref{app:particle}). Those particles could be used for constructing an
approximation of the belief distribution, a problem known as density
estimation. We nonetheless do not need an explicit estimate of this
distribution. Instead, the particles can be directly consumed by the MINE
network. In this case, the two sets of input samples of the MINE algorithm take
the form
\begin{align}
    \left\{ (\x^n, \y^n) \right\}_{n = 0}^{N - 1}
    &= \left\{ (\h^n, S^n) \right\}_{n = 0}^{N - 1} \\
    &= \left\{
        (\h^n, \left\{ \s^{n,m} \right\}_{m = 1}^M)
    \right\}_{n = 0}^{N - 1}.
\end{align}

In order to process particles from sets $S^n$ as input of the neural network
$T_\phi$, we choose an architecture that guarantees its invariance to
permutations of the particles. The deep set architecture
\citep{zaheer2017deep}, that is written as $\rho_\phi\left( \sum_{\s \in S}
\psi_\phi(\s) \right)$, provides such guarantees. Moreover, this architecture
is theoretically able to represent any function on sets, under the assumption
of having representative enough mappings $\rho_\phi$ and $\psi_\phi$ and the
additional assumption of using finite sets $S$ when particles come from an
uncountable set as in this work. The function $T_\phi$ is thus given by
\begin{equation}
    T_\phi(\h, S) =
        \mu_\phi\left(\h, \rho_\phi\left( \sum_{\s \in S}
        \psi_\phi(\s) \right)\right)
\end{equation}
when the belief is approximated by a set of particles.

\section{Hyperparameters} \label{app:hyperparameters}

The hyperparameters of the DRQN algorithm are given in \autoref{tab:hp_drqn}
and the hyperparameters of the MINE algorithm are given in
\autoref{tab:hp_mine}. The value of those hyperparameters have been chosen a
priori, except for the number of episodes of the DRQN algorithm and the number
of epochs of the MINE algorithm. These were chosen so as to ensure convergence
of the policy return and the MINE lower bound, respectively. The parameters of
the Mountain Hike and Varying Mountain Hike environments are given in
\autoref{tab:hp_hike}.

\begin{table}[!h]
    \centering
    \begin{tabular}{ccl}
        \toprule
        Name & Value & Description \\
        \midrule
        $S$ & \num{2} & Number of RNN layers \\
        $D$ & 1 & Number of linear layers (no activation function) \\
        $H$ & \num{32} & Hidden state size \\
        $N$ & \num{8192} & Replay buffer capacity \\
        $C$ & \num{10} & Target update period in term of episodes \\
        $I$ & \num{10} & Number of gradient steps after each episode \\
        $\epsilon$ & \num{0.2} & Exploration rate \\
        $B$ & \num{32} & Batch size \\
        $\alpha$ & \num{1e-3} & Adam learning rate \\
        \bottomrule
    \end{tabular}
    \caption{DRQN architecture and training hyperparameters.}
    \label{tab:hp_drqn}
\end{table}

\begin{table}[!h]
    \centering
    \begin{tabular}{ccl}
        \toprule
        Name & Value & Description \\
        \midrule
        $L$ & \num{2} & Number of hidden layers \\
        $H$ & \num{256} & Hidden layer size \\
        $N$ & \num{10000} & Training set size \\
        $E$ & \num{200} & Number of epochs \\
        $B$ & \num{1024} & Batch size \\
        $\alpha$ & \num{1e-3} & Adam learning rate \\
        $R$ & \num{16} & Representation size for the Deep Set architecture \\
        $\alpha$ & \num{0.01} & EMA rate for the bias corrected gradient \\
        \bottomrule
    \end{tabular}
    \caption{MINE architecture and training hyperparameters.}
    \label{tab:hp_mine}
\end{table}

\begin{table}[!h]
    \centering
    \begin{tabular}{ccl}
        \toprule
        Name & Value & Description \\
        \midrule
        $\sigma_O$ & 0.1 & Standard deviation of the observation noise \\
        $\sigma_T$ & 0.05 & Standard deviation of the transition noise \\
        \bottomrule
    \end{tabular}
    \caption{Mountain Hike and Varying Mountain Hike parameters.}
    \label{tab:hp_hike}
\end{table}

\newpage

\section{Generalisation to other distribution of histories}
\label{app:generalisation}

In this section, we study if the hidden state still provides information
about the belief under other distributions of histories than the one induced by
the learned policy \eqref{eq:hb_distribution}. This generalisation to other
distributions is desirable for building policies that are more robust to
perturbations of the histories.

We propose to study the evolution of the MI between the hidden state and the
belief when adding noise to the policy used to sample the histories. Formally,
instead of sampling the hidden states and beliefs according to
\eqref{eq:hb_distribution}, we propose to sample those according to
\begin{align}
    p_\varepsilon(\h, b)
    &= \sum_{t = 0}^\infty \; p(t)
    \int_{\H}
        p(\h, b \mid \eta) \;
        p_{\sigma_{\theta}^\varepsilon}(\eta \mid t)
        \d{\eta}
    \label{eq:hb_epsilon_distribution}
\end{align}
where $p(t)$ is once again chosen to the uniform distribution over the time
steps $p(t) = 1/H, \; t \in \{0, \dots, H - 1\}$, $\sigma_\theta^\varepsilon$
is the $\varepsilon$-greedy policy as defined in \autoref{app:drqn}, and
$p_{\sigma_{\theta}^\varepsilon}(\eta \mid t)$ gives the conditional
probability distribution induced by the policy $\sigma_\theta^\varepsilon$ over
histories $\eta \in \H$ given that their length is $t \in \N_0$. Note that the
training procedure remains unchanged.

\begin{figure}[!ht]
    \centering
    \begin{subfigure}[t]{0.49\textwidth}
        \includegraphics[width=\textwidth]
            {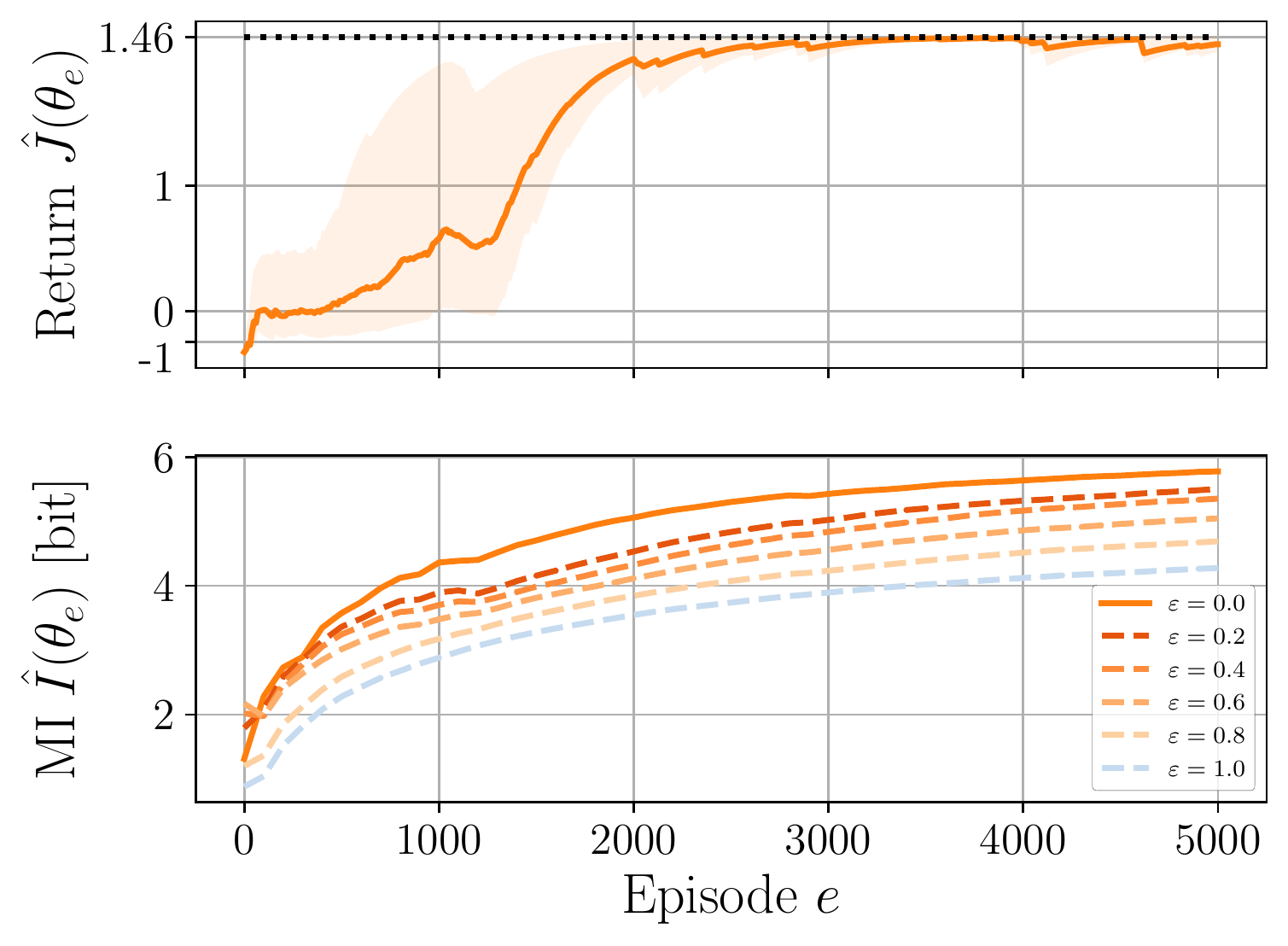}
        \vspace{-1.7em}
        \caption{Deterministic T-Maze ($L = 50$)}
        \vspace{1em}
        \label{fig:tm_generalisation}
    \end{subfigure}
    \begin{subfigure}[t]{0.49\textwidth}
        \includegraphics[width=\textwidth]
            {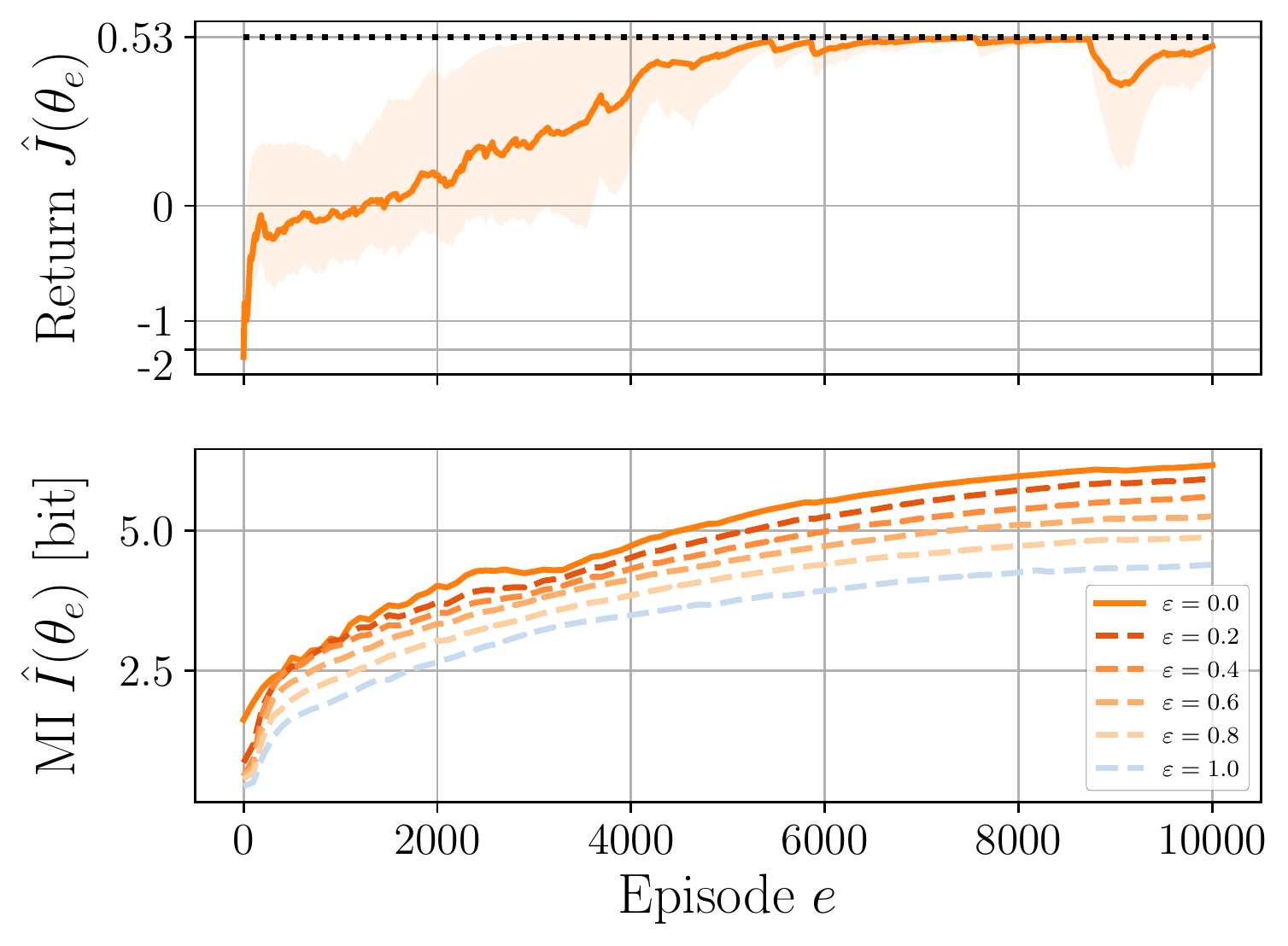}
        \vspace{-1.7em}
        \caption{Deterministic T-Maze ($L = 100$)}
        \vspace{1em}
        \label{fig:ltm_generalisation}
    \end{subfigure}
    \begin{subfigure}[t]{0.49\textwidth}
        \includegraphics[width=\textwidth]
            {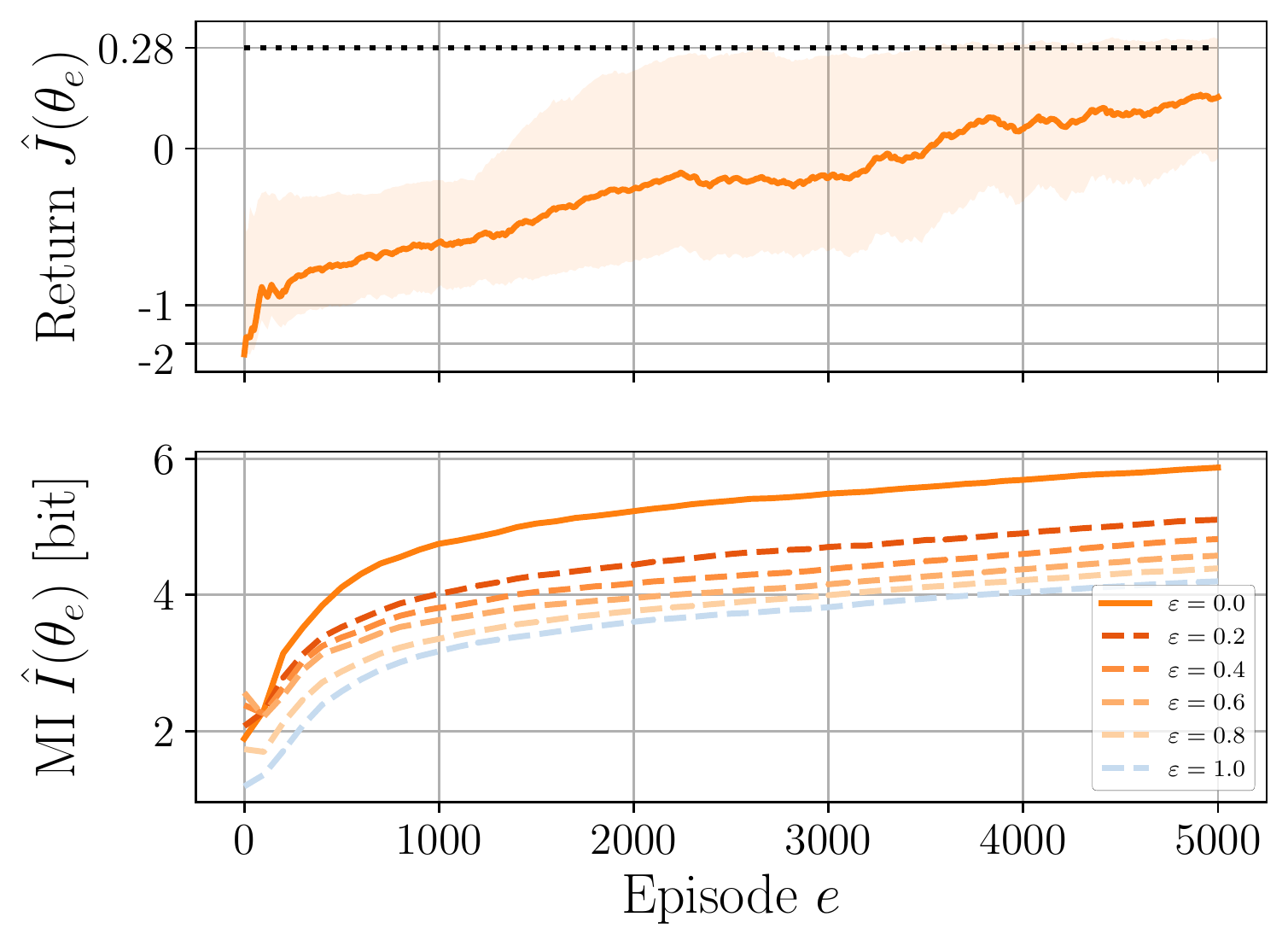}
        \vspace{-1.7em}
        \caption{Stochastic T-Maze ($L = 50$, $\lambda = 0.3$)}
        \vspace{1em}
        \label{fig:stm_generalisation}
    \end{subfigure}
    \begin{subfigure}[t]{0.49\textwidth}
        \includegraphics[width=\textwidth]
            {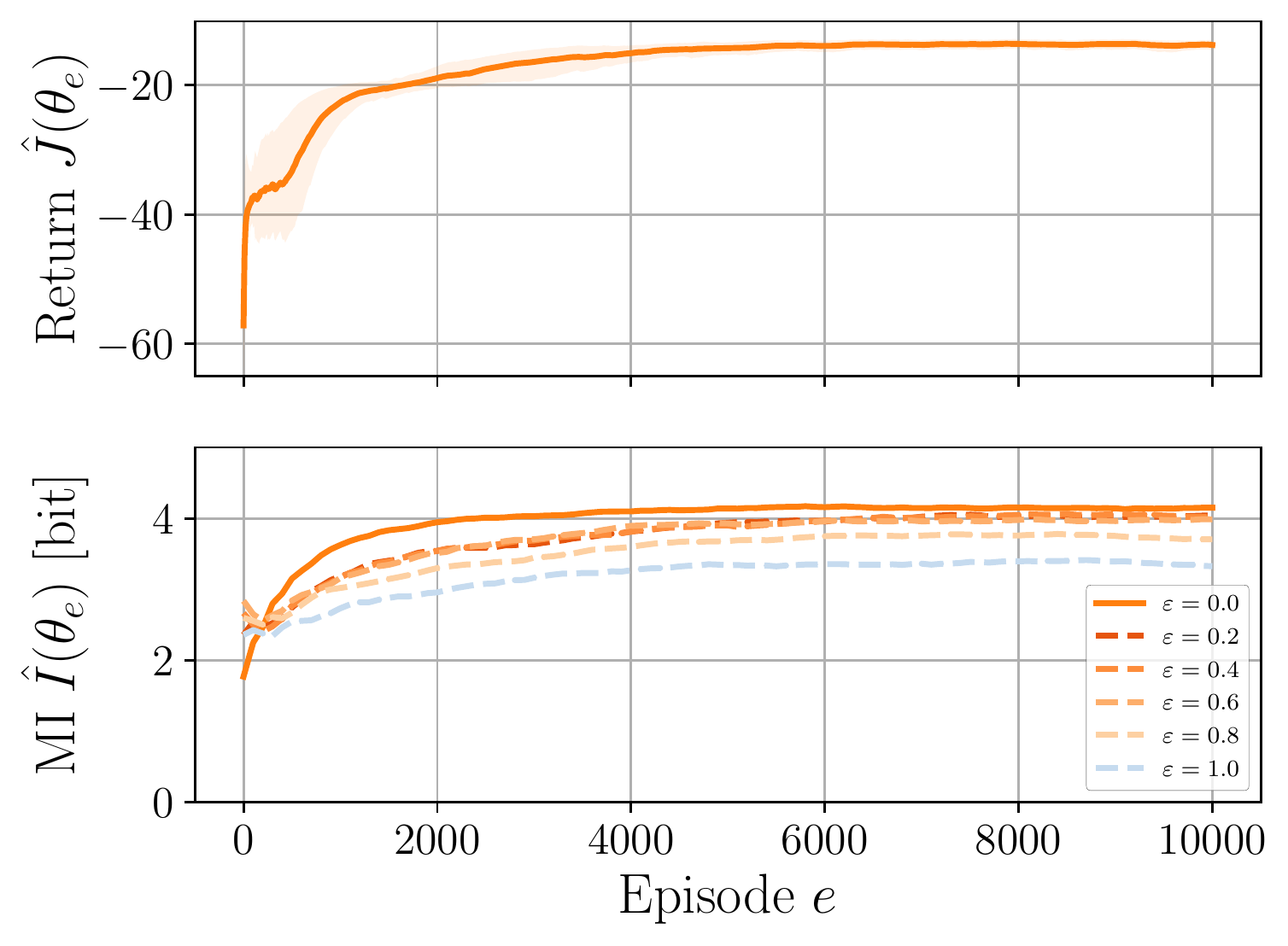}
        \vspace{-1.7em}
        \caption{Mountain Hike}
        \vspace{1em}
        \label{fig:mh_generalisation}
    \end{subfigure}
    \begin{subfigure}[t]{0.49\textwidth}
        \includegraphics[width=\textwidth]
            {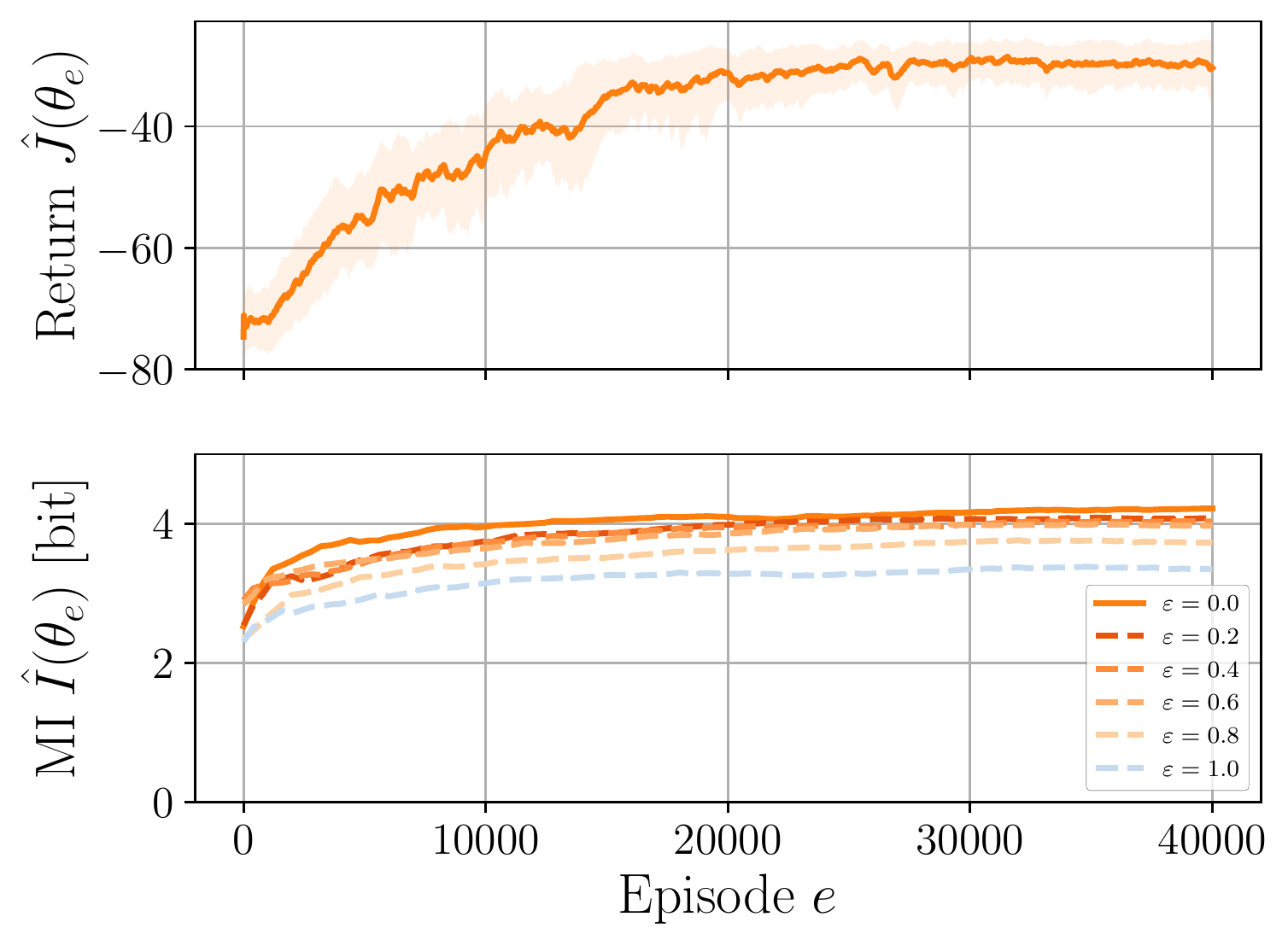}
        \vspace{-1.7em}
        \caption{Varying Mountain Hike}
        \vspace{1em}
        \label{fig:vmh_generalisation}
    \end{subfigure}
    \vspace{-1em}
    \caption{Evolution of the return $\hat{J}(\theta_e)$ and the MI
        $\hat{I}(\theta_e)$ after $e$ episodes, under distribution of histories
        induced by several $\varepsilon$-greedy policies, for the GRU cell.
        The maximal expected return is given by the dotted line.}
    \vspace{-1em}
    \label{fig:generalisation}
\end{figure}

The results of this additional study can be found in
\autoref{fig:generalisation}, for $\varepsilon \in \{0.0, 0.2, 0.4, 0.6, 0.8,
1.0\}$. It can be noted that $p_{0.0}$ is the distribution of hidden states and
beliefs induced by the learned policy \eqref{eq:hb_distribution}, and $p_{1.0}$
is the distribution of hidden states and beliefs induced by a fully random
policy. For reasons of computational capacity, this analysis was carried out
for the GRU cell only. This cell was chosen for being a standard cell that
performs well in all environments in terms of return, unlike the LSTM. As can
be seen in \autoref{fig:generalisation}, the MI between the hidden states and
the beliefs increases throughout the training process, under all considered
policies, even the fully random policy. We conclude that the correlation
between the hidden states and beliefs generalises reasonably well to other
distributions. In other words, the hidden states still capture information
about the beliefs even under other distributions of histories.

\section{Correlations between the empirical return and the estimated mutual
information} \label{app:correlations}

The correlation between the empirical return and the estimated MI are computed
with the Pearson's linear correlation coefficient and the Spearman's rank
correlation coefficient. These coefficients are reported for all environments
and all cells in \autoref{tab:pearson} and \autoref{tab:spearman}. The columns
named \emph{aggregated} give the correlation coefficients measured over all
samples of $\hat{I}$ and $\hat{J}$ from all cells.

\begin{table}[!h]
    \centering
    \begin{tabular}{r|cccccc}
        \toprule
        Environment &
            Aggregated &
            LSTM &
            GRU &
            BRC &
            nBRC &
            MGU \\
        \midrule
        T-Maze ($L = 50$, $\lambda = 0.0$) &
            \peartm{} &
            0.7329 &
            0.8500 &
            0.8747 &
            0.9314 &
            0.9178 \\
        T-Maze ($L = 100$, $\lambda = 0.0$) &
            \pearltm{} &
            0.3624 &
            0.6162 &
            0.6855 &
            0.6504 &
            0.6299 \\
        T-Maze ($L = 50$, $\lambda = 0.3$) &
            \pearstm{} &
            0.2882 &
            0.8008 &
            0.7229 &
            0.7424 &
            0.6159 \\
        Mountain Hike &
            \pearmh{} &
            0.7352 &
            0.6177 &
            0.4338 &
            0.5857 &
            0.5485 \\
        Varying Mountain Hike &
            \pearvmh{} &
            0.6712 &
            0.4530 &
            0.4446 &
            0.3669 &
            0.3006 \\
        \bottomrule
    \end{tabular}
    \caption{Pearson's linear correlation coefficient for each environment and
        cell.}
    \label{tab:pearson}
\end{table}

\begin{table}[!h]
    \centering
    \begin{tabular}{r|cccccc}
        \toprule
        Environment &
            Aggregated &
            LSTM &
            GRU &
            BRC &
            nBRC &
            MGU \\
        \midrule
        T-Maze ($L = 50$, $\lambda = 0.0$) &
            \speartm{} &
            0.7815 &
            0.5963 &
            0.5403 &
            0.4009 &
            0.5002 \\
        T-Maze ($L = 100$, $\lambda = 0.0$) &
            \spearltm{} &
            0.5969 &
            0.7108 &
            0.5058 &
            0.4605 &
            0.5534 \\
        T-Maze ($L = 50$, $\lambda = 0.3$) &
            \spearstm{} &
            0.3730 &
            0.6600 &
            0.5090 &
            0.4706 &
            0.6497 \\
        Mountain Hike &
            \spearmh{} &
            0.5933 &
            0.1443 &
            0.2762 &
            0.4337 &
            0.2630 \\
        Varying Mountain Hike &
            \spearvmh{} &
            0.6869 &
            0.3677 &
            0.4355 &
            0.2955 &
            0.2266 \\
        \bottomrule
    \end{tabular}
    \caption{Spearman's rank correlation coefficient for each environment and
        cell.}
    \label{tab:spearman}
\end{table}

\section{Belief of variables irrelevant for the optimal control}
\label{subsec:irrelevant_additional}

In this section, we report the evolution of the return and the MI between the
hidden states and the belief of both the relevant and irrelevant variables for
the LSTM, BRC, nBRC and MGU architectures. It completes the results obtained
for the GRU cell in \autoref{subsec:irrelevant}.

\begin{figure}[h]
    \centering
    \begin{subfigure}[t]{0.49\textwidth}
        \includegraphics[width=\textwidth]
            {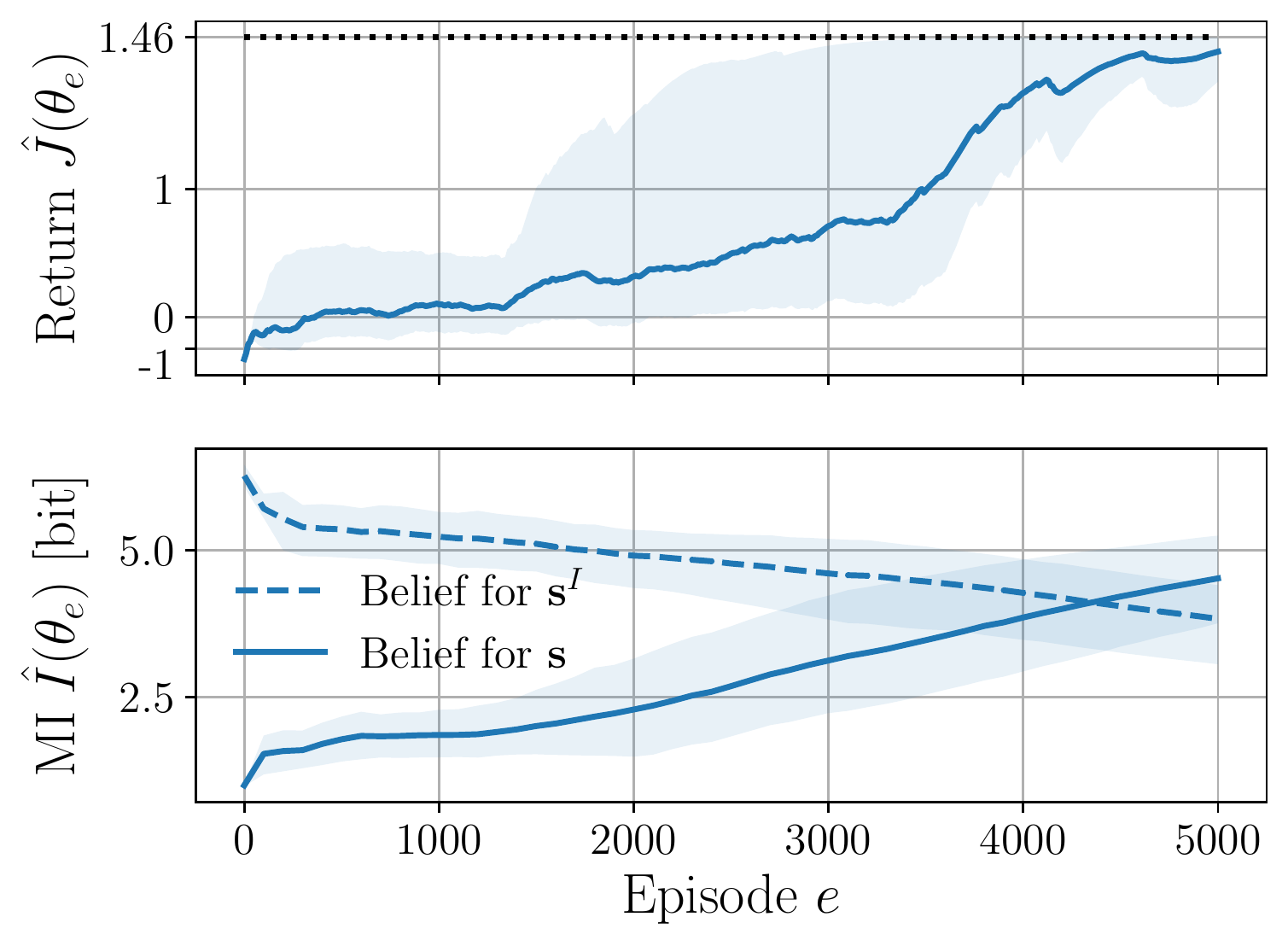}
        \vspace{-1.7em}
        \caption{\(d = 1\)}
        \vspace{1em}
        \label{fig:tmaze_irr_1_lstm}
    \end{subfigure}
    \begin{subfigure}[t]{0.49\textwidth}
        \includegraphics[width=\textwidth]
            {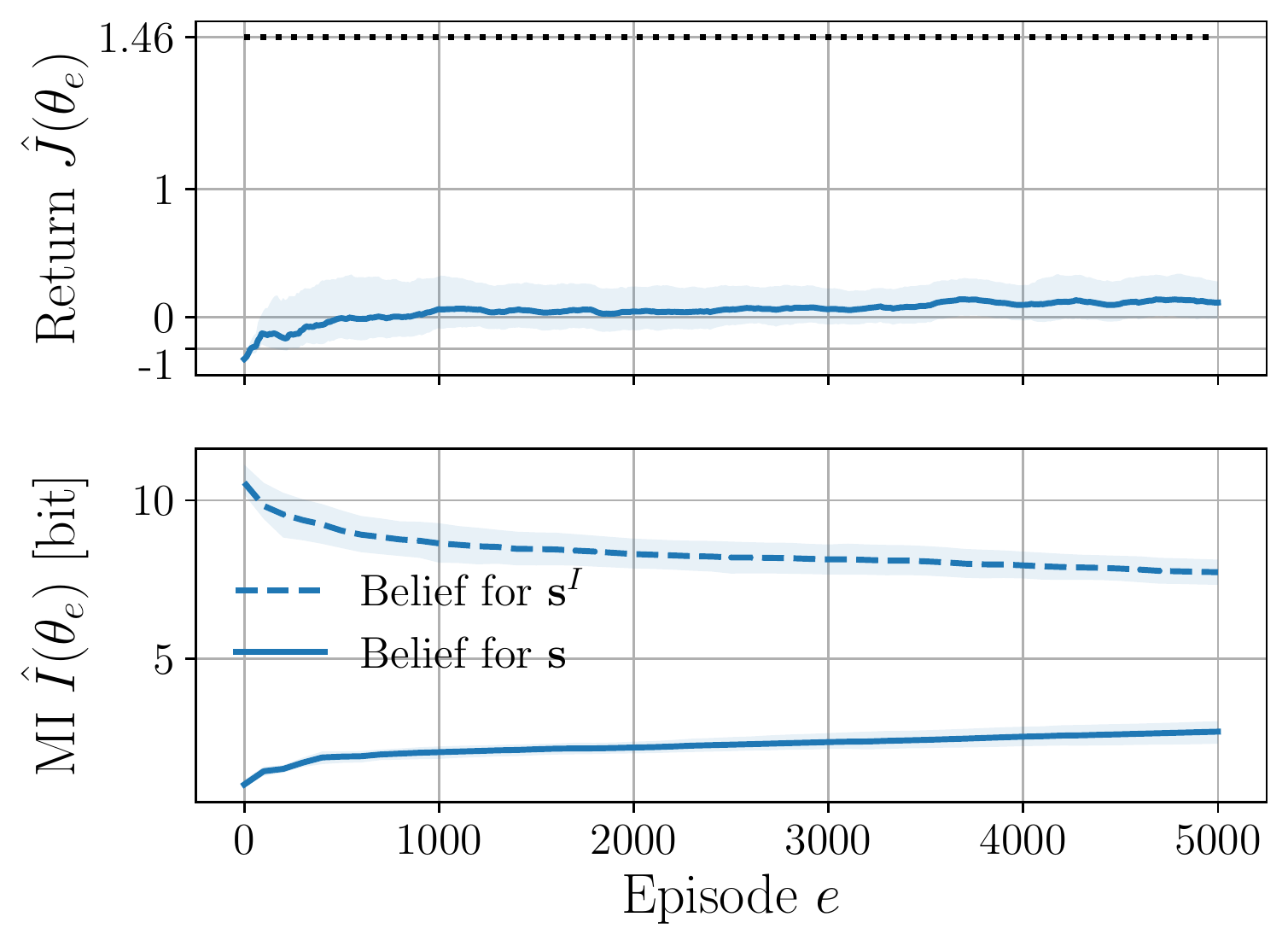}
        \vspace{-1.7em}
        \caption{\(d = 4\)}
        \vspace{1em}
        \label{fig:tmaze_irr_4_lstm}
    \end{subfigure}
    \vspace{-1em}
    \caption{Deterministic T-Maze (\(L = 50\)) with \(d\) irrelevant state
        variables.
        Evolution of the return \(\hat{J}(\theta_e)\) and the MI
        \(\hat{I}(\theta_e)\) for the belief of the irrelevant and relevant
        state variables after \(e\) episodes, for the LSTM cell.}
    \label{fig:tmaze_irr_lstm}
\end{figure}

\begin{figure}[h]
    \centering
    \begin{subfigure}[t]{0.49\textwidth}
        \includegraphics[width=\textwidth]
            {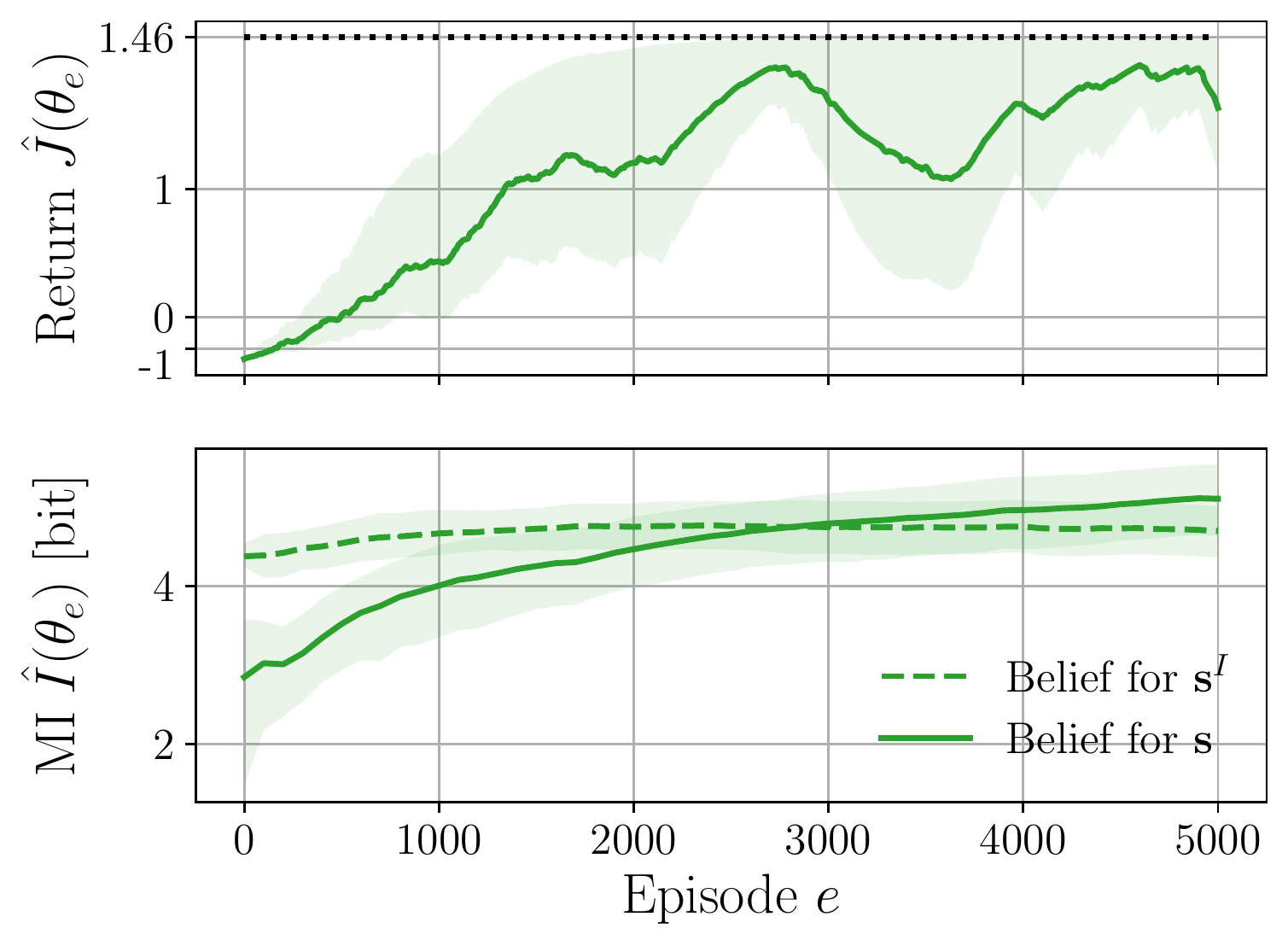}
        \vspace{-1.7em}
        \caption{\(d = 1\)}
        \vspace{1em}
        \label{fig:tmaze_irr_1_brc}
    \end{subfigure}
    \begin{subfigure}[t]{0.49\textwidth}
        \includegraphics[width=\textwidth]
            {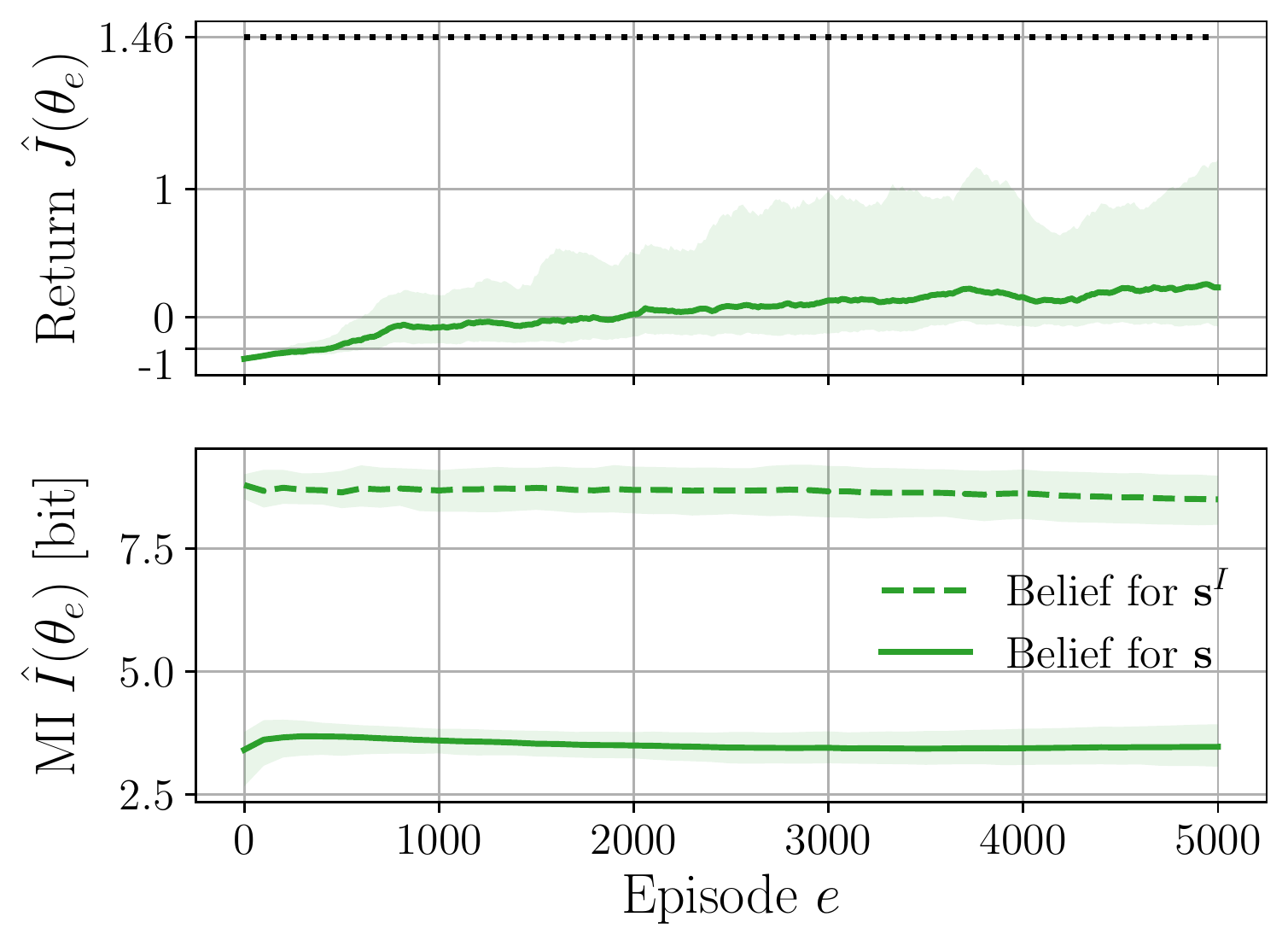}
        \vspace{-1.7em}
        \caption{\(d = 4\)}
        \vspace{1em}
        \label{fig:tmaze_irr_4_brc}
    \end{subfigure}
    \vspace{-1em}
    \caption{Deterministic T-Maze (\(L = 50\)) with \(d\) irrelevant state
        variables.
        Evolution of the return \(\hat{J}(\theta_e)\) and the MI
        \(\hat{I}(\theta_e)\) for the belief of the irrelevant and relevant
        state variables after \(e\) episodes, for the BRC cell.}
    \label{fig:tmaze_irr_brc}
\end{figure}

\begin{figure}[!hb]
    \centering
    \begin{subfigure}[t]{0.49\textwidth}
        \includegraphics[width=\textwidth]
            {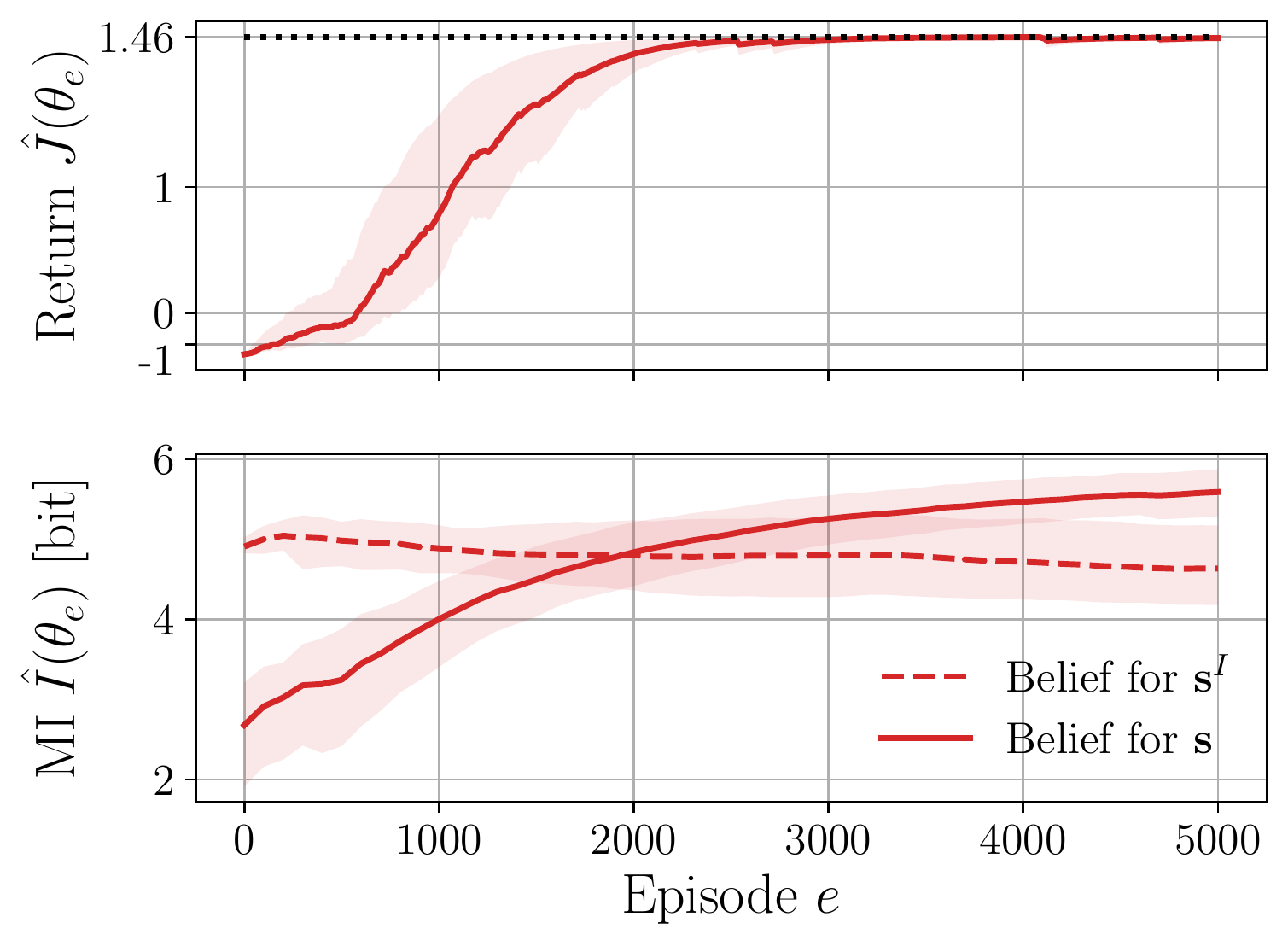}
        \vspace{-1.7em}
        \caption{\(d = 1\)}
        \vspace{1em}
        \label{fig:tmaze_irr_1_nbrc}
    \end{subfigure}
    \begin{subfigure}[t]{0.49\textwidth}
        \includegraphics[width=\textwidth]
            {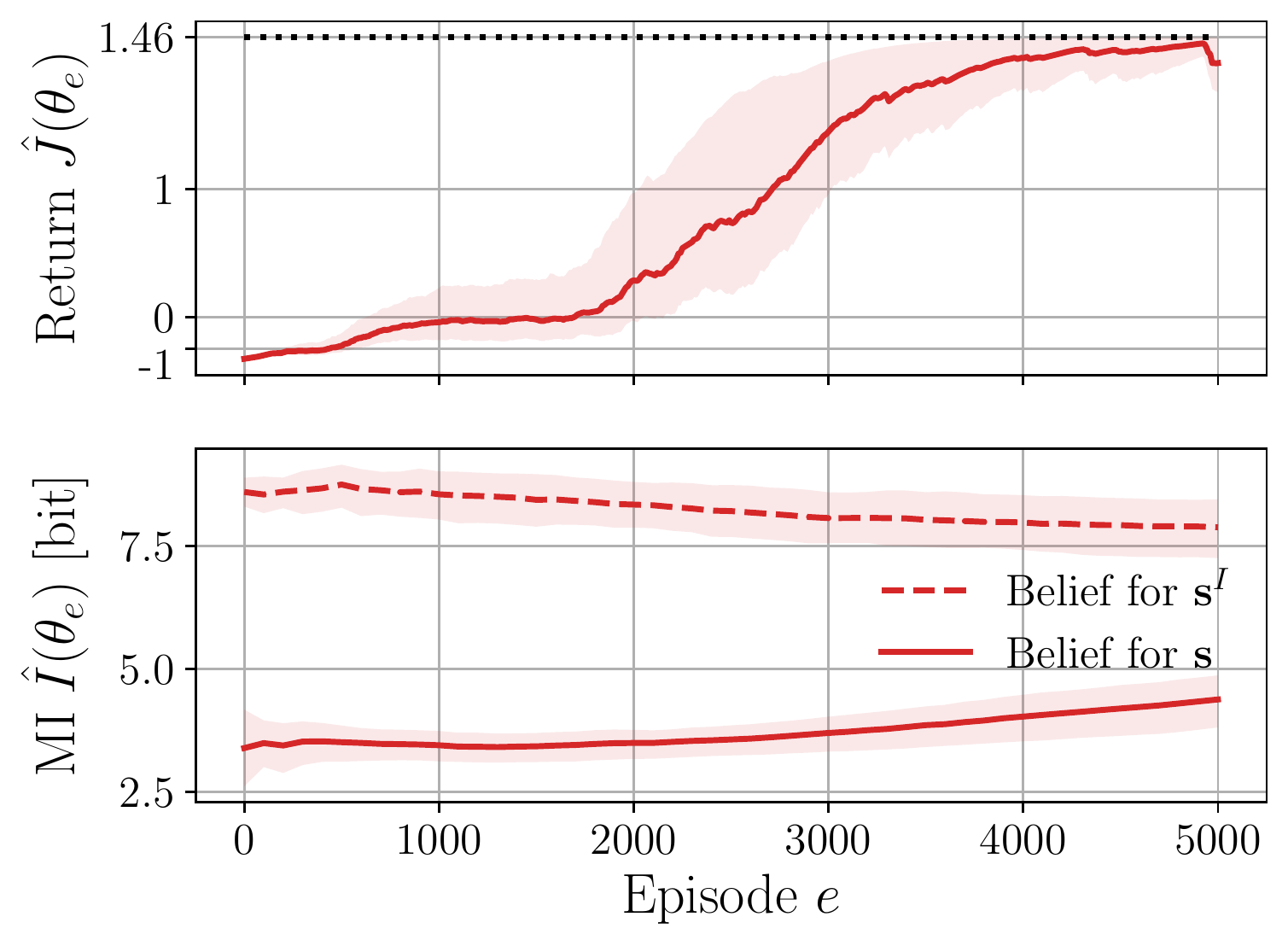}
        \vspace{-1.7em}
        \caption{\(d = 4\)}
        \vspace{1em}
        \label{fig:tmaze_irr_4_nbrc}
    \end{subfigure}
    \vspace{-1em}
    \caption{Deterministic T-Maze (\(L = 50\)) with \(d\) irrelevant state
        variables.
        Evolution of the return \(\hat{J}(\theta_e)\) and the MI
        \(\hat{I}(\theta_e)\) for the belief of the irrelevant and relevant
        state variables after \(e\) episodes, for the nBRC cell.}
    \label{fig:tmaze_irr_nbrc}
\end{figure}

\begin{figure}[!hb]
    \centering
    \begin{subfigure}[t]{0.49\textwidth}
        \includegraphics[width=\textwidth]
            {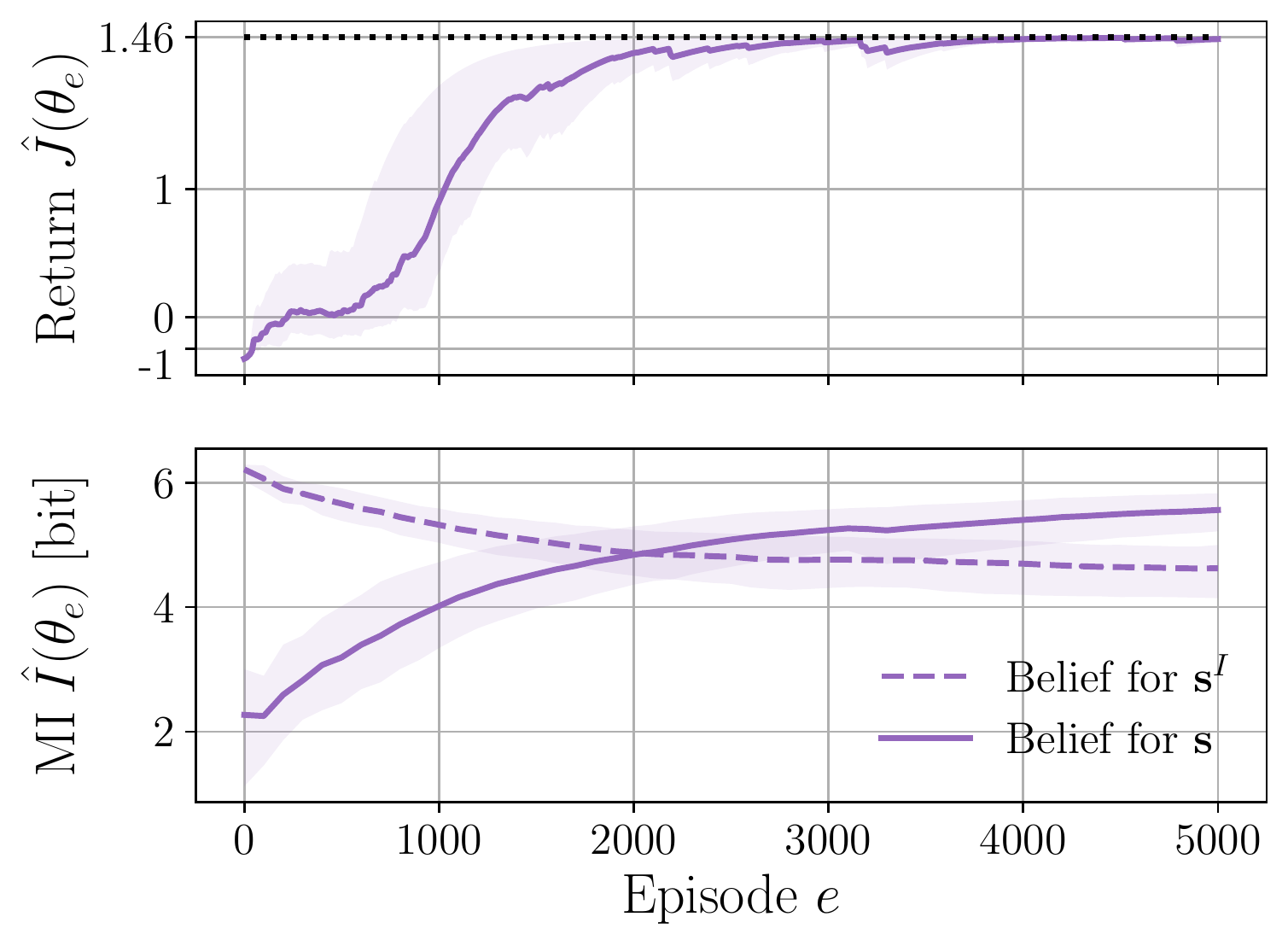}
        \vspace{-1.7em}
        \caption{\(d = 1\)}
        \vspace{1em}
        \label{fig:tmaze_irr_1_mgu}
    \end{subfigure}
    \begin{subfigure}[t]{0.49\textwidth}
        \includegraphics[width=\textwidth]
            {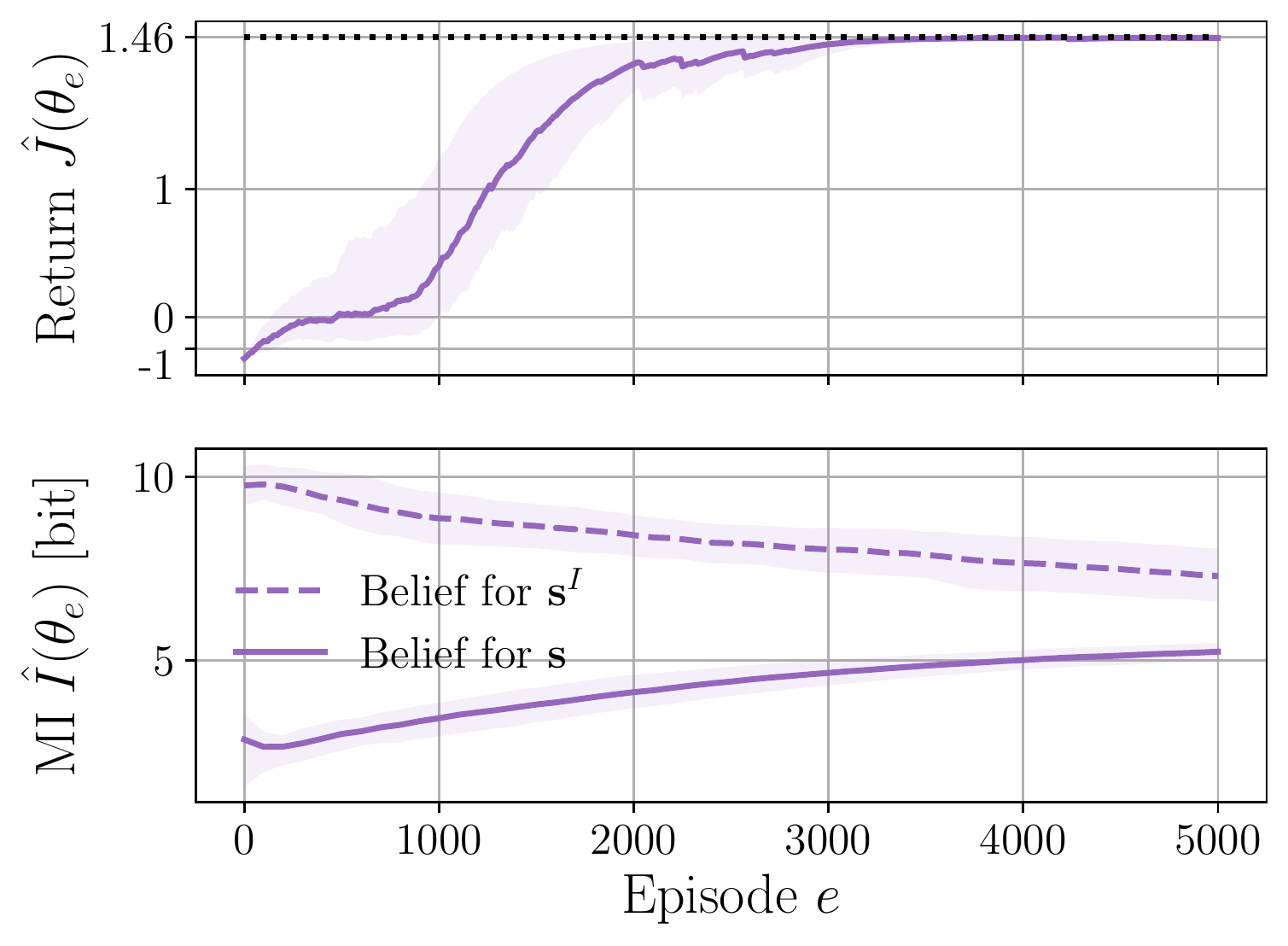}
        \vspace{-1.7em}
        \caption{\(d = 4\)}
        \vspace{1em}
        \label{fig:tmaze_irr_4_mgu}
    \end{subfigure}
    \vspace{-1em}
    \caption{Deterministic T-Maze (\(L = 50\)) with \(d\) irrelevant state
        variables.
        Evolution of the return \(\hat{J}(\theta_e)\) and the MI
        \(\hat{I}(\theta_e)\) for the belief of the irrelevant and relevant
        state variables after \(e\) episodes, for the MGU cell.}
    \label{fig:tmaze_irr_mgu}
\end{figure}

\autoref{fig:tmaze_irr_lstm}, \autoref{fig:tmaze_irr_brc},
\autoref{fig:tmaze_irr_nbrc}, and \autoref{fig:tmaze_irr_mgu} show the
evolution of the return and the MI for a T-Maze of length \(L = 50\) with $d
\in \{1, 4\}$ irrelevant state variables added to the process for these cells.
These results are reported for the GRU cell in \autoref{fig:tmaze_irr_gru} (see
\autoref{subsec:tmaze}). As can be seen from these figures, the return
generally increases with the MI between the hidden states and the belief of
state variables that are relevant for optimal control. Moreover, as for the GRU
cell, the MI between the hidden states and the belief of irrelevant state
variables generally decreases throughout the learning process.

Additionally, it can be observed that the LSTM and BRC cells fail in achieving
a near-optimal return when $d = 4$. As far as the LSTM is concerned, it is
reflected in its MI that reaches a lower value than the other RNNs. Likewise,
the BRC cell does not reach a high return, and the MI does not increase at all.
For this cell, it can be seen that the MI with the belief of irrelevant state
variables is not decreasing, even with $d = 1$. The inability of the BRC cell
to increase its MI with the belief of relevant variables and to decrease its MI
with the belief of irrelevant variables might explain its bad performance in
this environment.

\begin{figure}
    \centering
    \begin{subfigure}[t]{0.49\textwidth}
        \includegraphics[width=\textwidth]
            {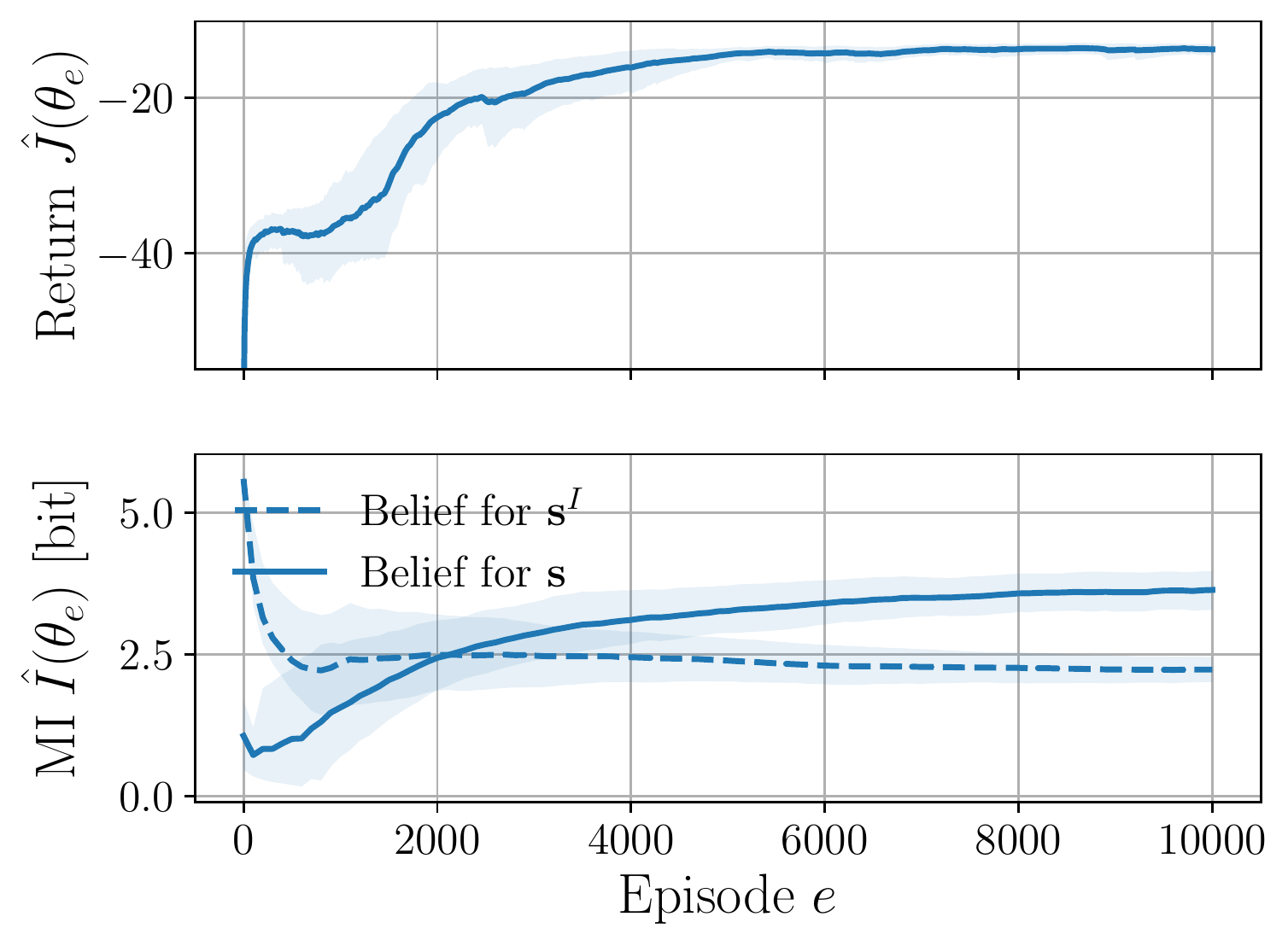}
        \vspace{-1.7em}
        \caption{\(d = 1\)}
        \vspace{1em}
        \label{fig:hike_irr_1_lstm}
    \end{subfigure}
    \begin{subfigure}[t]{0.49\textwidth}
        \includegraphics[width=\textwidth]
            {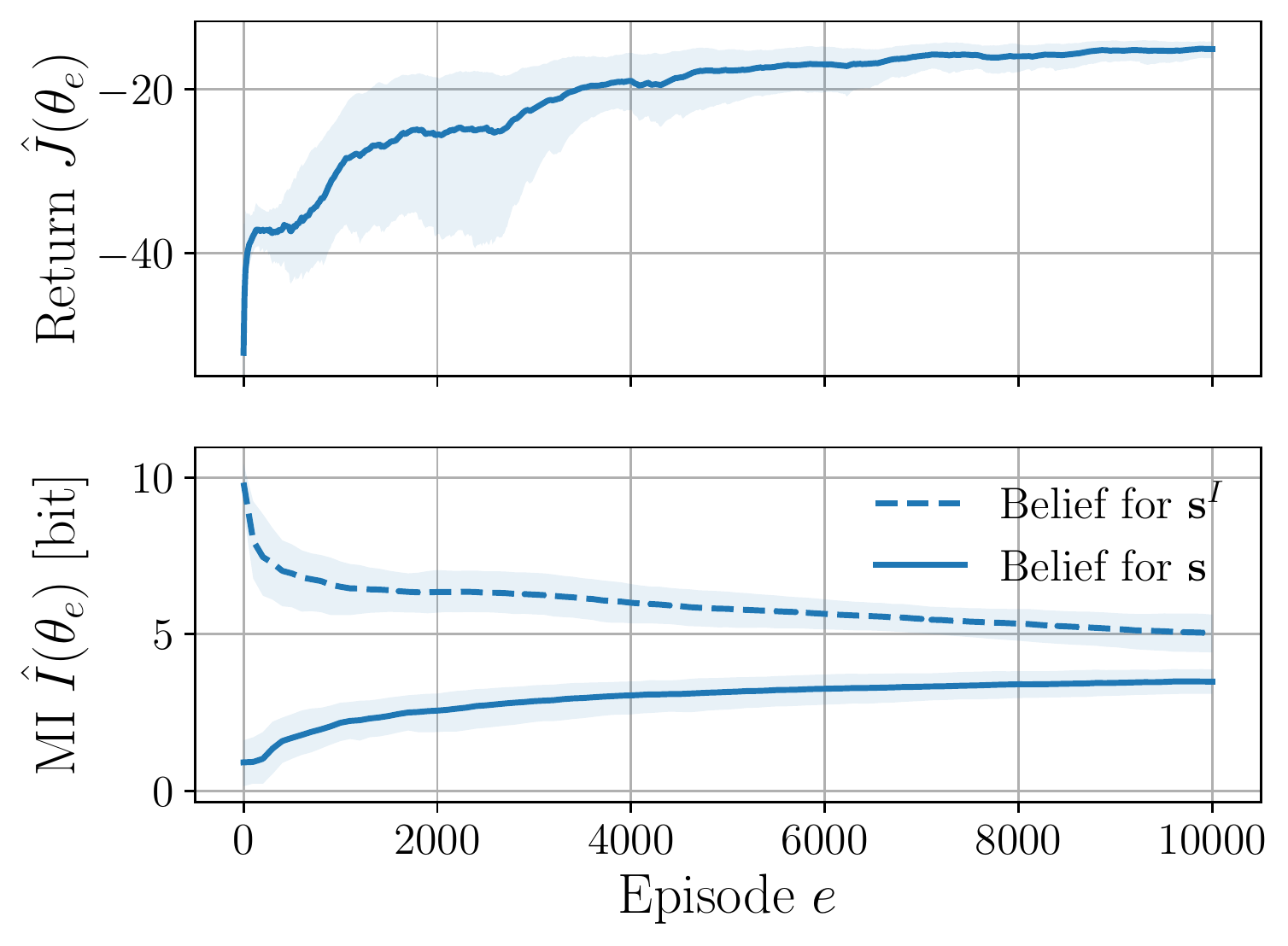}
        \vspace{-1.7em}
        \caption{\(d = 4\)}
        \vspace{1em}
        \label{fig:hike_irr_4_lstm}
    \end{subfigure}
    \vspace{-1em}
    \caption{Mountain Hike with with \(d\) irrelevant state variables.
        Evolution of the return \(\hat{J}(\theta_e)\) and the MI
        \(\hat{I}(\theta_e)\) for the belief of the irrelevant and relevant
        state variables after \(e\) episodes, for the LSTM cell.}
    \label{fig:hike_irr_lstm}
\end{figure}

\begin{figure}
    \centering
    \begin{subfigure}[t]{0.49\textwidth}
        \includegraphics[width=\textwidth]
            {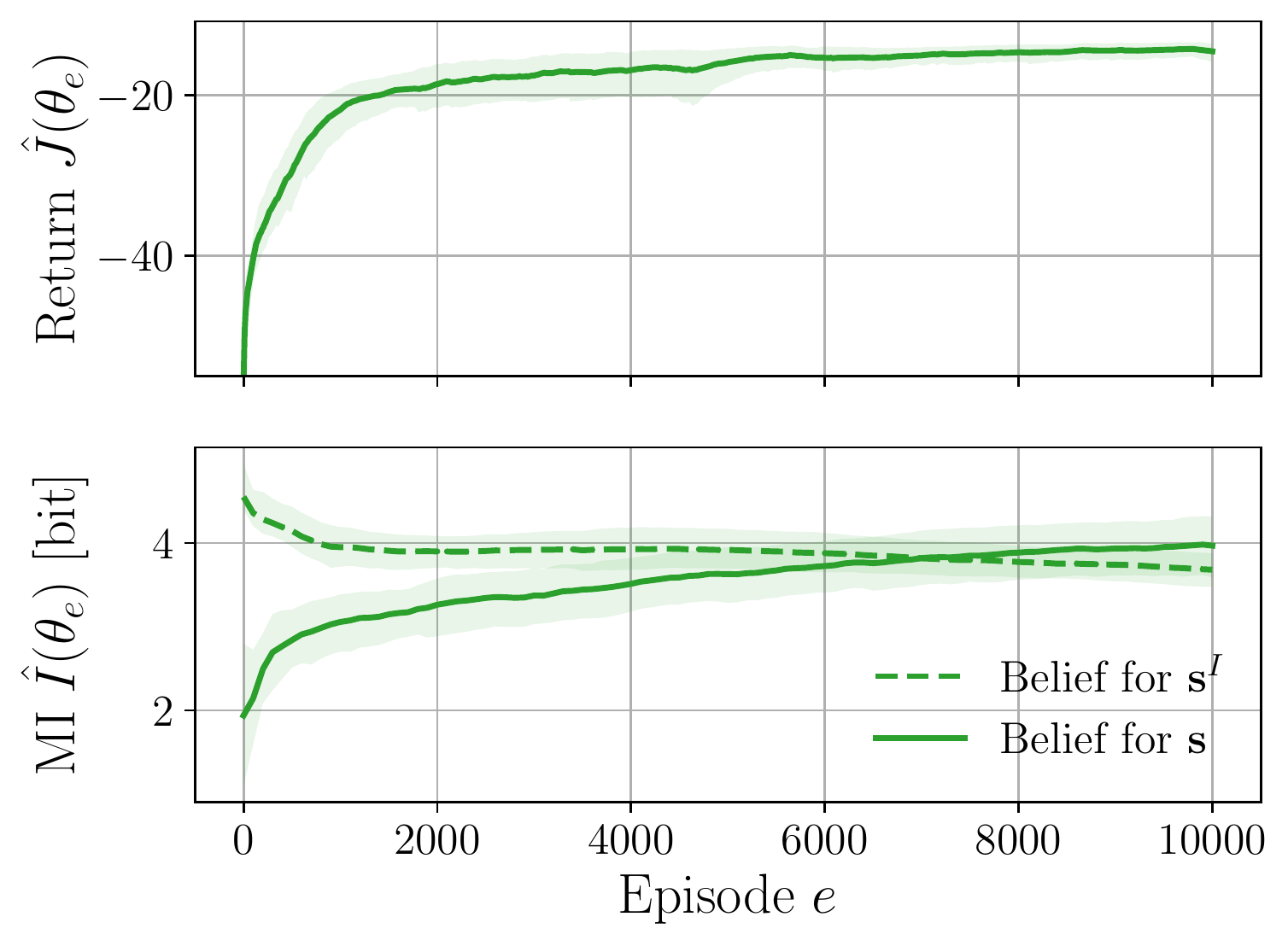}
        \vspace{-1.7em}
        \caption{\(d = 1\)}
        \vspace{1em}
        \label{fig:hike_irr_1_brc}
    \end{subfigure}
    \begin{subfigure}[t]{0.49\textwidth}
        \includegraphics[width=\textwidth]
            {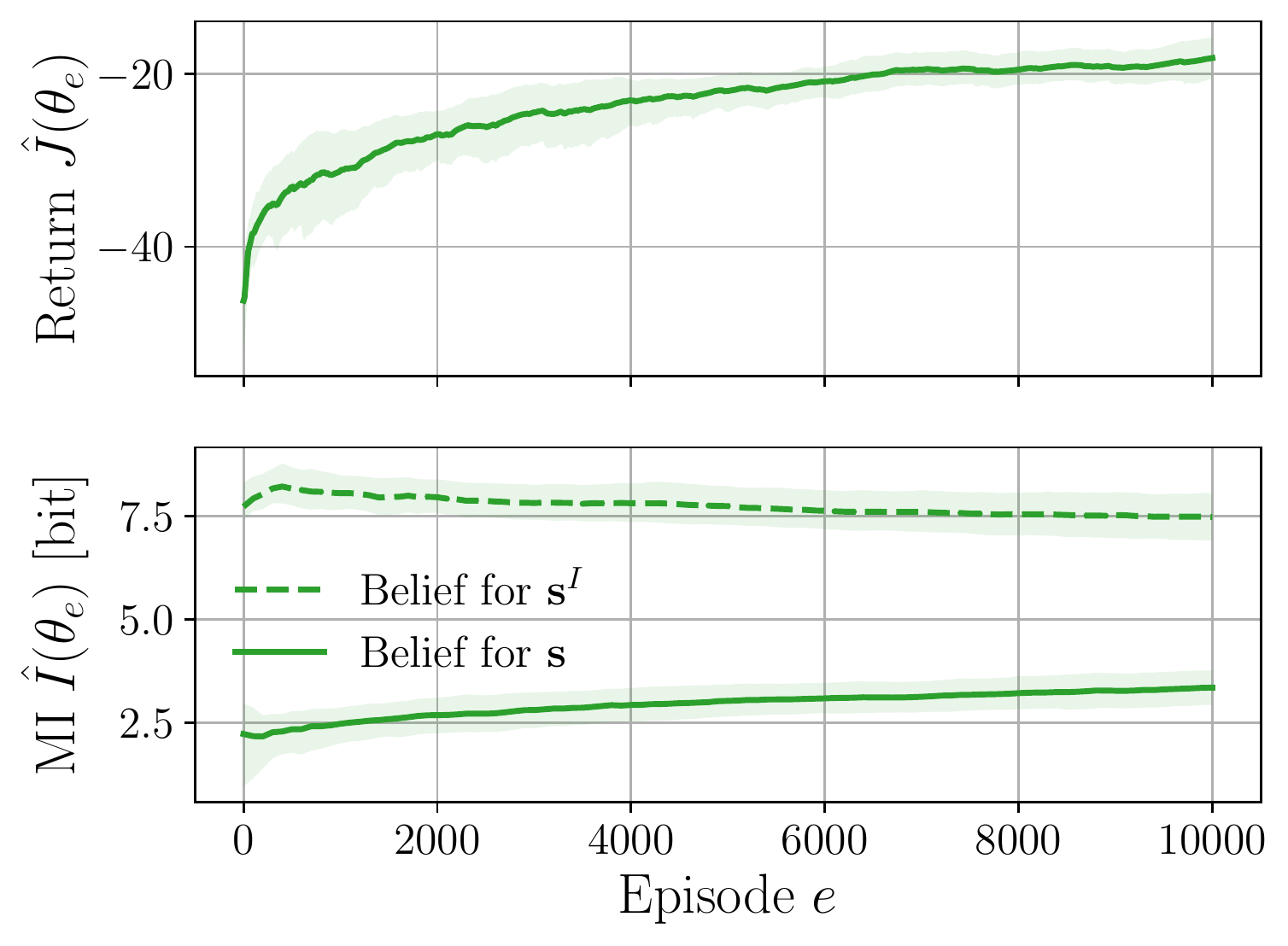}
        \vspace{-1.7em}
        \caption{\(d = 4\)}
        \vspace{1em}
        \label{fig:hike_irr_4_brc}
    \end{subfigure}
    \vspace{-1em}
    \caption{Mountain Hike with with \(d\) irrelevant state variables.
        Evolution of the return \(\hat{J}(\theta_e)\) and the MI
        \(\hat{I}(\theta_e)\) for the belief of the irrelevant and relevant
        state variables after \(e\) episodes, for the BRC cell.}
    \label{fig:hike_irr_brc}
\end{figure}

\begin{figure}
    \centering
    \begin{subfigure}[t]{0.49\textwidth}
        \includegraphics[width=\textwidth]
            {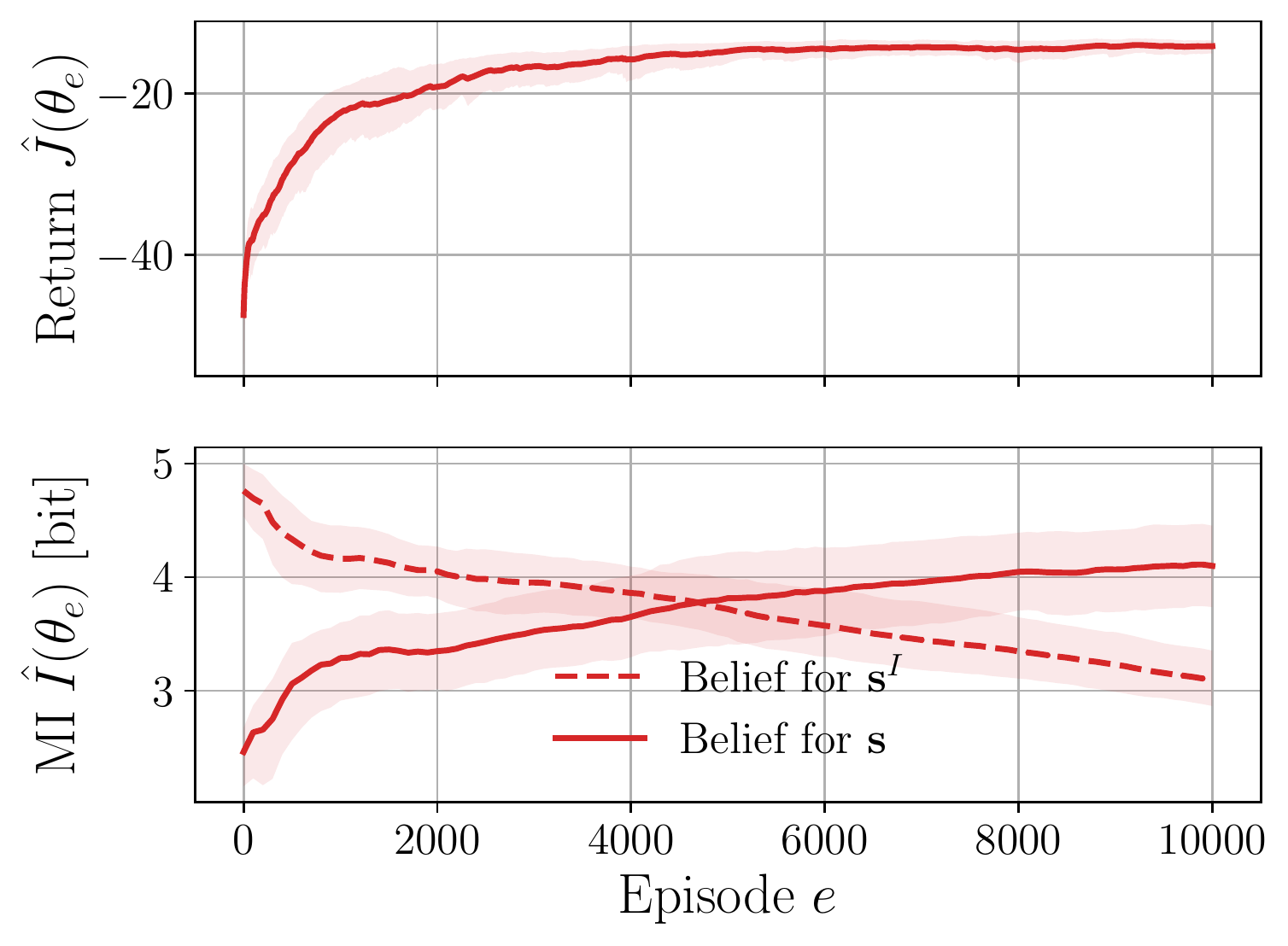}
        \vspace{-1.7em}
        \caption{\(d = 1\)}
        \vspace{1em}
        \label{fig:hike_irr_1_nbrc}
    \end{subfigure}
    \begin{subfigure}[t]{0.49\textwidth}
        \includegraphics[width=\textwidth]
            {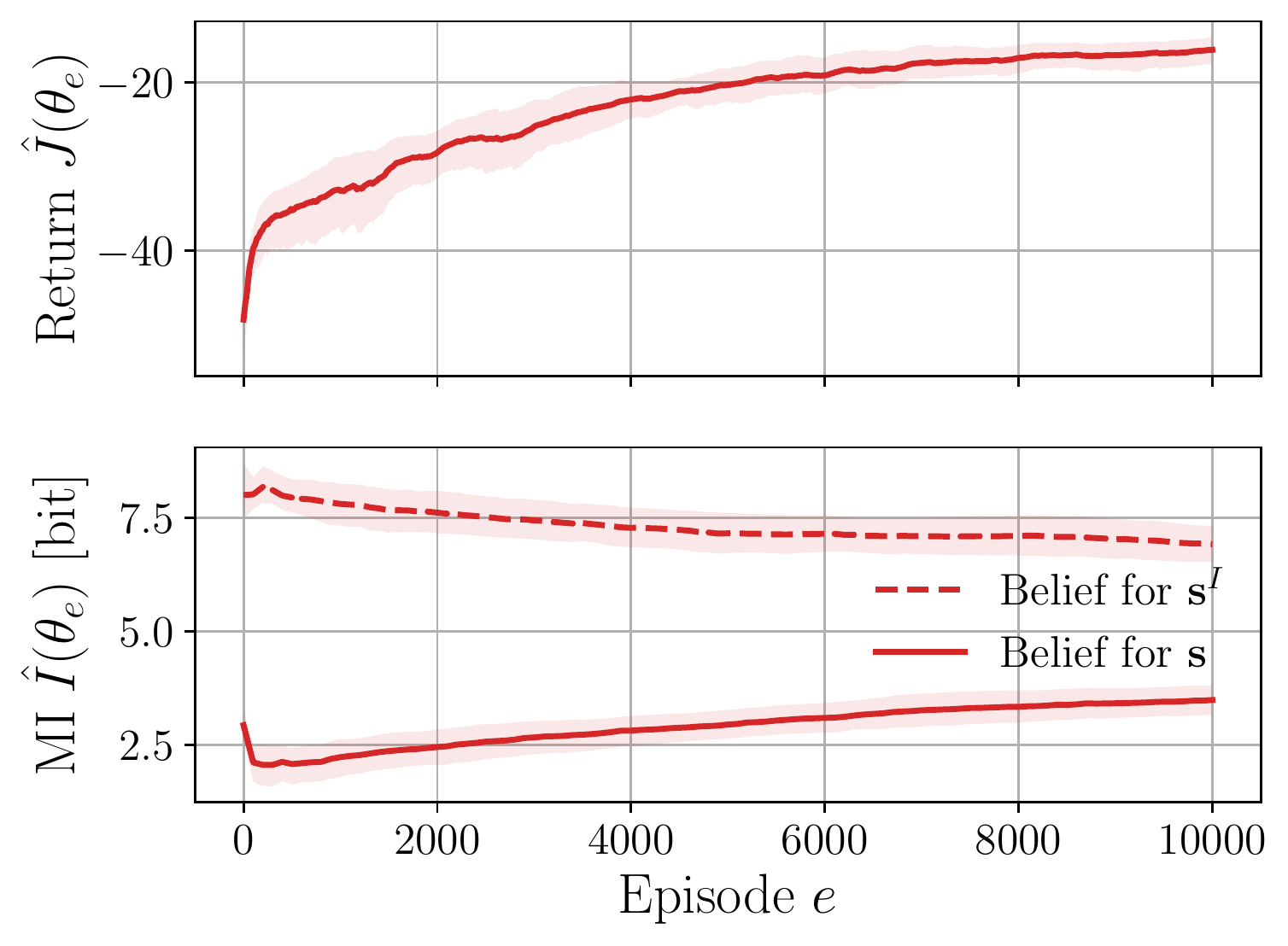}
        \vspace{-1.7em}
        \caption{\(d = 4\)}
        \vspace{1em}
        \label{fig:hike_irr_4_nbrc}
    \end{subfigure}
    \vspace{-1em}
    \caption{Mountain Hike with with \(d\) irrelevant state variables.
        Evolution of the return \(\hat{J}(\theta_e)\) and the MI
        \(\hat{I}(\theta_e)\) for the belief of the irrelevant and relevant
        state variables after \(e\) episodes, for the nBRC cell.}
    \label{fig:hike_irr_nbrc}
\end{figure}

\begin{figure}
    \centering
    \begin{subfigure}[t]{0.49\textwidth}
        \includegraphics[width=\textwidth]
            {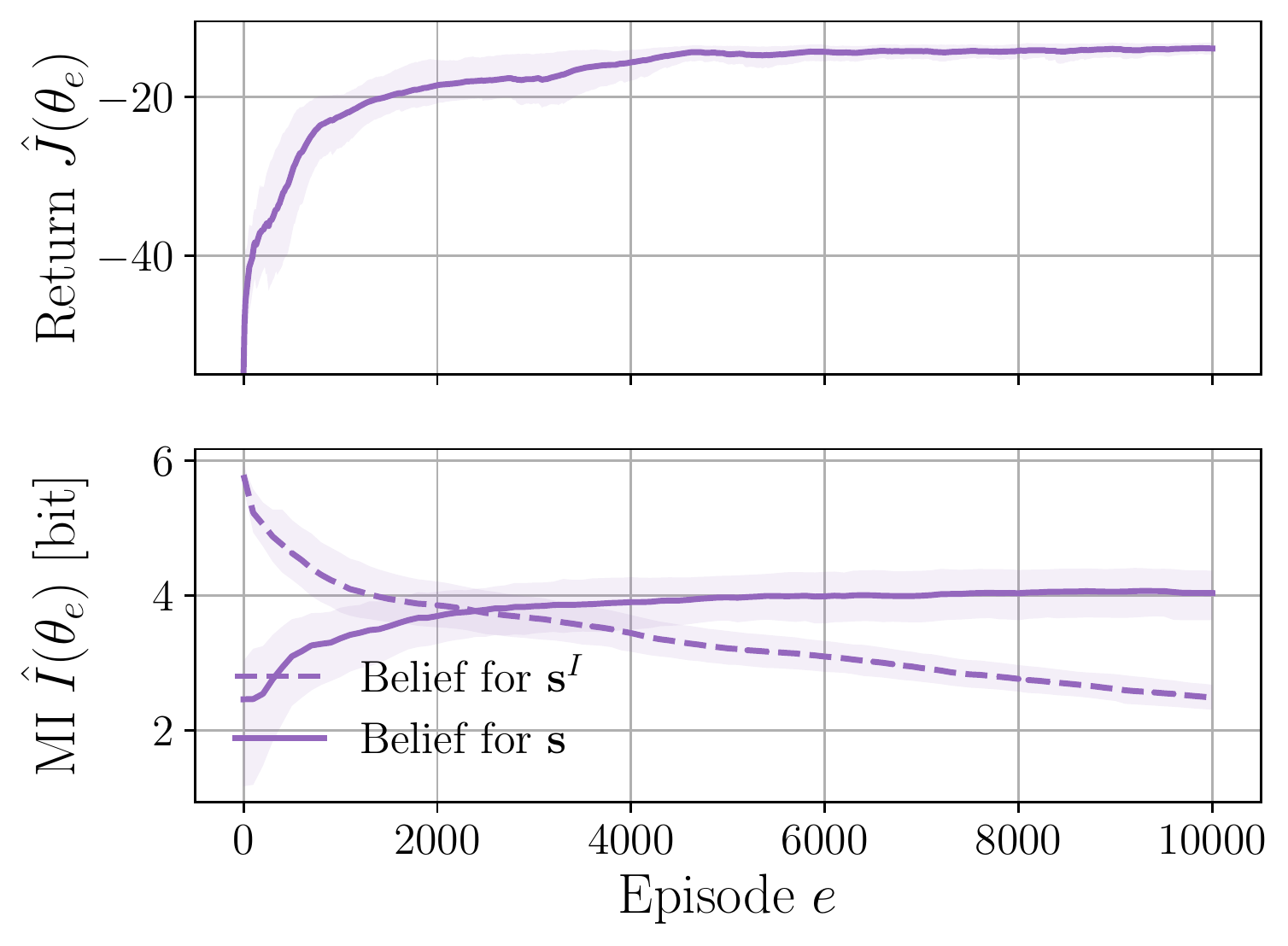}
        \vspace{-1.7em}
        \caption{\(d = 1\)}
        \vspace{1em}
        \label{fig:hike_irr_1_mgu}
    \end{subfigure}
    \begin{subfigure}[t]{0.49\textwidth}
        \includegraphics[width=\textwidth]
            {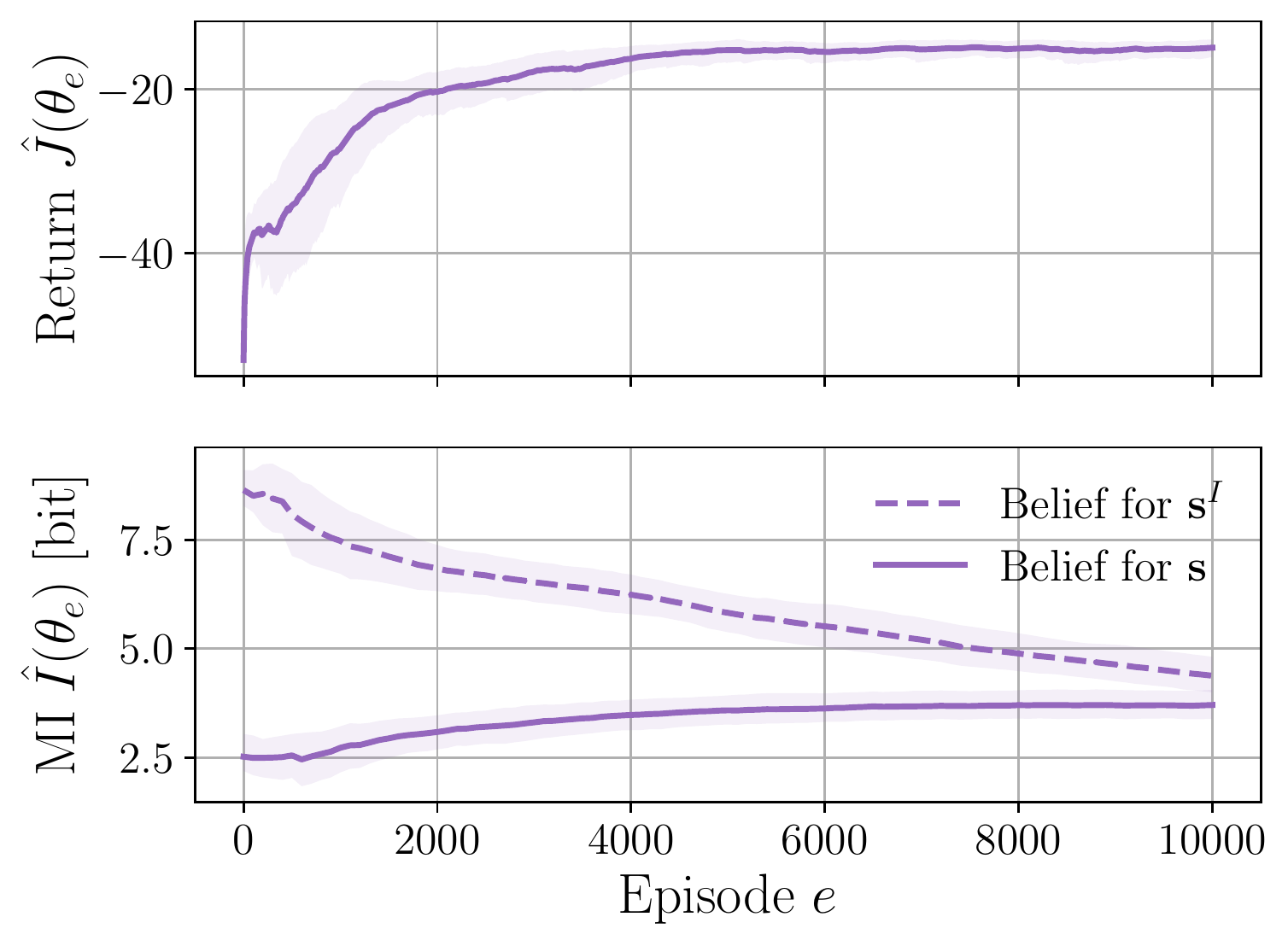}
        \vspace{-1.7em}
        \caption{\(d = 4\)}
        \vspace{1em}
        \label{fig:hike_irr_4_mgu}
    \end{subfigure}
    \vspace{-1em}
    \caption{Mountain Hike with with \(d\) irrelevant state variables.
        Evolution of the return \(\hat{J}(\theta_e)\) and the MI
        \(\hat{I}(\theta_e)\) for the belief of the irrelevant and relevant
        state variables after \(e\) episodes, for the MGU cell.}
    \label{fig:hike_irr_mgu}
\end{figure}

As far as the Mountain Hike is concerned, \autoref{fig:hike_irr_lstm},
\autoref{fig:hike_irr_brc}, \autoref{fig:hike_irr_nbrc} and
\autoref{fig:hike_irr_mgu} show that all previous observations also hold for
this environment with the LSTM, BRC, nBRC and MGU cells. These results are
reported for the GRU cell in \autoref{fig:hike_irr_gru} (see
\autoref{subsec:tmaze}). As can be seen from these figures, the return clearly
increases with the MI between the hidden states and the belief of relevant
state variables, for all cells. In contrast, the MI with the belief of
irrelevant state variables decreases throughout the learning process.

\end{document}